%% file: 00_main.tex
\pdfoutput=1

\documentclass[11pt]{article}

\usepackage{acl}
\usepackage{lipsum}
\usepackage{times}
\usepackage{latexsym}
\usepackage{subcaption}
\usepackage{multirow}
\usepackage{xcolor}
\usepackage[T1]{fontenc}
\usepackage{setspace}
\usepackage{multicol}
\usepackage{tikz}

\usepackage[utf8]{inputenc}

\usepackage{booktabs}
\usepackage{graphicx}
\usepackage{subcaption}
\usepackage{microtype}

\usepackage{inconsolata}
\usepackage[most]{tcolorbox}

\title{An Empirical Analysis on Large Language Models in Debate Evaluation}

\author{
Xinyi Liu\(^{*}\) \quad Pinxin Liu\thanks{\(^{*}\) Equal contribution} \quad  Hangfeng He\\ 
 University of Rochester\\
 \texttt{xinyi.liu1@simon.rochester.edu} \\
 \texttt{pliu23@u.rochester.edu} \\
 \texttt{hangfeng.he@rochester.edu} \\
}

\begin{document}

\maketitle
\begin{abstract}


In this study, we investigate the capabilities and inherent biases of advanced large language models (LLMs) such as GPT-3.5 and GPT-4 in the context of debate evaluation. We discover that LLM's performance exceeds humans and surpasses the performance of state-of-the-art methods fine-tuned on extensive datasets in debate evaluation. We additionally explore and analyze biases present in LLMs, including positional bias, lexical bias, order bias, which may affect their evaluative judgments. Our findings reveal a consistent bias in both GPT-3.5 and GPT-4 towards the second candidate response presented, attributed to prompt design. We also uncover lexical biases in both GPT-3.5 and GPT-4, especially when label sets carry connotations such as numerical or sequential, highlighting the critical need for careful label verbalizer selection in prompt design. Additionally, our analysis indicates a tendency of both models to favor the debate's concluding side as the winner, suggesting an end-of-discussion bias.\footnote{Our code is publicly available at \url{https://github.com/XinyiLiu0227/LLM\_Debate\_Bias/}} 



\end{abstract}

\input{01_introduction}

\input{02_methodology}

\input{03_experiments}

\input{04_results}

\input{05_conclusion}


\bibliography{custom}

\appendix

\clearpage

\input{06_appendix}

\end{document}

%% file: 01_introduction.tex
\section{Introduction}


Prior research in automatic debate evaluation has predominantly relied on pre-trained encoders and the modeling of argument relations and structures~\cite{hsiao2022modeling, li2020exploring, ruiz2022automatic, zhang2023argument}. A significant drawback of these approaches is their dependency on feature engineering and extensive data training, limiting their generalizability across diverse datasets.

\begin{figure}
    \centering
    \includegraphics[height=0.25\textheight]{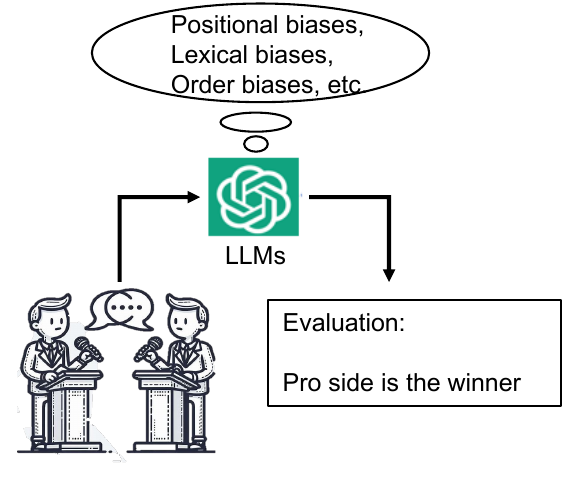}
    \vspace{-2mm}
    \caption{Large Language Models presents various biases during the evaluation of long debates.}
    \label{fig:enter-label}
\end{figure}

The advent of advanced large language models (LLMs) such as GPT-3.5 and GPT-4~\cite{achiam2023gpt} has marked the beginning of a new era in automating a wide spectrum of complex tasks 
~\cite{
wei2022chain, thirunavukarasu2023large, jingyang2023a, wang2023document, tang2023llmva, zhang2024cocot, jiang2023llm}. These models have been increasingly utilized as automatic evaluators \cite{chiang2023can, chiang2023closer, lin2023llm, chan2023chateval, zeng2023evaluating, he2023socreval}. 
Leveraging LLMs for debate evaluation presents more challenges, including the extended duration of debates, evolving argument dynamics, and the necessity for evaluators to rely on comprehensive knowledge and reasoning that extend beyond the immediate scope of the debate. Our research delves into the utilization of LLMs for debate evaluation, uncovering their zero-shot capabilities that parallel human evaluators and surpass all existing state-of-the-art (SOTA) methods fine-tuned on ample data~\cite{li2020exploring, hsiao2022modeling}. 

We further investigate potential biases in GPT-3.5 and GPT-4 within the context of debate evaluation. While previous research has identified various biases in LLMs, such as persona bias~\cite{wan2023personalized}, political bias~\cite{feng2023pretraining}, and positional bias~\cite{wang2023large}, our investigation uniquely concentrates on biases affecting debate evaluation performance, a relatively unexplored domain.


Specifically, upon comparing outcomes between scenarios where the positions of candidate responses are switched, persistent bias has been observed in both GPT-3.5 and GPT-4 toward the second candidate response presented, a \textit{positional bias} induced 
by the prompt design. Beyond this, both models also display  significant \textit{lexical biases}, particularly when label sets carry connotations such as sequential or magnitude, underscoring the importance of careful selection of label verbalizers in prompt design to mitigate unintended biases \cite{liu2023pre}. Moreover, our study reveals that both GPT-3.5 and GPT-4 exhibit a tendency to favor the concluding side of a debate as the winner, pointing to a potential end-of-discussion \textit{order bias}. Interestingly, after all the identified biases are eliminated, GPT-3.5 still demonstrates a consistent bias, while this residual bias is less obvious for GPT-4. These insights highlight the nuanced nature of biases in LLMs and the complexity of designing fair and unbiased evaluation methodologies for debate evaluation.


%% file: 02_methodology.tex
\label{sec:methodology}

\section{Methodology}


\paragraph{LLMs' capability for debate evaluation.} As illustrated in Table~\ref{tab:template}, we utilize an evaluation template $T$ with two placeholders, $T (\text{Side1\_label, Side2\_label})$, to solicit debate assessments from LLMs. For each debate topic, two labels are assigned to represent the Pro and Con sides, respectively. These labels are then inserted into the designated slots of the evaluation template, creating a prompt $T (\text{Side1\_label}=\text{P}, \text{Side2\_label}=\text{C})$. This tailored prompt is subsequently employed to query the LLM for its judgment on the debate. Experiments with other templates are presented in the Appendix \ref{sec: appendix 3}.



\paragraph{LLMs' biases in debate evaluation.} Analyzing errors and biases in LLMs is complex due to their multifaceted decision-making processes, which are influenced by inherent orders, stances, and the nuances of prompt design. To separate LLM-induced biases from dataset biases, we use a balanced dataset, ensuring class representation equality and mitigating skewness from overrepresentation. This method focuses on LLM-specific biases within a controlled environment, minimizing dataset composition effects. We explore LLM sensitivity to label choice from alphabetic (A/B, P/C), numerical (1/-1), to textual format (Pro/Con), providing a comprehensive bias analysis. Our study also examines the applicability of these findings in real-world, original distribution settings, extending the relevance of our insights.




\begin{table}[t]
\centering 
\small
\begin{tcolorbox}
\textbf{Content}

The content of the whole debate:

The current speech in the debate is from the user \textcolor[rgb]{0,0,0.9}{\{Side1\_label\}}:

[The content of the side1]

The current speech in the debate is from the user \textcolor[rgb]{0,0,0.9}{\{Side2\_label\}}:

[The content of the side2]

The current speech in the debate is from the user \textcolor[rgb]{0,0,0.9}{\{Side1\_label\}}:

[The content of the side1]

\ldots

\textbf{Prompt}

Assume you are a debate evaluator, there are two participants in this debate. Given the above context of the whole debate. Please give the decision on which participant is the winner, you only need to give the character(number) of either \textcolor[rgb]{0,0,0.9}{\{Side1\_label\}}, or \textcolor[rgb]{0,0,0.9}{\{Side2\_label\}}. \textcolor[rgb]{0,0,0.9}{\{Side1\_label\}} means user \textcolor[rgb]{0,0,0.9}{\{Side1\_label\}} wins. \textcolor[rgb]{0,0,0.9}{\{Side2\_label\}} means user \textcolor[rgb]{0,0,0.9}{\{Side2\_label\}} wins. Please only give the result without any other words.

\end{tcolorbox}
\caption{Our evaluation template for determining debate outcomes for Large Language Models. We provide a specific 3-round debate example in Appendix~\ref{sec: appendix 4}}
\label{tab:template}
\end{table}

\begin{figure*}[t]
    \centering
    \captionsetup[subfigure]{font=scriptsize}

    \begin{subfigure}[b]{0.24\linewidth}
        \includegraphics[width=\linewidth]{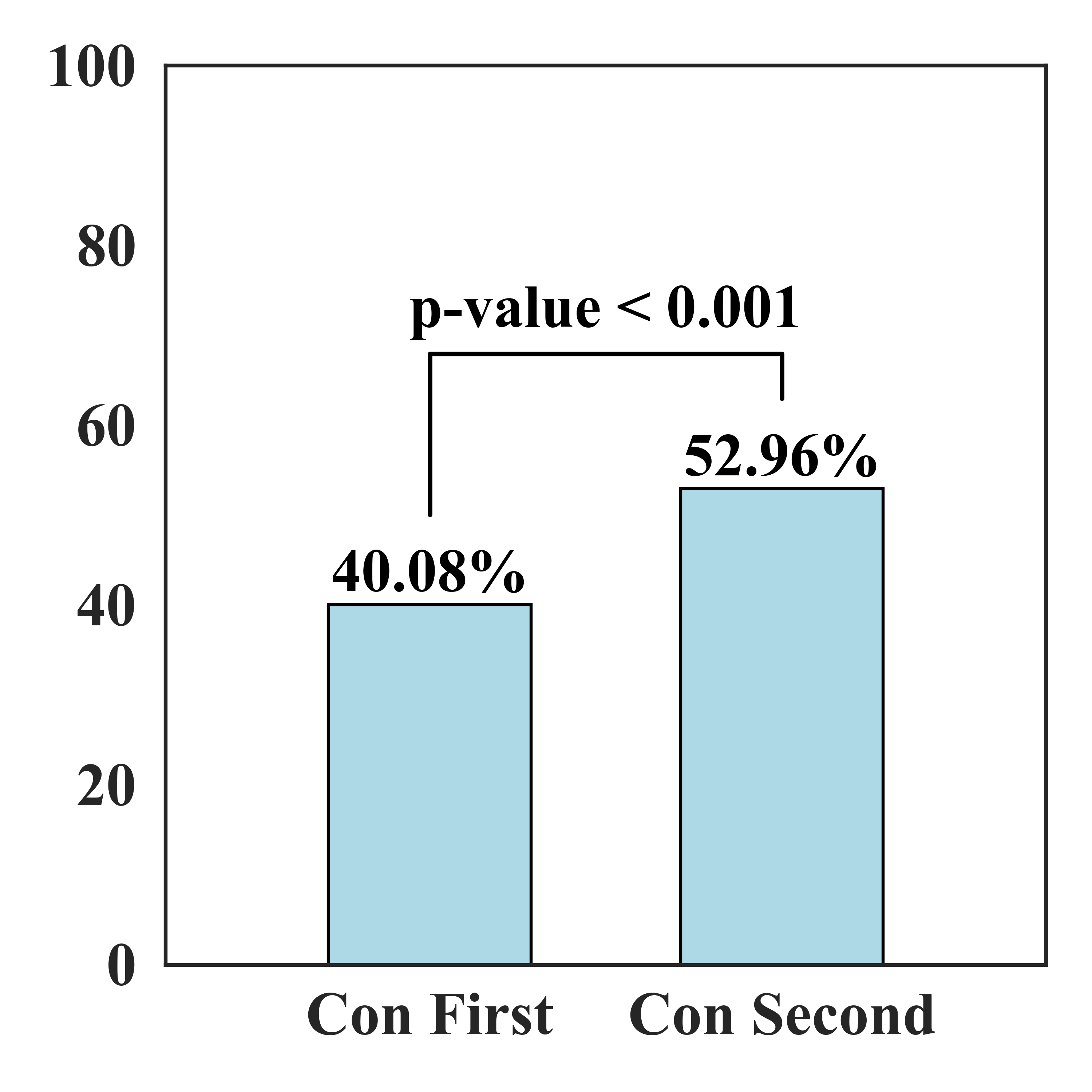}
        \caption{A/B label set}
        \label{fig:positional bias GPT3.5 sub1}
    \end{subfigure}
    \hfill 
    \begin{subfigure}[b]{0.24\linewidth}
        \includegraphics[width=\linewidth]{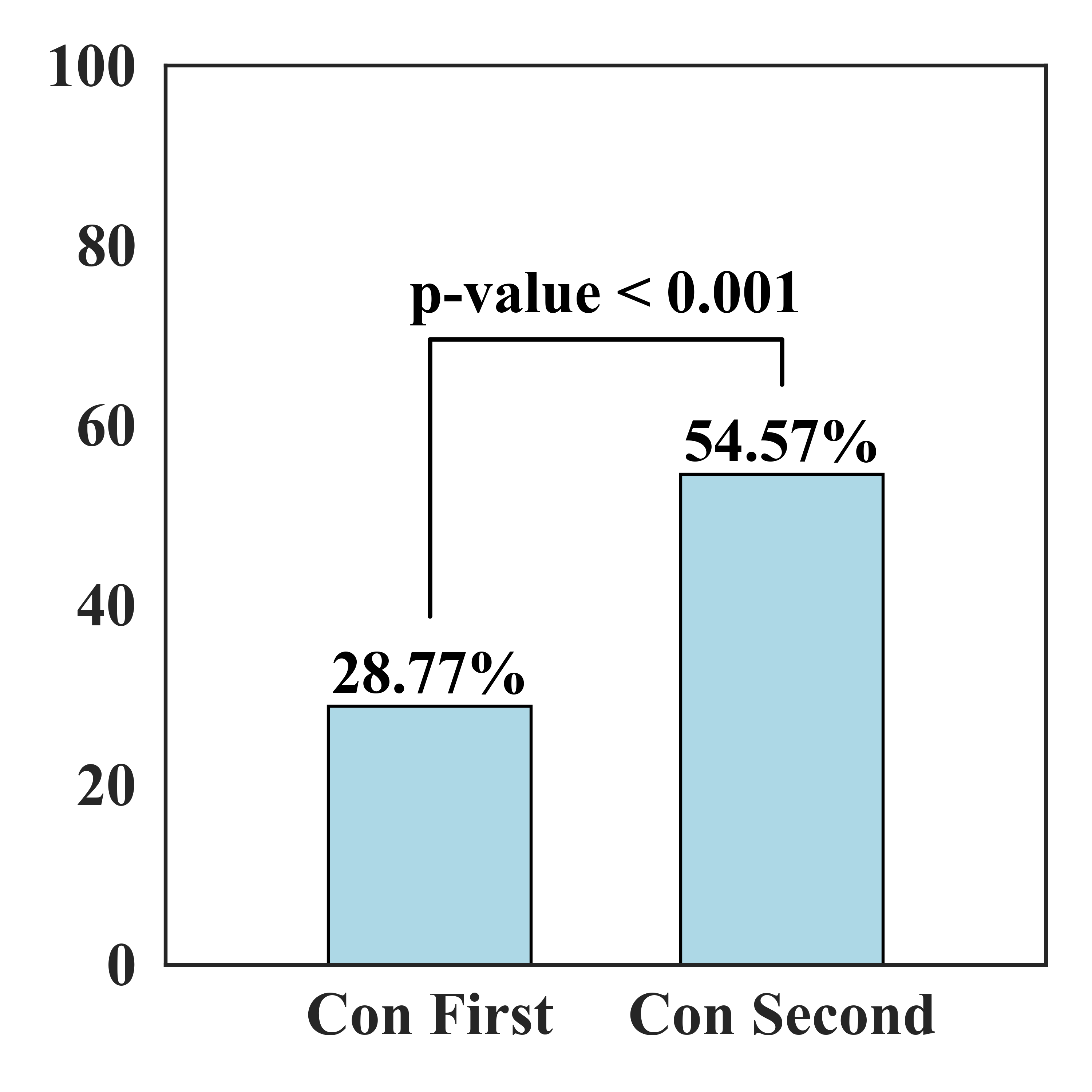}
        \caption{P/C label set}
        \label{fig:positional bias GPT3.5 sub2}
    \end{subfigure}
    \hfill 
    \begin{subfigure}[b]{0.24\linewidth}
        \includegraphics[width=\linewidth]{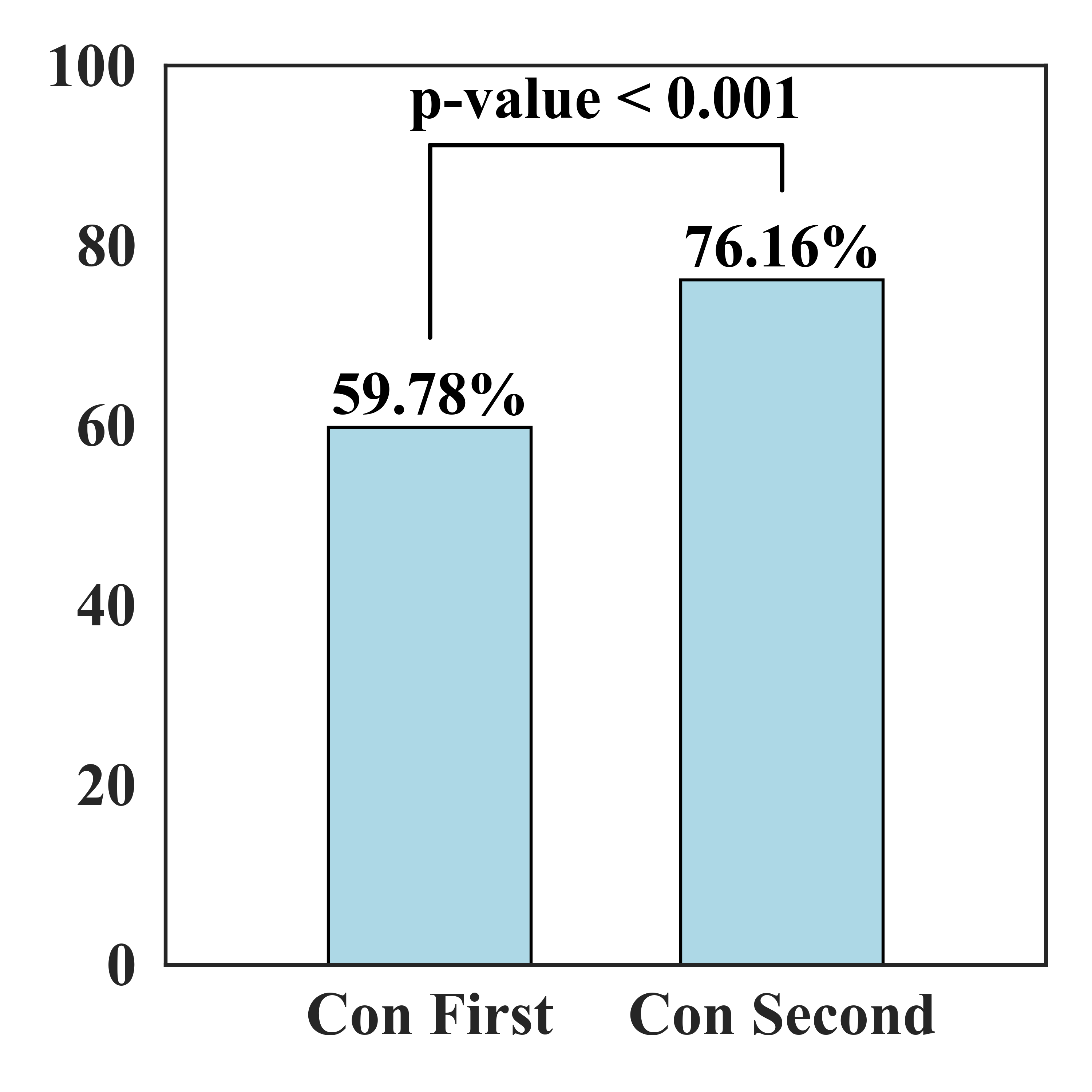}
        \caption{1/-1 label set}
        \label{fig:positional bias GPT3.5 sub3}
    \end{subfigure}
    \hfill 
    \begin{subfigure}[b]{0.24\linewidth}
        \includegraphics[width=\linewidth]{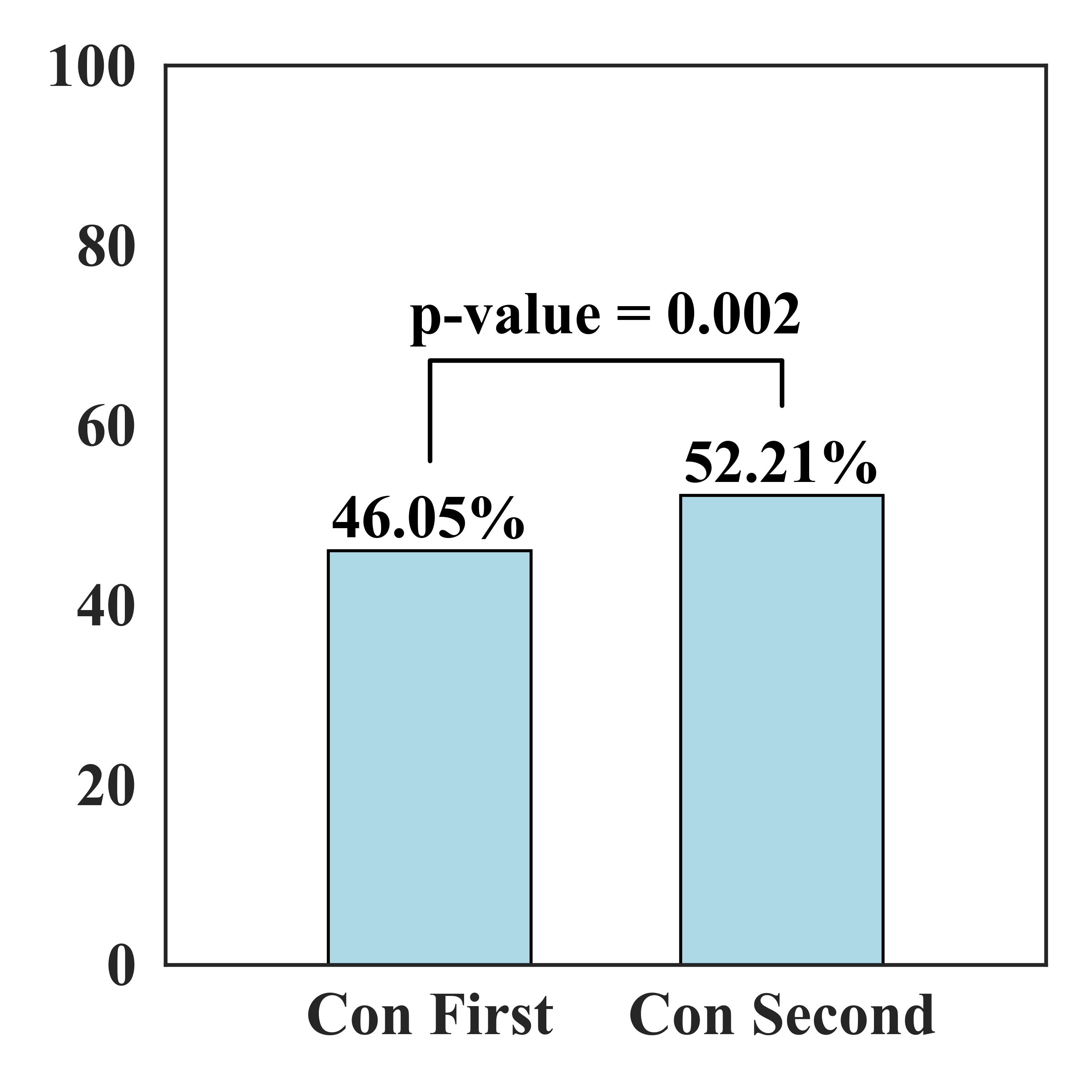}
        \caption{Pro/Con label set}
        \label{fig:positional bias GPT3.5 sub4}
    \end{subfigure}

    \caption{The observed positional bias in GPT-3.5 is evident through the alteration in the proportion of Predicted Con outcomes, which increases when Con is positioned as the second candidate response compared to its placement as the first. This consistent preference across all label configurations suggests a systematic positional bias favoring the second candidate, underscoring the model's sensitivity to the order in which options are presented.}
    \label{fig: positional bias in GPT-3.5}
\end{figure*}

\begin{figure*}[t]
    \centering
    \captionsetup[subfigure]{font=scriptsize}

    \begin{subfigure}[b]{0.24\linewidth}
        \includegraphics[width=\linewidth]{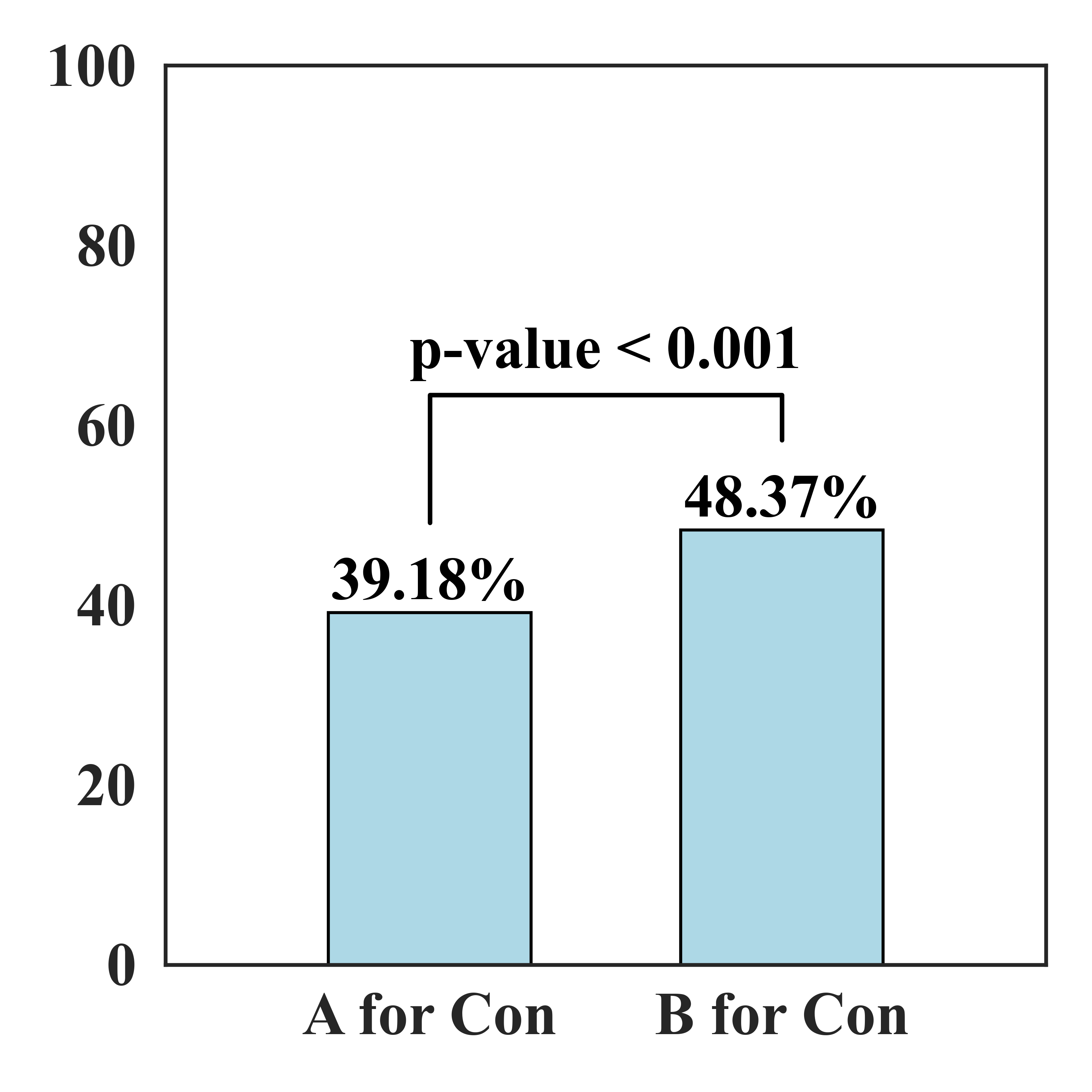}
        \caption{Shuffled A/B vs. Shuffled B/A}
        \label{fig:lexical bias GPT3.5 sub1}
    \end{subfigure}
    \hfill 
    \begin{subfigure}[b]{0.24\linewidth}
        \includegraphics[width=\linewidth]{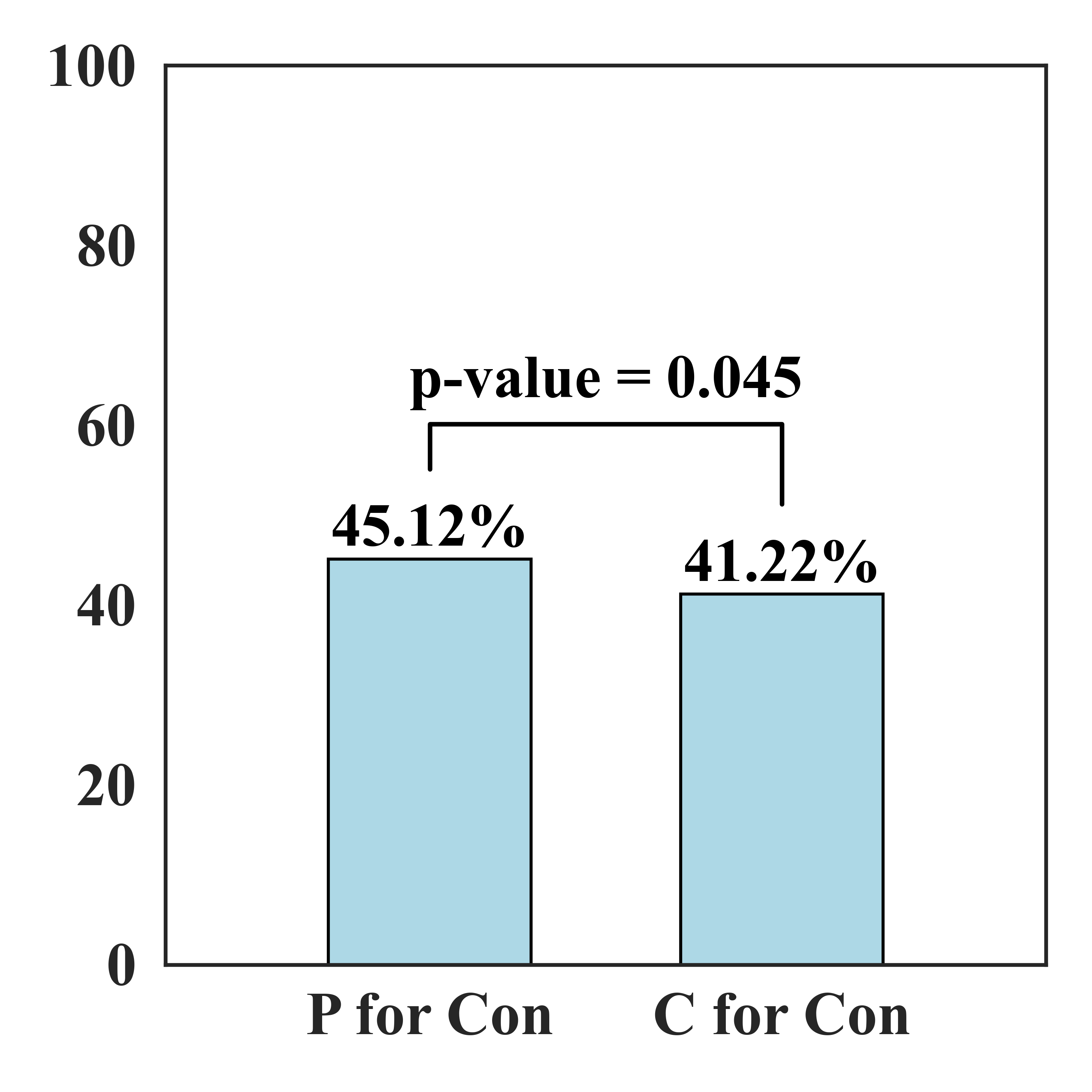}
        \caption{Shuffled P/C vs. Shuffled C/P}
        \label{fig:lexical bias GPT3.5 sub2}
    \end{subfigure}
    \hfill 
    \begin{subfigure}[b]{0.24\linewidth}
        \includegraphics[width=\linewidth]{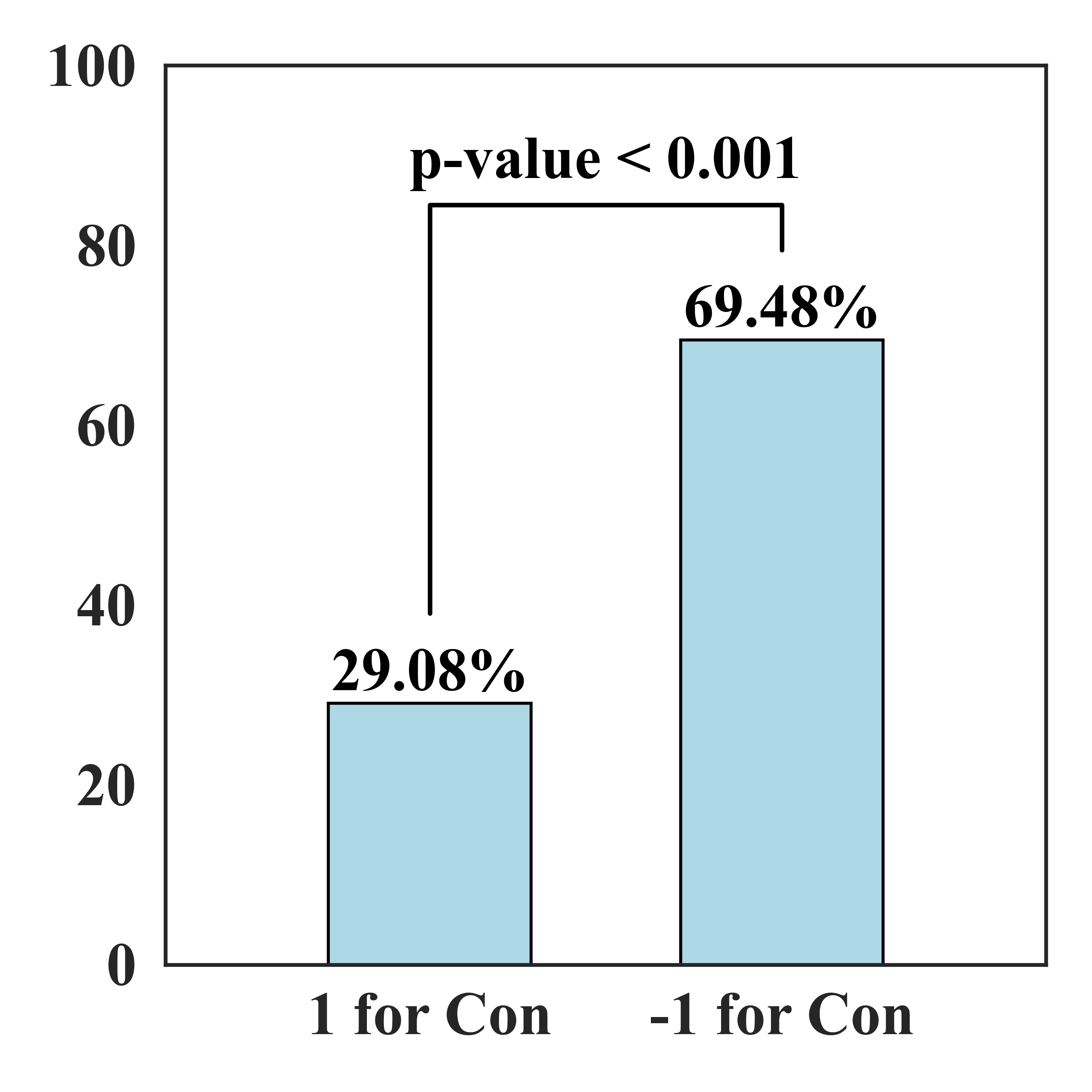}
        \caption{Shuffled 1/-1 vs. Shuffled -1/1}
        \label{fig:lexical bias GPT3.5 sub3}
    \end{subfigure}
    \hfill 
    \begin{subfigure}[b]{0.24\linewidth}
        \includegraphics[width=\linewidth]{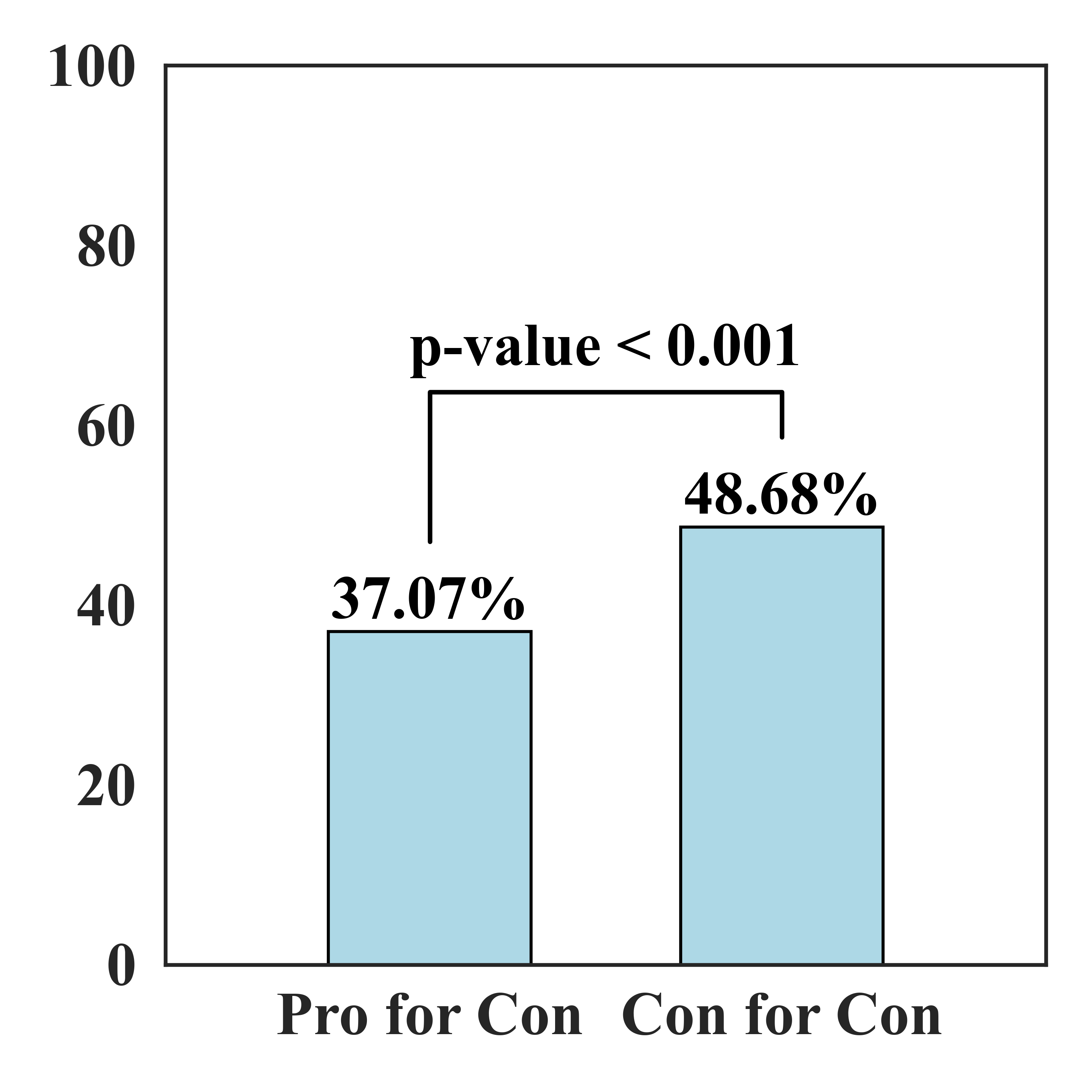}
        \caption{\tiny Shuffled Pro/Con vs. Shuffled Con/Pro}
        \label{fig:lexical bias GPT3.5 sub4}
    \end{subfigure}
    \vspace{1mm}
    \begin{subfigure}[b]{0.24\linewidth}
        \includegraphics[width=\linewidth]{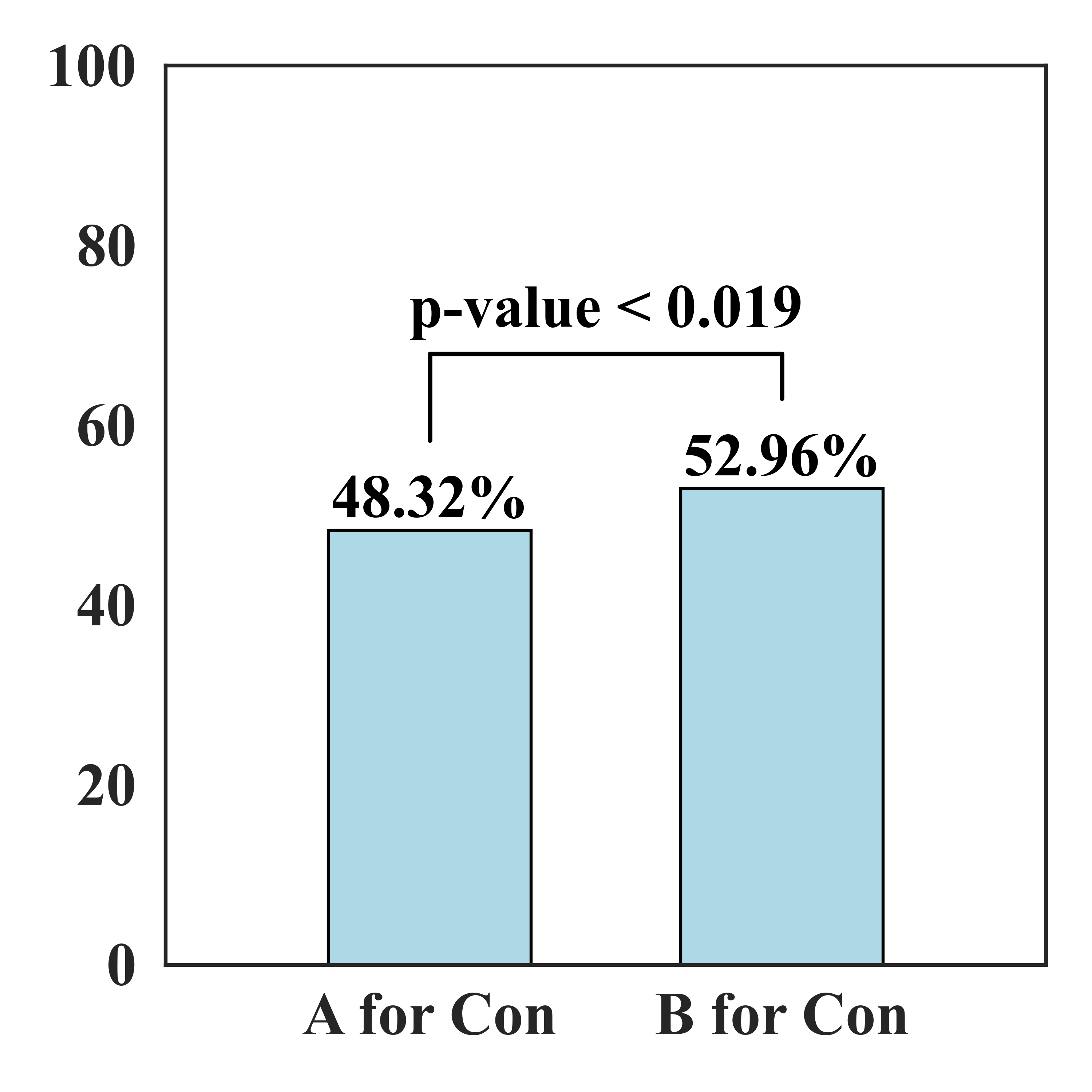}
        \caption{A/B vs. B/A}
        \label{fig:lexical bias GPT3.5 sub5}
    \end{subfigure}
    \hfill 
    \begin{subfigure}[b]{0.24\linewidth}
        \includegraphics[width=\linewidth]{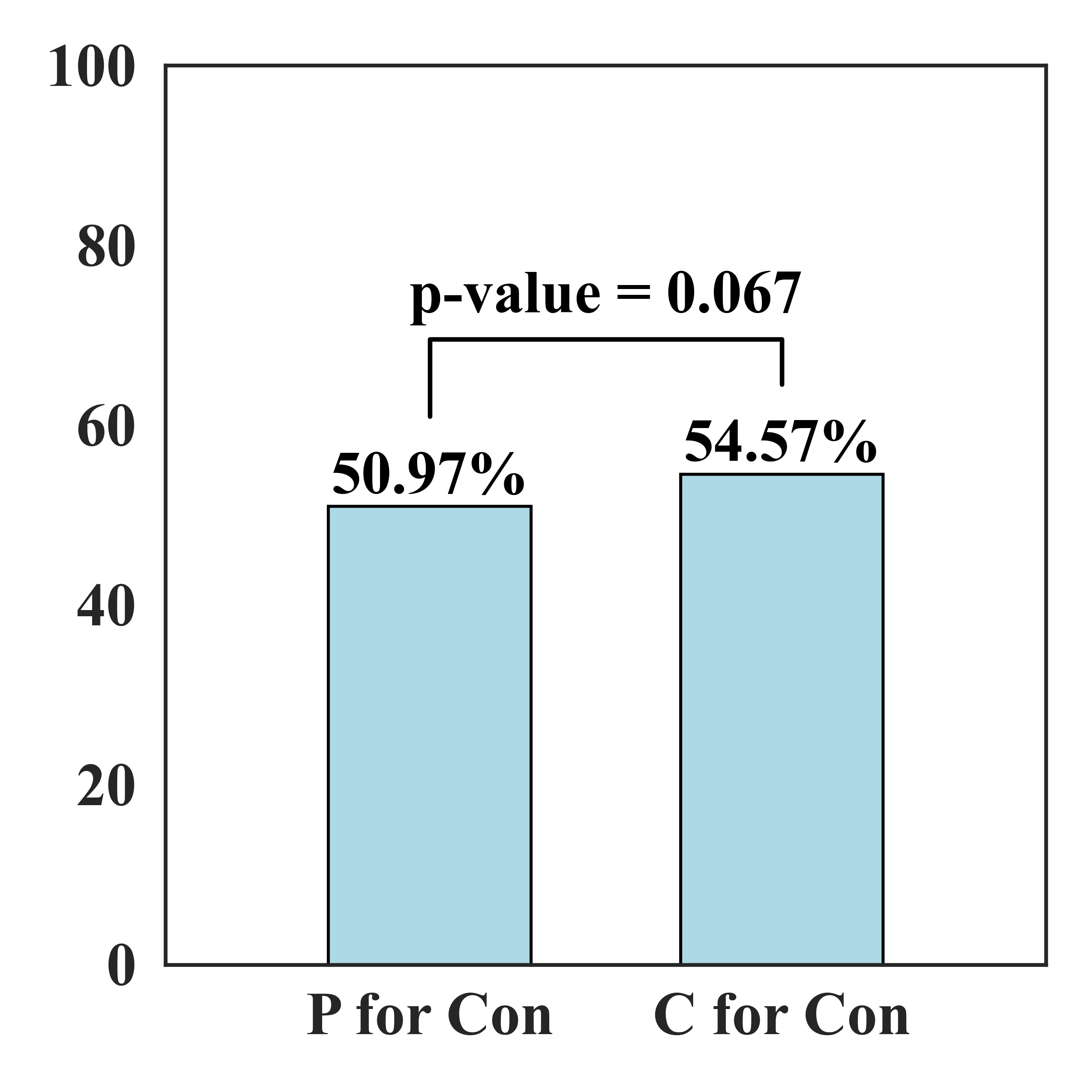}
        \caption{P/C vs. C/P}
        \label{fig:lexical bias GPT3.5 sub6}
    \end{subfigure}
    \hfill 
    \begin{subfigure}[b]{0.24\linewidth}
        \includegraphics[width=\linewidth]{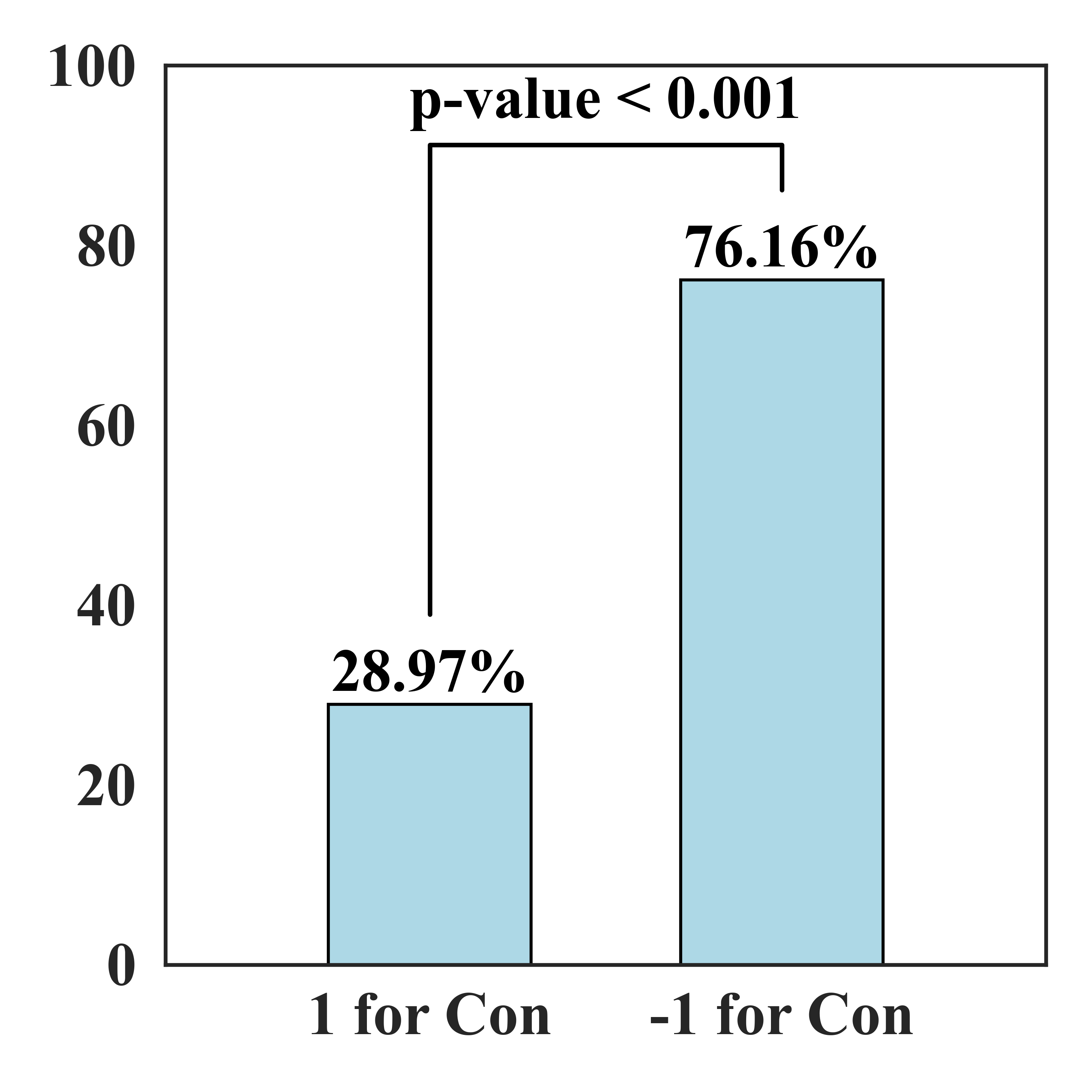}
        \caption{1/-1 vs. -1/1}
        \label{fig:lexical bias GPT3.5 sub7}
    \end{subfigure}
    \hfill 
    \begin{subfigure}[b]{0.24\linewidth}
        \includegraphics[width=\linewidth]{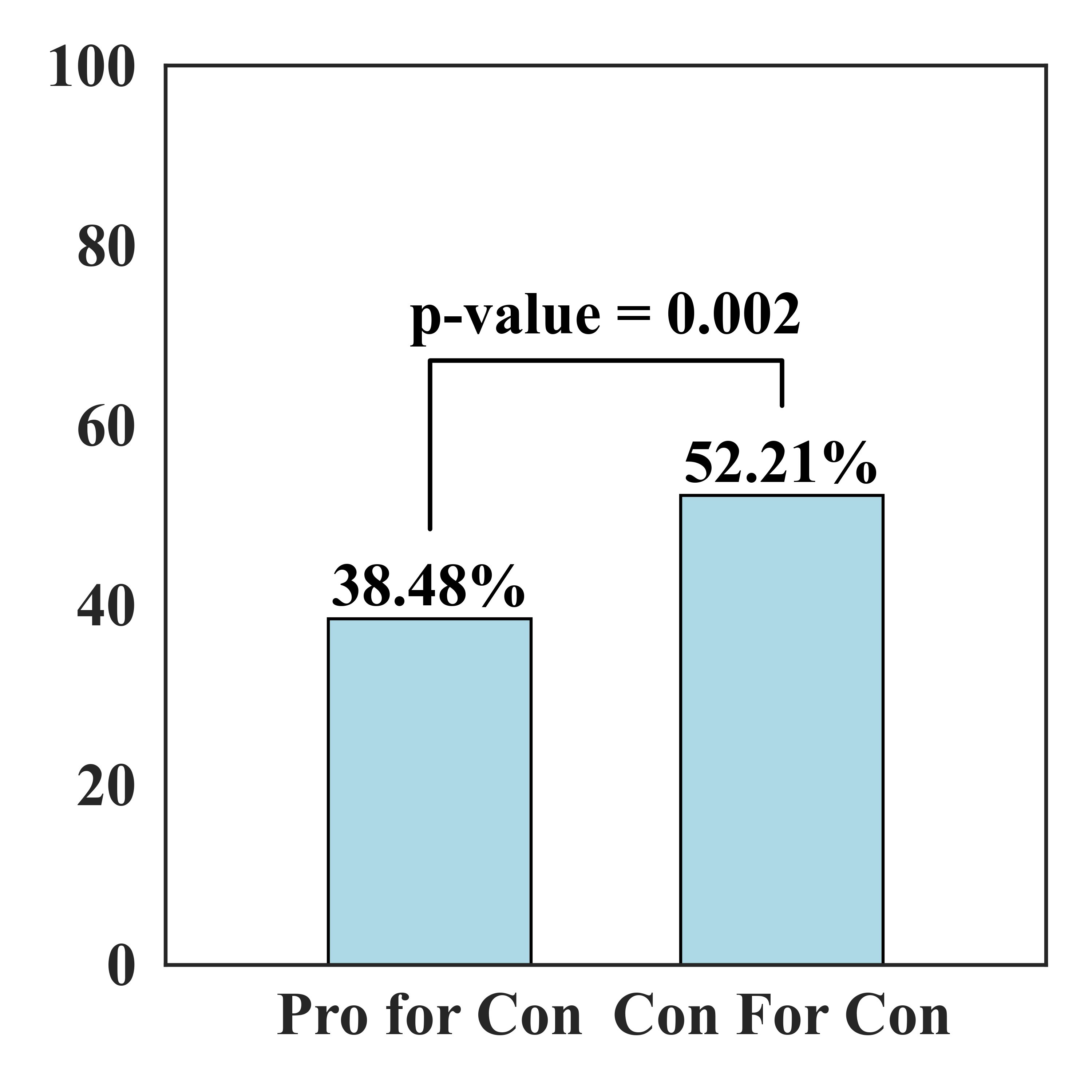}
        \caption{Pro/Con vs. Con/Pro}
        \label{fig:lexical bias GPT3.5 sub8}
    \end{subfigure}
    \vspace{-0.1cm}
    \caption{Each subfigure's legend delineates the Pro/Con label set across different verbalizer configurations. GPT-3.5 demonstrates a consistent lexical bias, which persists across shuffled positions aimed at counteracting positional bias, and in settings where Pro consistently precedes Con, except for the insignificant bias within P/C.}
    \label{fig: lexical bias in GPT-3.5}
\end{figure*}
\vspace{-0.15cm}

%% file: 03_experiments.tex
\section{Experiments}
\label{sec:experiments}

\paragraph{Dataset.}
We utilize DDO dataset \cite{durmus2019corpus}, which comprises 77,655 debates from 23 topics on debate.org, structured into rounds with a single utterance from each of the Pro and Con side. We focus on debates of 3 to 5 rounds, defining winners by audience vote differences exceeding two, and exclude debates with forfeits to maintain analysis integrity, following previous works' setting \cite{li2020exploring, hsiao2022modeling}. The length of these debates aligns well with the input length capacities of current LLMs, making it more suitable than other datasets derived from transcribed debate videos. We present experiments on an additional dataset in the Appendix \ref{sec: appendix 5}, which demonstrate consistent findings.

The dataset exhibits a win bias towards the Con side across 3 to 5-round debates (36.9\% vs. 63.1\%, 44.9\% vs. 55.1\%, 37.9\% vs. 62.1\%, respectively), likely due to a concluding side bias with Con frequently concluding debates. To evaluate LLMs in debate assessment, we propose two settings: balanced and unbalanced. The unbalanced setting replicates the original dataset's distribution, sampling 500 debates for each round count (totaling 1500). Conversely, the balanced setting aims to examine LLMs' inherent bias by ensuring equal representation of four scenarios—Pro or Con initiating and winning or losing—with 125 debates each for 3 and 4 rounds, and due to data constraints, 75 debates each for 5 rounds, resulting in 500 debates for 3 and 4 rounds and 300 for 5 rounds.


\paragraph{Evaluation metrics.}
In addition to the accuracy reported by previous works, we measure 
weighted F-1 score to accommodate the imbalance between Pro win and Con win in the original data distribution, aiming for a more comprehensive understanding.


\paragraph{Models.} For open-source model, we select LLaMA2-70B~\cite{touvron2023llama} as it has been demonstrated as the most powerful model in the LLaMA family. For close-source models, we select the latest stable versions of OpenAI's GPT-3.5 and GPT-4 models at the time to conduct our experiments, namely gpt-3.5-turbo-1106 and gpt-4-1106-preview. 

\paragraph{Human annotation.}
To assess the effectiveness of LLMs, two 
authors manually annotated the “win/lose” outcomes of randomly selected debates independently for 75 debates. 
Unlike the collective voting in multi-audience settings, this annotation was independently completed by a single annotator.


%% file: 04_results.tex
\section{Results and Analysis}
\label{sec:results}

\subsection{LLMs’ Performance}

Table \ref{tab:model_perforamance} reveals that GPT-3.5 and GPT-4 match human evaluators in assessing debates, highlighting their effectiveness. Using 75 debates labeled by two of the authors enables a direct comparison with GPT-3.5 and GPT-4. They achieve significant accuracy and F1 scores—82.04\% and 81.85\% for GPT-3.5, and 86.22\% and 86.01\% for GPT-4, respectively, outperforming previous SOTA models. LLaMA2-70B, on the other hand, performs significantly worse than existing methods, being only comparable to the ruble-based method. Thus, it is less likely for LLaMA2 to be adopted as the automatic debate evaluator. Our further experiments for bias analysis therefore mainly focus on GPT-3.5 and GPT-4.


Notably, the word choice in the prompt can have a profound impact on the performance of LLMs, as shown in Table \ref{tab:model_settings}. Within our study, employing the label set 1/-1 results in a marked decline in the performance of GPT-3.5, and using the label set Pro/Con leads to the lowest observed outcomes in GPT-4. GPT-3.5 is particularly sensitive to negative phrasing; its performance degrades below that of random selection when prompted to identify the debate's loser rather than the winner. In contrast, GPT-4 demonstrates much less sensitivity to such changes, showing only a minor decrease in performance.

\subsection{Biases Analysis}
Our study explores biases present in GPT-3.5 using a balanced setting of DDO dataset. Additional analyses of the GPT-3.5 on the 
original unbalanced DDO data and analysis of GPT-4 are in Appendix \ref{sec: appendix 2} and \ref{sec: appendix 1}, respectively. The experiments with an extra dataset that confirm our findings are presented in Appendix \ref{sec: appendix 5}.


\begin{table}[th]
\centering
\small
\begin{tabular}{lccc}
\toprule
\textbf{Evaluators} & \textbf{Size} & \textbf{Acc} & \textbf{F1} \\ 
\midrule
Rule-based & 6058 & 67.53 & 46.68\\
\midrule
LLaMA2-70B  & 1500 & 65.69 & 56.07 \\
\midrule
BERT + Structure & - & 78.89 & -- \\
BERT + Relation & 1964 & 80.04 & -- \\
\midrule
GPT-3.5 & 1500 & 82.04 & 81.85 \\
GPT-4 & 1500 & 86.22 & 86.01 \\
\midrule
Human 1 & 75 & 77.33 & 77.39 \\
Human 2 & 75 & 78.67 & 78.15 \\
\bottomrule
\end{tabular}
\caption{GPT-3.5 and GPT-4's performance are on par with human performance and outperform the existing state-of-the-art BERT-based methods with fine-tuning \cite{li2020exploring, hsiao2022modeling}. The rule-based model predicts the winner as the side that concludes the debate. LLaMA2-70B has similar performance to the rule-based model.}
\label{tab:model_perforamance}
\end{table}

\begin{table}[th]
\centering
\small
\begin{tabular}{cccccc}
\toprule
\textbf{Evaluators} & \textbf{Verbalizer} & \textbf{Outcome} & \textbf{Acc} & \textbf{F1} \\ 
\midrule
\multirow{5}{*}{GPT-3.5} & A/B & Winner & \textbf{82.04} & \textbf{81.85} \\
 & P/C & Winner & 81.39 & 81.02 \\
 & 1/-1 & Winner & \textit{72.08} & \textit{68.24} \\
 & Pro/Con & Winner & 81.86 & 81.60 \\
 & A/B & Loser & 37.72 & 24.74 \\
\midrule
\multirow{5}{*}{GPT-4} & A/B & Winner & 84.49 & 84.49 \\
 & P/C & Winner & 85.11 & 84.78 \\
 & 1/-1 & Winner & \textbf{86.22} & \textbf{86.01} \\
 & Pro/Con & Winner & \textit{79.72} & \textit{78.16} \\
 & A/B & Loser & 80.94 & 81.11 \\
\bottomrule
\end{tabular}%

\caption{The ``Verbalizer" column lists Pro\_label and Con\_label sets, and the ``Outcome" column shows whether GPT-3.5 and GPT-4 are tasked with identifying debate winners or losers. Bold formatting indicates the top-performing verbalizer choice, while italics highlight the least effective choice.}
\label{tab:model_settings}
\end{table}

\paragraph{Positional Bias.}

Figure \ref{fig: positional bias in GPT-3.5} compares the proportion of predictions labeled as "Con" between instances where the Con is positioned at the first candidate response and the instances where Con is placed as the second candidate response. It shows that GPT-3.5 systematically favors the candidate response in the second position across all tested verbalizer settings. The two-sided P-values of the two-proportion z-test consistently suggest the positional bias is significant. This finding confirms the second position preference of GPT-3.5 as reported by \citet{wang2023large}. On the unbalanced data that reflects the original distribution, we also investigate the changes in the counts of predicted Pros and predicted Cons between the settings with shuffled candidate response positions and fixed positions. The details are shown in Appendix \ref{sec: appendix 1}, and \ref{sec: appendix 2}, suggesting a consistent trend.

\paragraph{Lexical Bias.}

GPT-3.5 is affected by the lexical choice of labels representing the two sides of a debate, as demonstrated by Figure \ref{fig: lexical bias in GPT-3.5}. These differences highlight the inherent lexical bias of GPT-3.5 within the selected label set. GPT-3.5 prefers the label `B'(`-1') over `A'(`1'), predicting Con as the winner significantly more frequently when `B'(`-1') represents Con as opposed to when `A'(`1') does, as shown in Figure \ref{fig: lexical bias in GPT-3.5}. There is no significant lexical bias found within the P/C label set for GPT-3.5. The Con/Pro label configuration, which swaps the position names of the two sides, could confuse LLMs about each label's corresponding side, as the content of the debate usually reveals the actual position of each side. This ambiguity might contribute to the poorer performance observed in the Con/Pro label setting and raises questions about the inferred preference for the 'Con' label. The analysis of lexical bias is further detailed in Appendix~\ref{sec: appendix 1} and \ref{sec: appendix 2}.

\paragraph{Order Bias.} GPT-3.5 exhibits a significant order bias, favoring the side that concludes the debate, as shown in experiments where the Pro\_label consistently ranked as the primary response (Table \ref{tab:chi_test_for_order_bias_in_GPT35}). This bias is statistically significant across all verbalizer options. The results suggest an inherent tendency in LLMs to give more weight to the final arguments.

\begin{table}[ht]
\centering
\footnotesize 
\scalebox{0.95}{
\begin{tabular}{ccccr}
\toprule
\textbf{Verbalizer} & \textbf{End-Side} & \textbf{\# P-Pro} & \textbf{\# P-Con} & \textbf{P-Value} \\
\midrule
 \multirow{2}{*}{A/B} & Pro & {\bf 389} & 253 & < 0.001 \\
    & Con & 215 & {\bf 427} &  \\
\midrule
\multirow{2}{*}{P/C} & Pro & {\bf 408} & 245 & < 0.001 \\
    & Con & 184 & {\bf 460} &  \\
\midrule
\multirow{2}{*}{1/-1} & Pro & 238 & {\bf 409*} & < 0.001 \\
             & Con &  70 & {\bf 575} &  \\
\midrule
\multirow{2}{*}{Pro/Con} & Pro & {\bf 399} & 218 & < 0.001 \\
        & Con & 248 & {\bf 426} &  \\
\bottomrule
\end{tabular}}
\caption{Analysis of GPT-3.5 predictions correlating with debate orders, using Chi-square tests for significance. "\# P-Pro" and "\# P-Con" indicate the counts of Pro and Con sides predicted as winners, respectively. The results reveal a significant association with order for all verbalizer choices. * here highlights the strong lexical bias for `-1' that dominates the others.}
\label{tab:chi_test_for_order_bias_in_GPT35}
\end{table}



%% file: 05_conclusion.tex
\section{Discussion}

Our research demonstrates that LLMs outperform current SOTA models in evaluating debates but are influenced by specific word choices, affecting their efficacy. We highlight LLMs' embedded biases—positional, lexical, and order—offering insights for future LLM training enhancements. 

Despite attempts to neutralize positional bias by shuffling labels in Figs \ref{fig:lexical bias GPT3.5 sub1} and \ref{fig:lexical bias GPT3.5 sub4}, GPT-3.5 still exhibits a Pro bias, contradicting its lexical preference for 'B'('Con'). This might suggest a confirmation bias-like tendency in GPT-3.5, favoring agreement with the debate topic. We further conduct experiments shuffling A/B with B/A and 1/-1 with -1/1 label sets, where each label randomly represents Pro or Con in 50\% of cases, with positions also shuffled. Despite eliminating lexical and positional biases, results indicate a persistent Pro bias, detailed in Appendix Figure \ref{fig:stance bias}, pointing to an underlying tendency warranting further investigation.


\section{Limitations}
The insights from our investigation, based on the examination of GPT-3.5 and GPT-4, indicate that the discerned behavioral patterns might be unique to these specific models and not necessarily extend to other language models with divergent architectures or training approaches. With the relentless advancement in language model technology and the anticipation of updated versions, the biases detected in GPT-3.5 and GPT-4 could become obsolete in subsequent iterations. Highlighting the significance of prompt types and training techniques on the efficacy of models, our research underlines the imperative for continued research to identify the optimal prompt types for various scenarios and the optimal training methods for reducing bias.

Although various biases may interact and potentially counterbalance each other, leading to improvements, the intensity of distinct bias types can vary significantly across different contexts. Consequently, a prompt that appears to exhibit balanced bias in one scenario may manifest more pronounced bias under slightly altered conditions.

%% file: 06_appendix.tex
\onecolumn
\section{Appendix}
\label{sec:appendix}

\subsection{More details of the Dataset}

The dataset extends beyond textual debate content to audience votes across four evaluation criteria: making more convincing arguments, better conduct, use of reliable sources, and spelling and grammar proficiency. Consistent with prior research \cite{li2020exploring, hsiao2022modeling}, our analysis utilizes the criterion of "making more convincing arguments" for assessing debate outcomes. To ensure alignment with these studies and enhance comparability, we narrow our focus to debates with a definitive margin of victory—requiring a vote difference exceeding two—and limit our analysis to debates spanning three to five rounds, which represent the bulk of the dataset. Debates compromised by forfeits, identified either through explicit forfeit labels or instances of one side forfeiting a round, are omitted from consideration.

The debates within the dataset have an average length of 1574.93 words, with the majority fit within the input length constraints of contemporary LLMs. Regarding audience engagement, the average vote counts for 3-round, 4-round, and 5-round debates stand at 10.05, 7.02, and 7.03, respectively. Furthermore, the average vote differences for these debate formats are 5.52, 4.69, and 4.79, indicating a clear preference in outcomes that facilitate our focused analysis on convincing arguments. The percentages of Con conclude the debates are 77.84\%, 78.24\%, and 78.13\% for 3-round, 4-round and 5-round debates respectively.

\subsection{Additional Results of DDO Dataset in the Balanced Setting}
\label{sec: appendix 1}

\begin{figure*}[t]
    \centering
    
    \captionsetup[subfigure]{font=scriptsize}

    \newlength{\subfigwidth}
    \setlength{\subfigwidth}{0.245\linewidth}

    \begin{subfigure}[b]{\subfigwidth}
        \includegraphics[width=\linewidth]{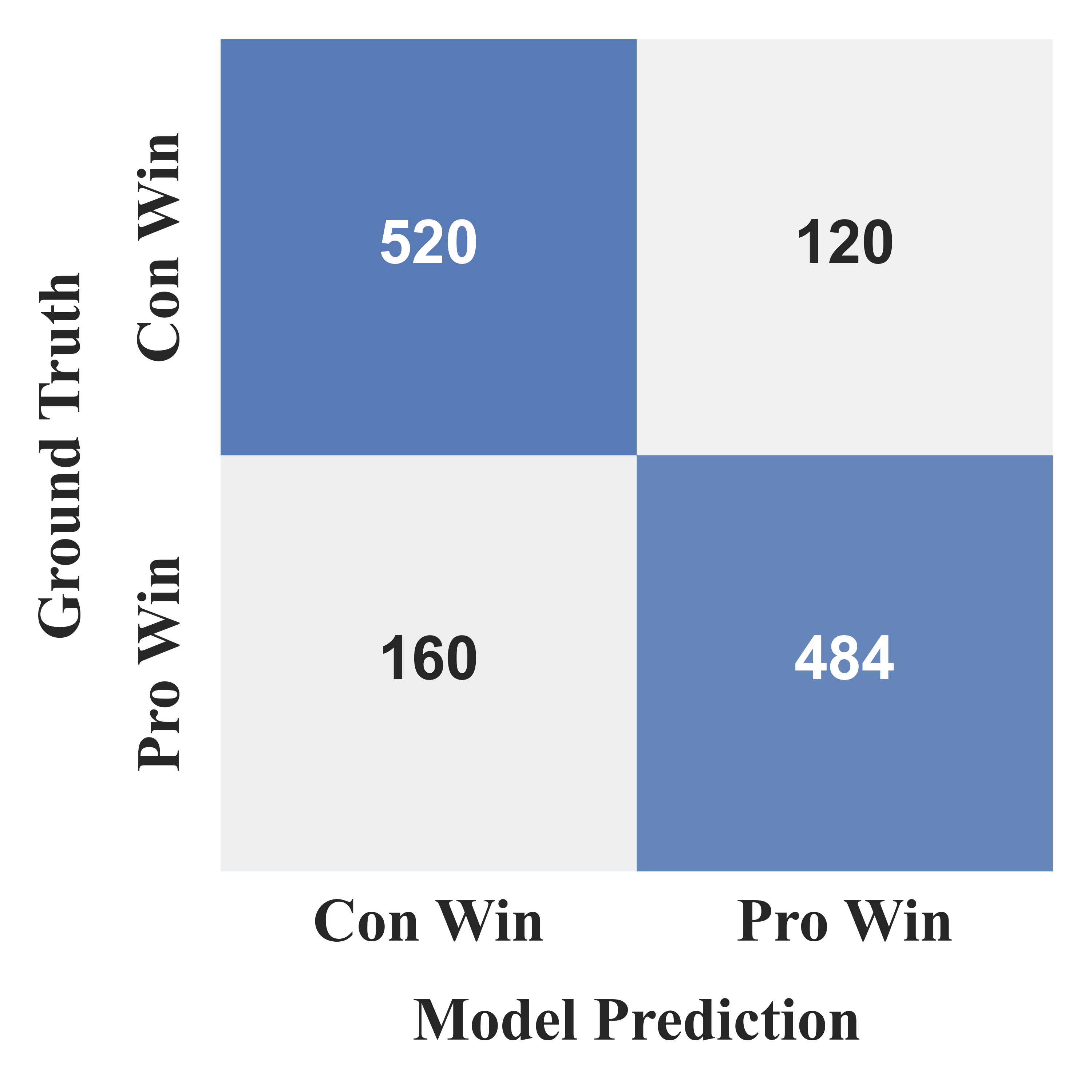}
        \caption{A/B label set}
        \label{fig:confusion matrices GPT-3.5 sub1}
    \end{subfigure}
    \hfill
    \begin{subfigure}[b]{\subfigwidth}
        \includegraphics[width=\linewidth]{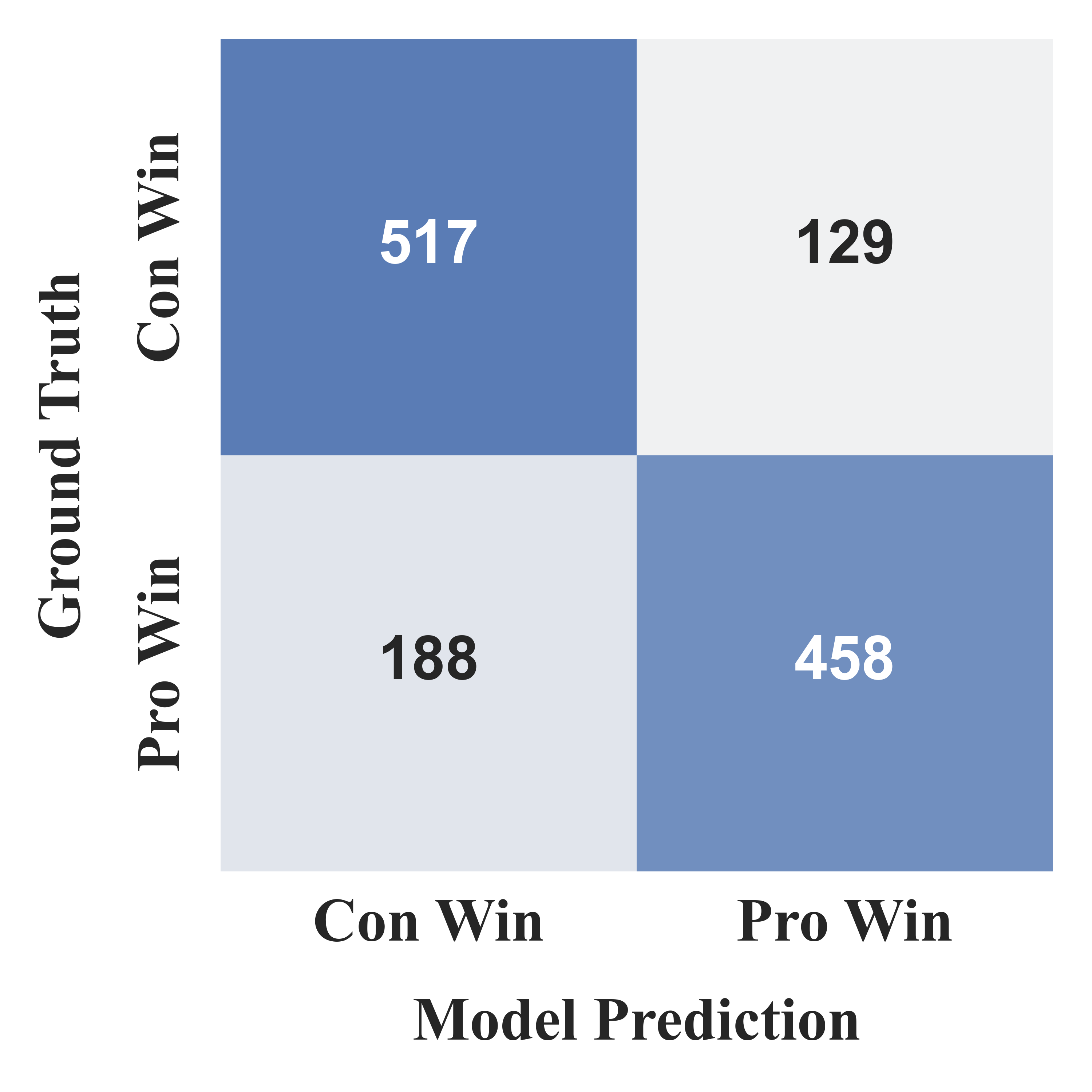}
        \caption{P/C label set}
        \label{fig:confusion matrices GPT-3.5 sub2}
    \end{subfigure}
    \hfill
    \begin{subfigure}[b]{\subfigwidth}
        \includegraphics[width=\linewidth]{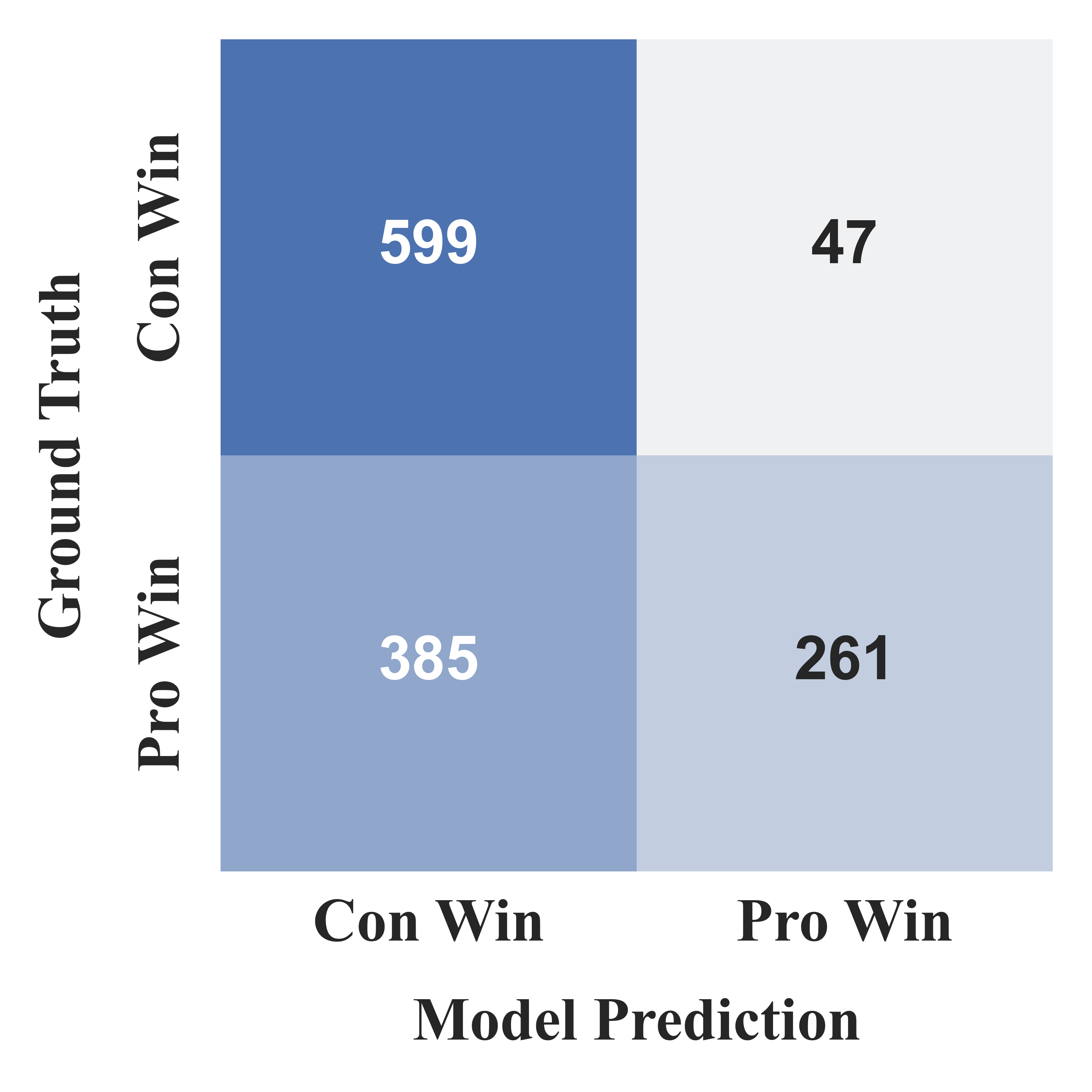}
        \caption{1/-1 label set}
        \label{fig:confusion matrices GPT-3.5 sub3}
    \end{subfigure}
    \hfill
    \begin{subfigure}[b]{\subfigwidth}
        \includegraphics[width=\linewidth]{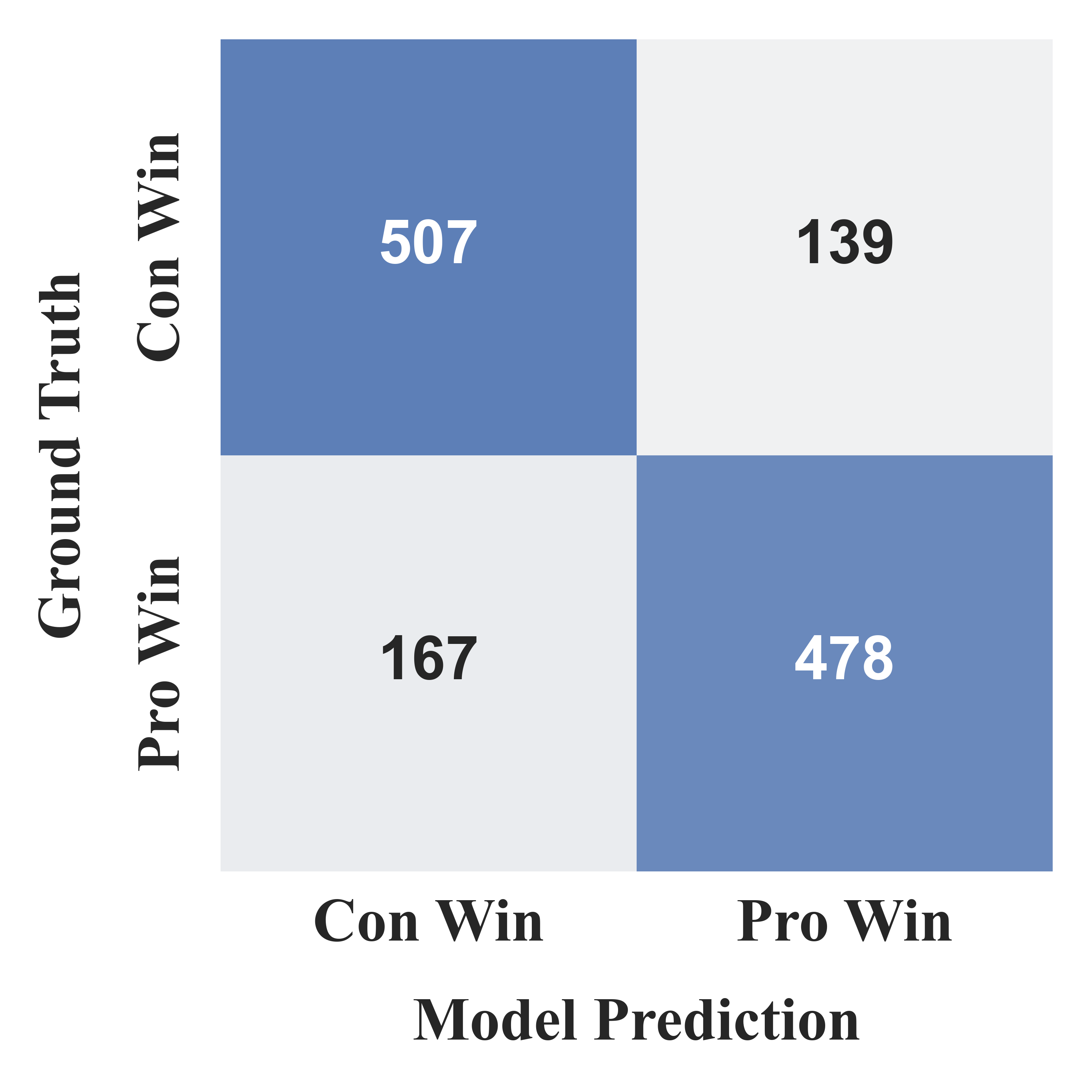}
        \caption{Pro/Con label set}
        \label{fig:confusion matrices GPT-3.5 sub4}
    \end{subfigure}

    \vspace{1mm} 



    \begin{subfigure}[b]{\subfigwidth}
        \includegraphics[width=\linewidth]{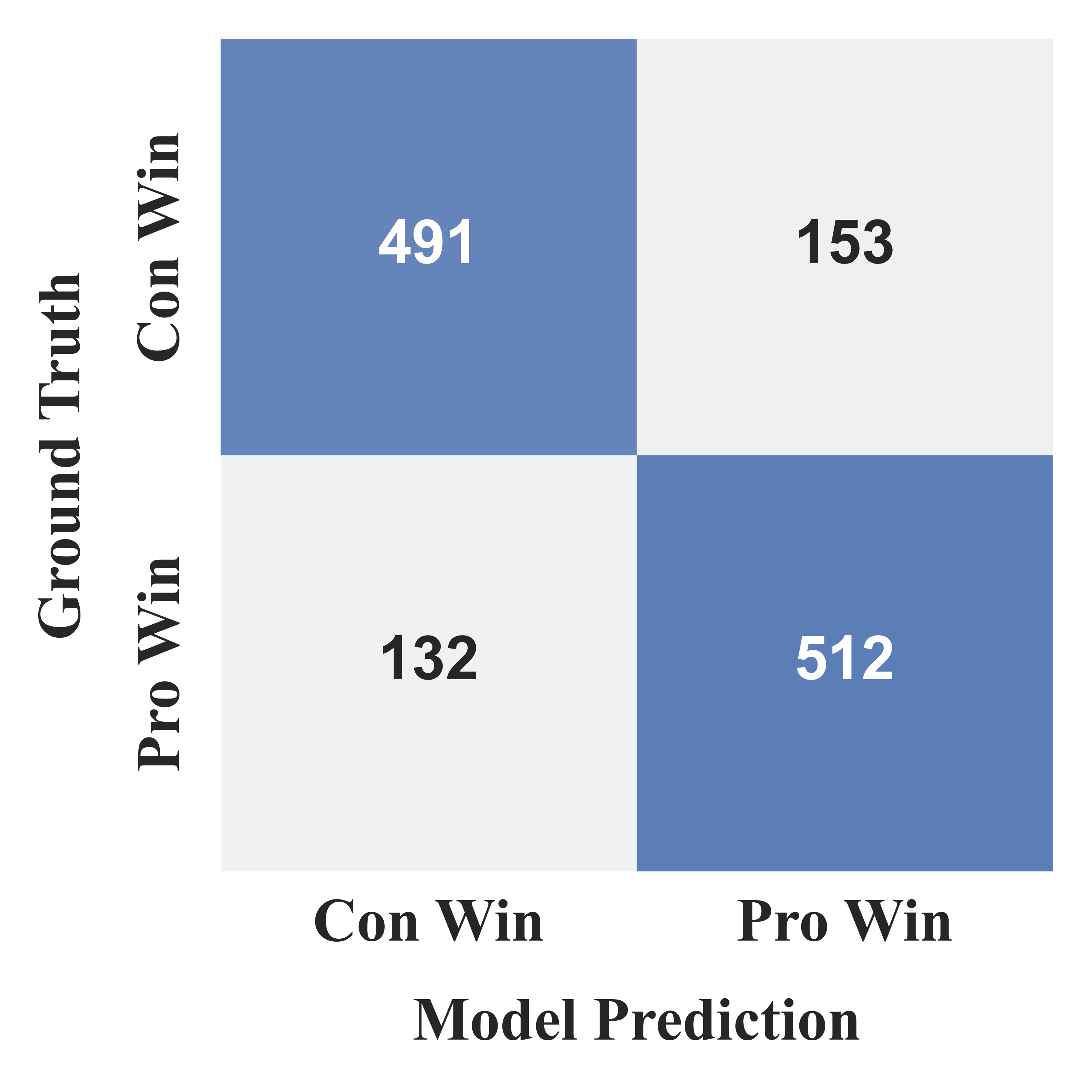}
        \caption{Shuffled A/B label set}
        \label{fig:confusion matrices GPT-3.5 sub9}
    \end{subfigure}
    \hfill
    \begin{subfigure}[b]{\subfigwidth}
        \includegraphics[width=\linewidth]{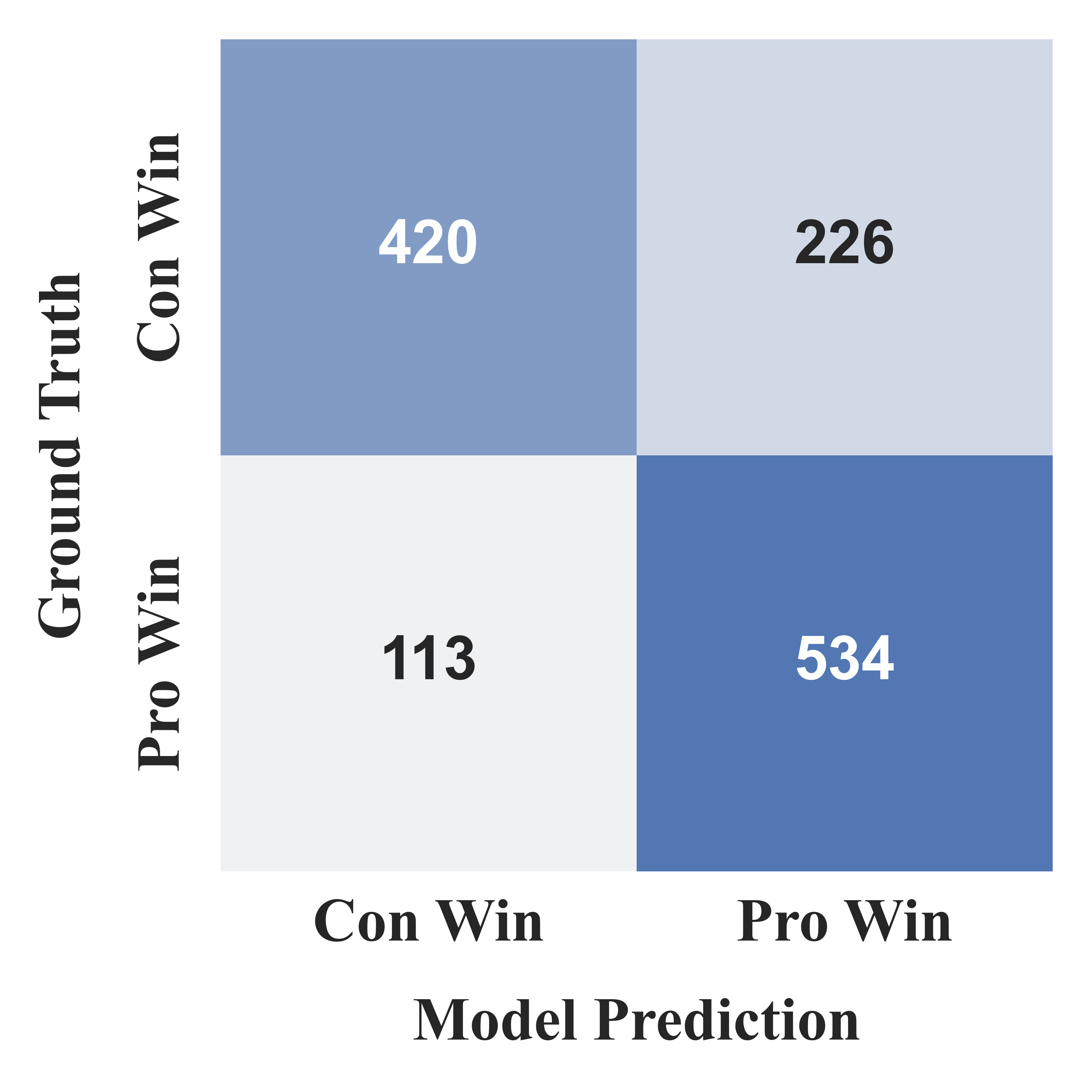}
        \caption{Shuffled P/C label set}
        \label{fig:confusion matrices GPT-3.5 sub10}
    \end{subfigure}
    \hfill
    \begin{subfigure}[b]{\subfigwidth}
        \includegraphics[width=\linewidth]{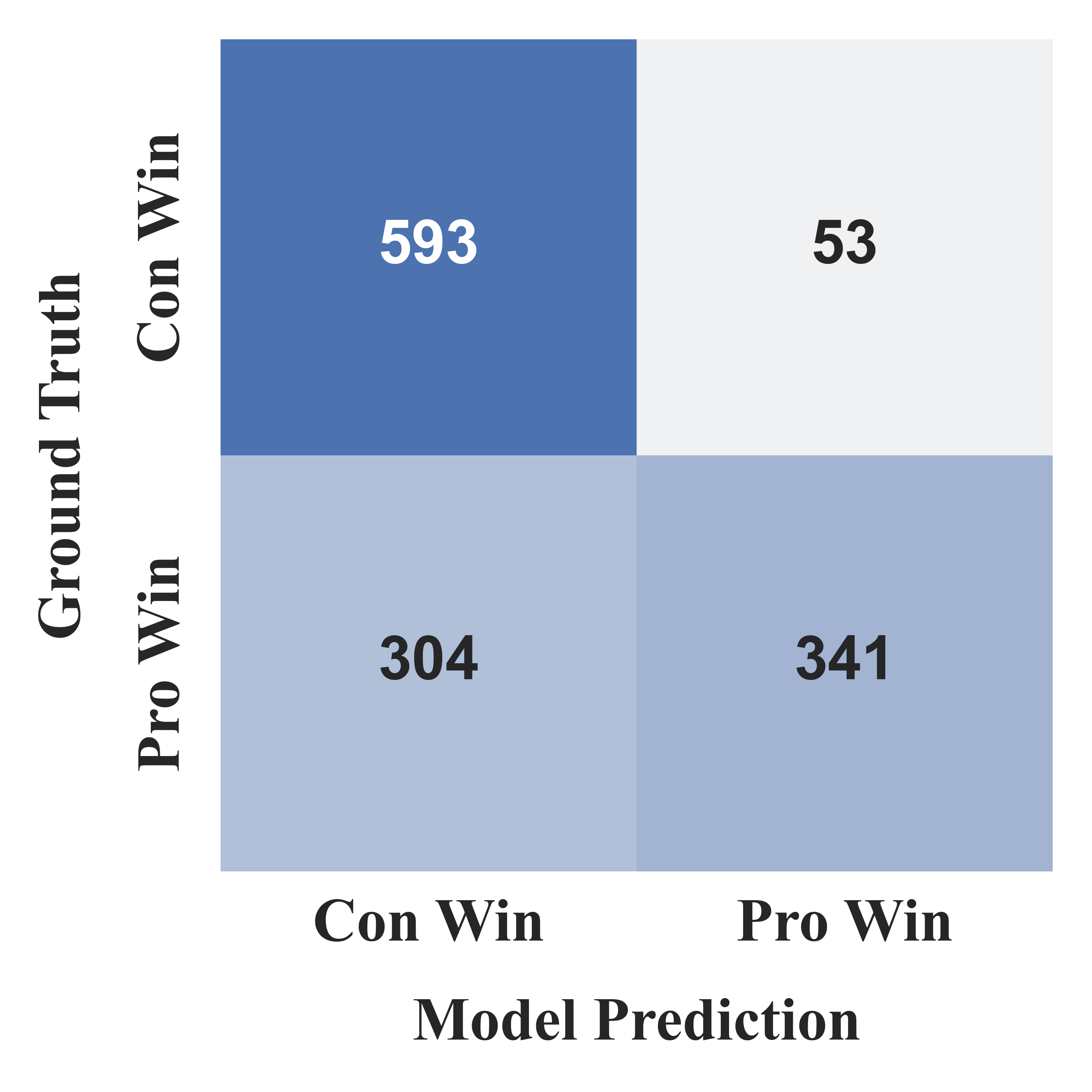}
        \caption{Shuffled 1/-1 label set}
        \label{fig:confusion matrices GPT-3.5 sub11}
    \end{subfigure}
    \hfill
    \begin{subfigure}[b]{\subfigwidth}
        \includegraphics[width=\linewidth]{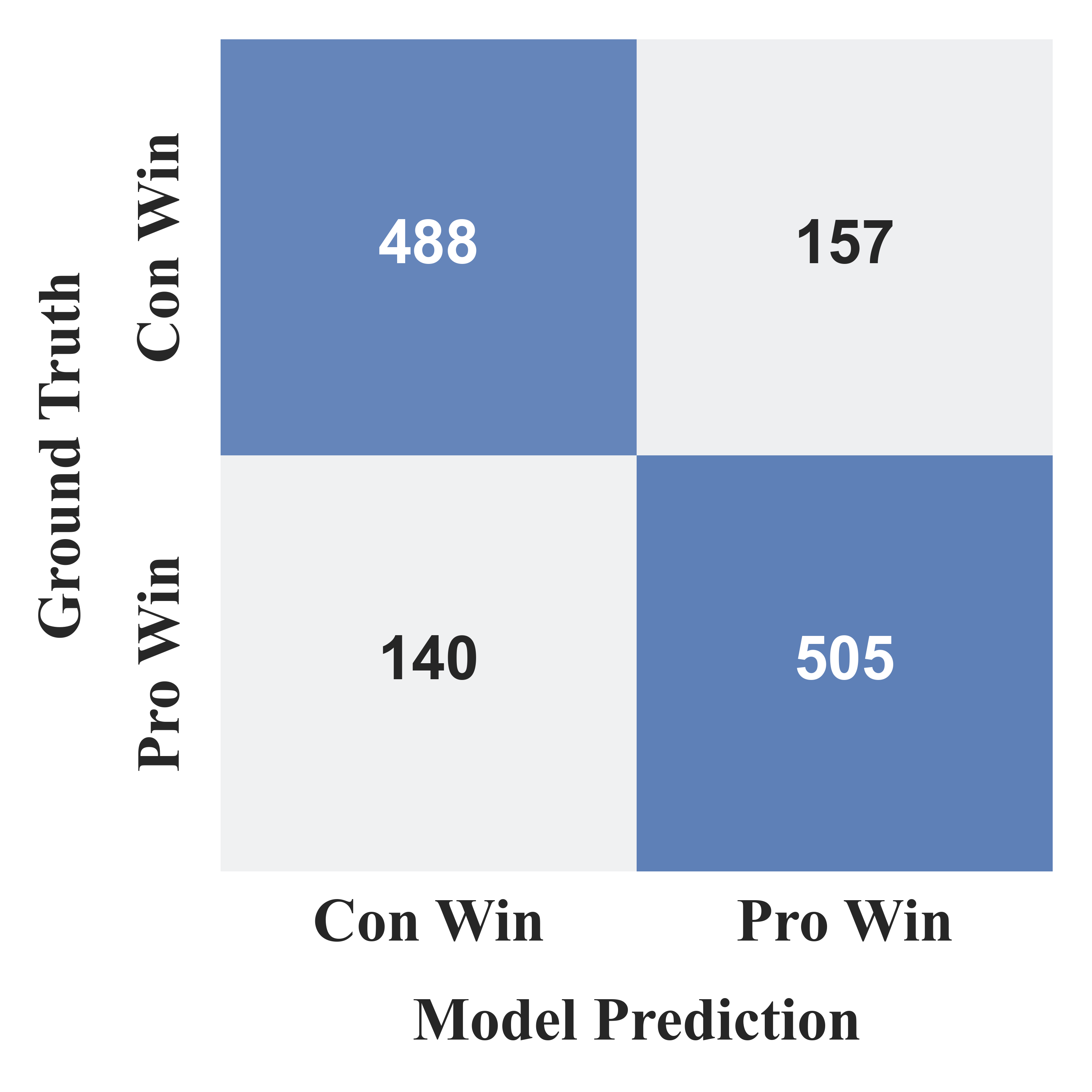}
        \caption{Shuffled Pro/Con label set}
        \label{fig:confusion matrices GPT-3.5 sub12}
    \end{subfigure}

    \vspace{1mm} 

    \begin{subfigure}[b]{\subfigwidth}
        \includegraphics[width=\linewidth]{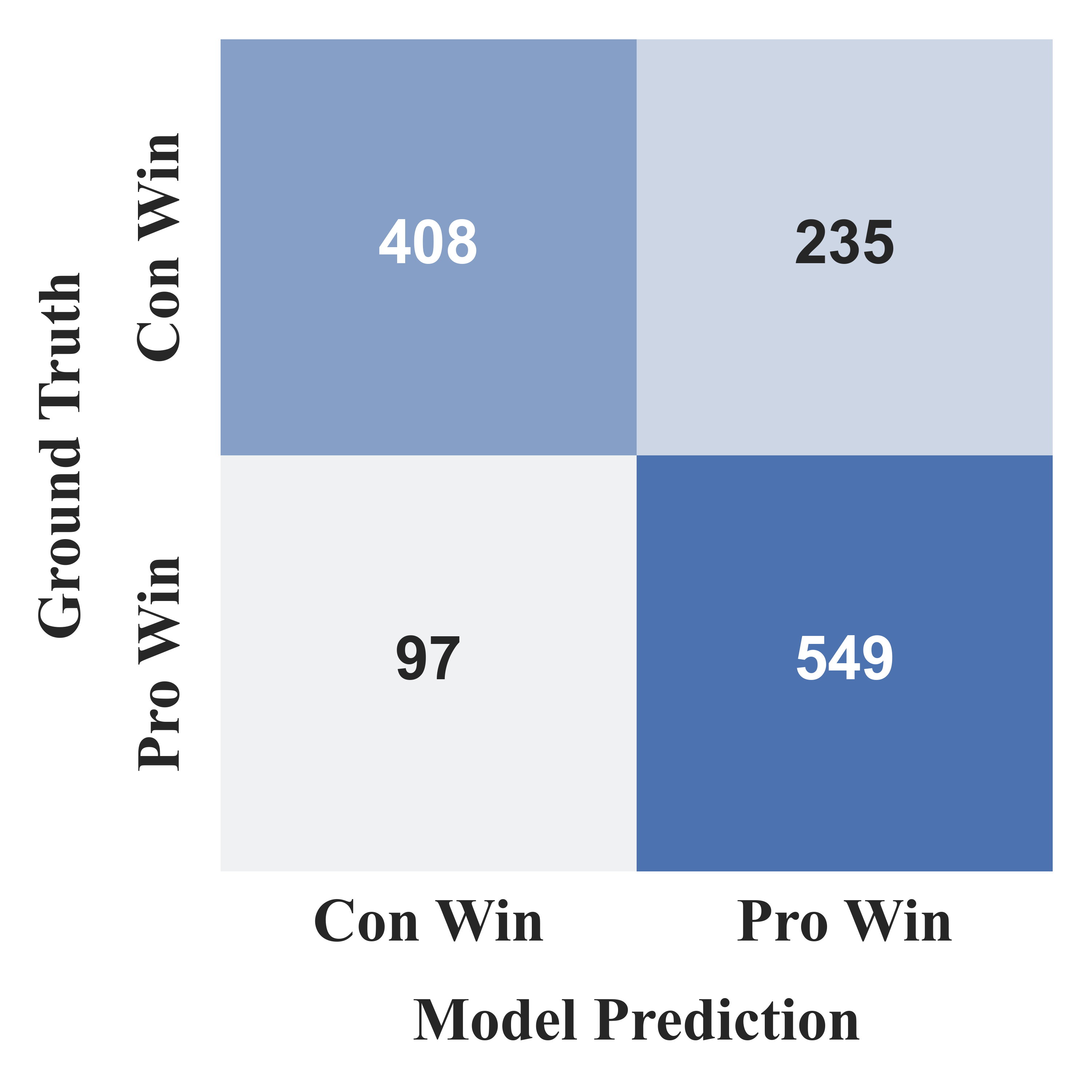}
        \caption{Shuffled B/A label set}
        \label{fig:confusion matrices GPT-3.5 sub13}
    \end{subfigure}
    \hfill
    \begin{subfigure}[b]{\subfigwidth}
        \includegraphics[width=\linewidth]{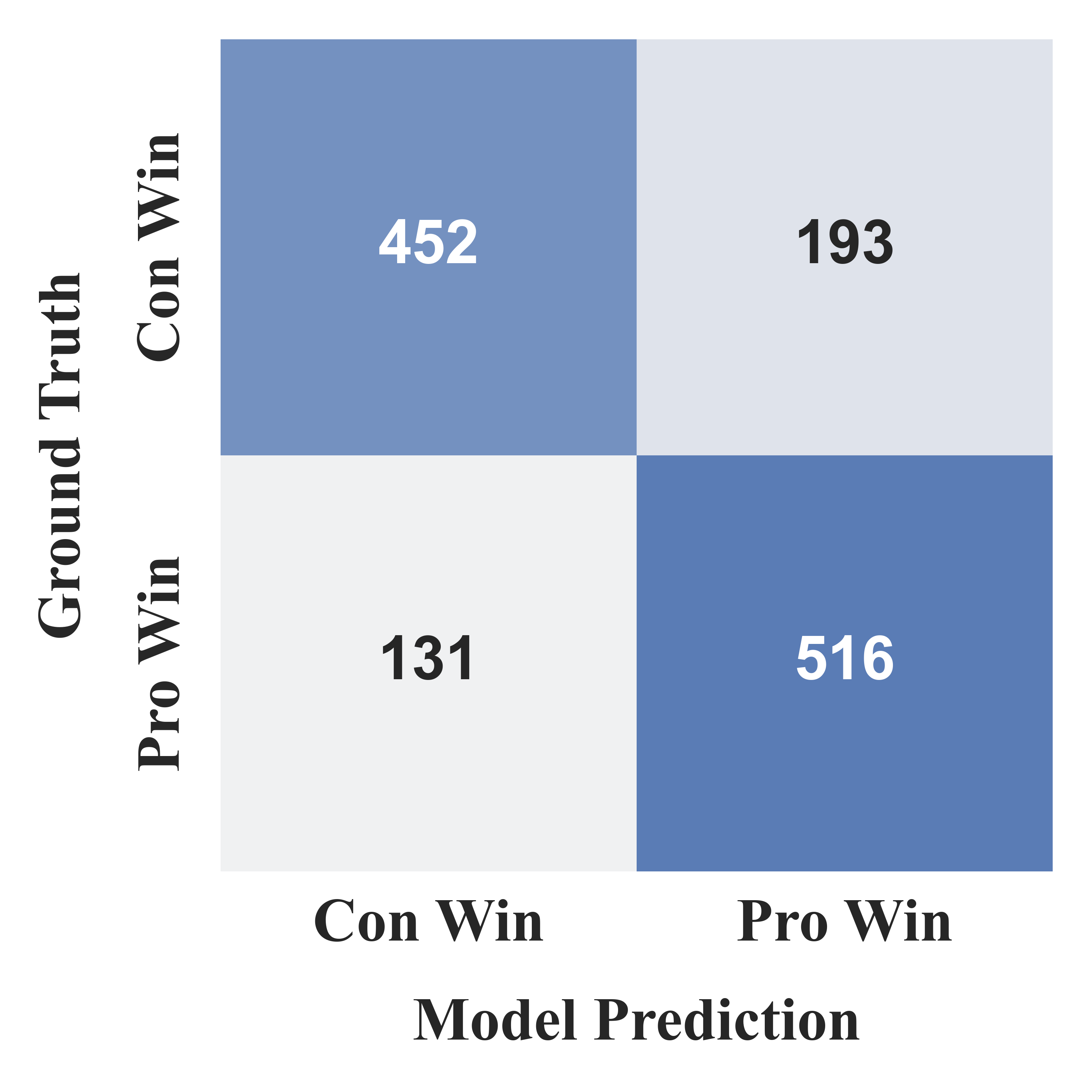}
        \caption{Shuffled C/P label set}
        \label{fig:confusion matrices GPT-3.5 sub14}
    \end{subfigure}
    \hfill
    \begin{subfigure}[b]{\subfigwidth}
        \includegraphics[width=\linewidth]{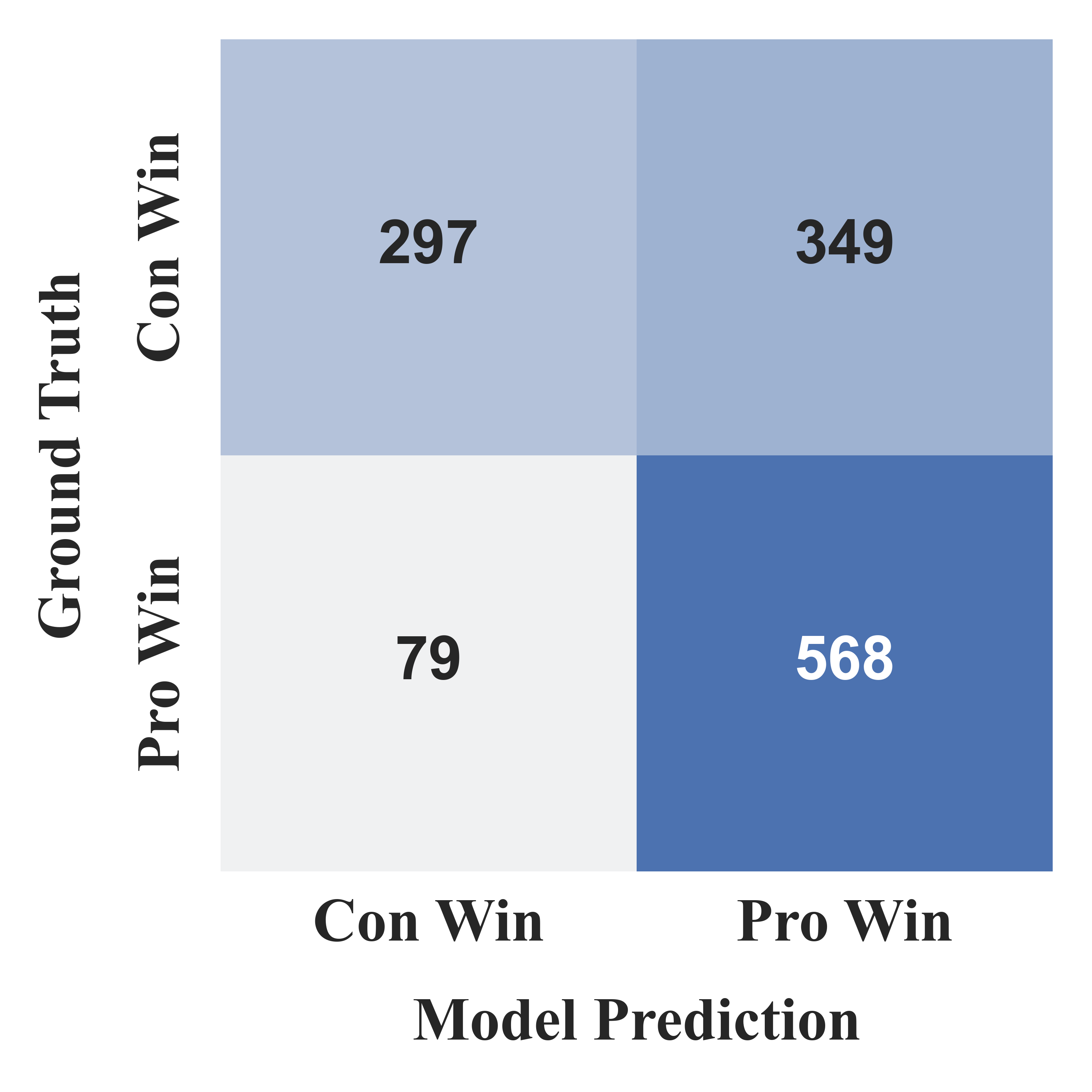}
        \caption{Shuffled -1/1 label set}
        \label{fig:confusion matrices GPT-3.5 sub15}
    \end{subfigure}
    \hfill
    \begin{subfigure}[b]{\subfigwidth}
        \includegraphics[width=\linewidth]{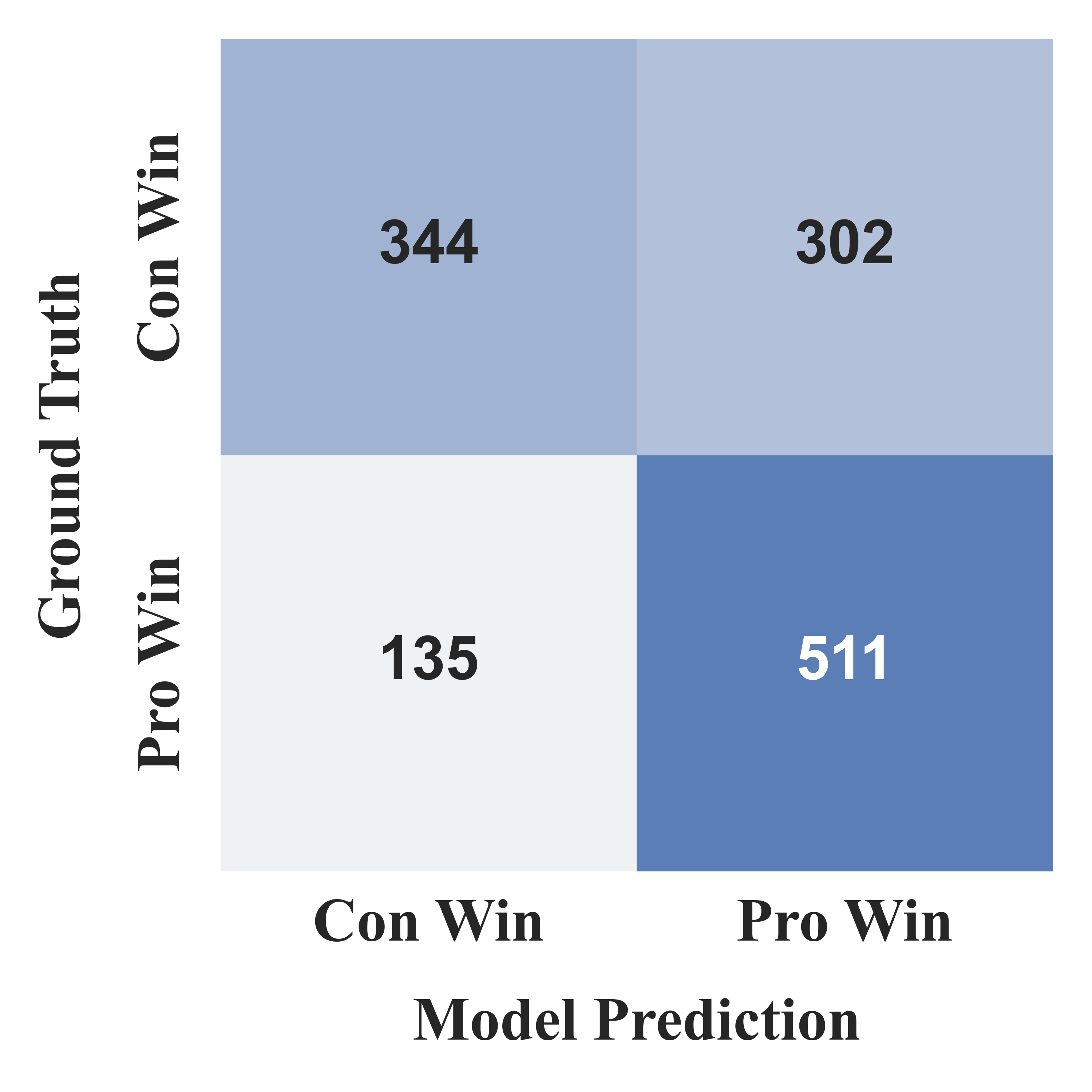}
        \caption{Shuffled Con/Pro label set}
        \label{fig:confusion matrices GPT-3.5 sub16}
    \end{subfigure}
    
    \caption{This figure displays confusion matrices for GPT-3.5 with various Pro\_label/Con\_label sets. The matrices in the first row correspond to scenarios where the Pro\_label consistently occupies the leading position in the instruction prompt, potentially introducing a positional bias. In contrast, the second and third rows present matrices from experiments where the positions of Pro\_label and Con\_label are shuffled, aiming to mitigate this bias for pure comparisons between switching corresponding label verbalizers of Pro and Con.}
    \label{fig:confusion matrices GPT-3.5}
    \vspace{-6mm}
\end{figure*}

\begin{figure*}[t]
    \centering
    
    \captionsetup[subfigure]{font=scriptsize}

    \begin{subfigure}[b]{0.245\linewidth}
        \includegraphics[width=\linewidth]{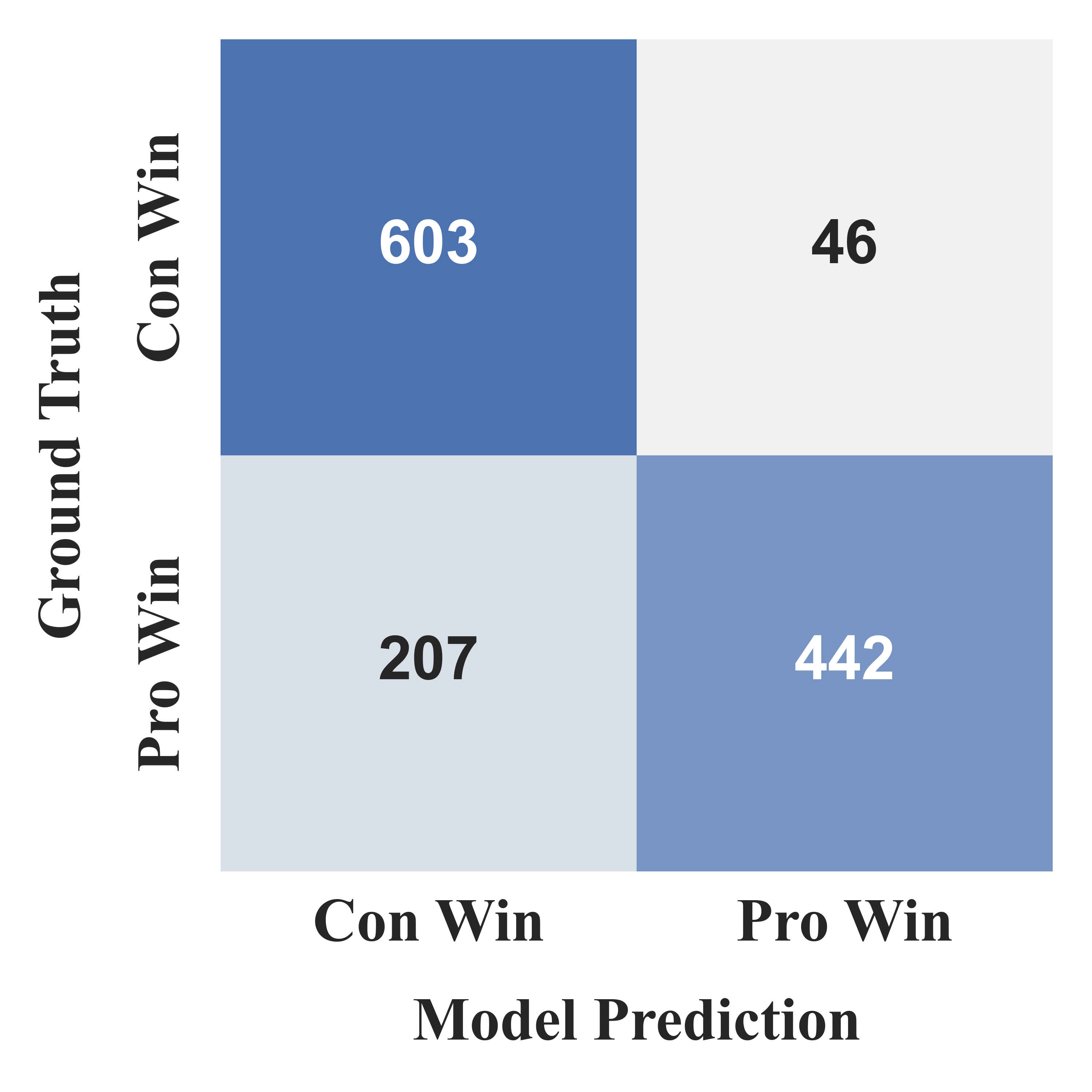}
        \caption{A/B label set}
        \label{fig:GPT4 sub1}
    \end{subfigure}
    \hfill
    \begin{subfigure}[b]{0.245\linewidth}
        \includegraphics[width=\linewidth]{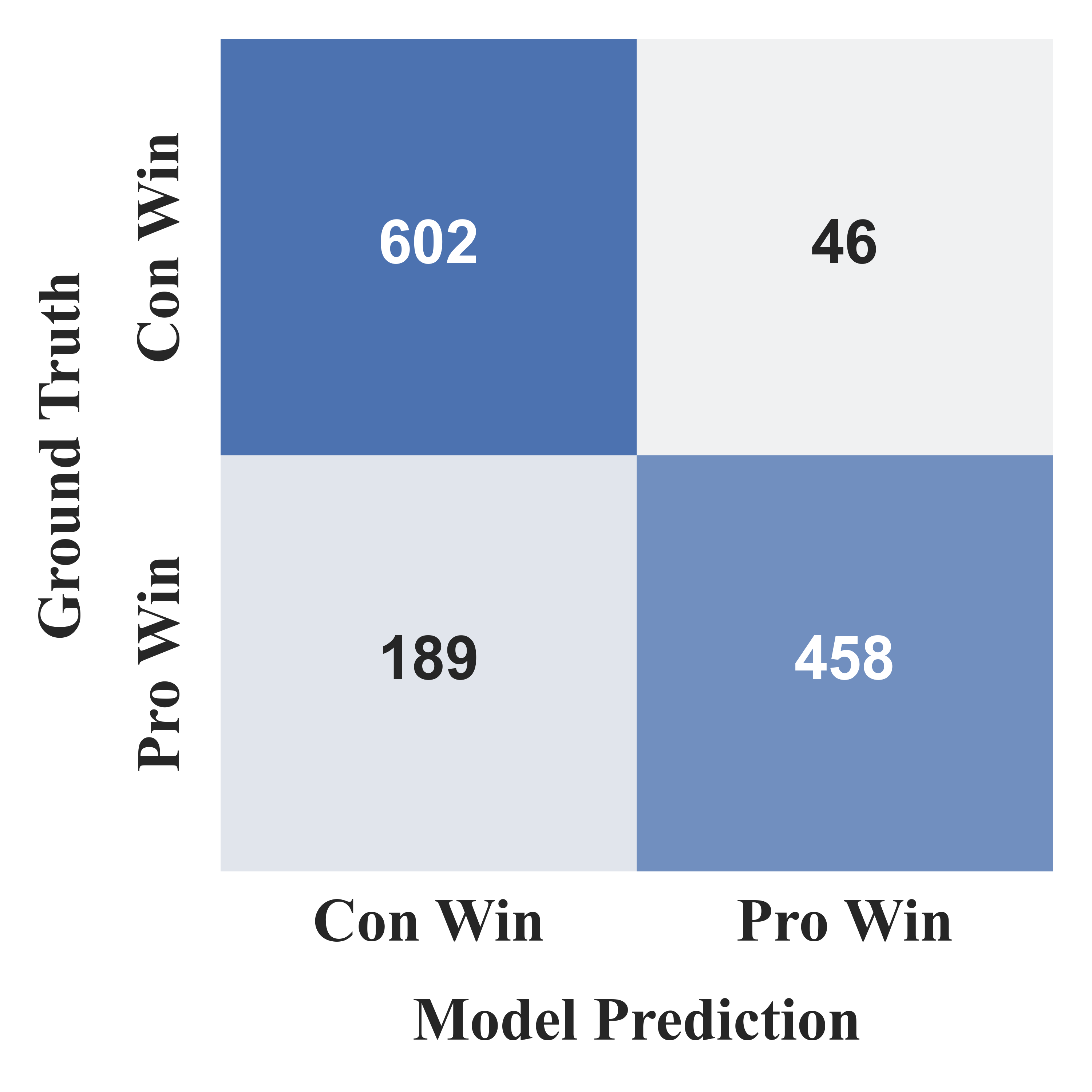}
        \caption{P/C label set}
        \label{fig:GPT4 sub2}
    \end{subfigure}
    \hfill
    \begin{subfigure}[b]{0.245\linewidth}
        \includegraphics[width=\linewidth]{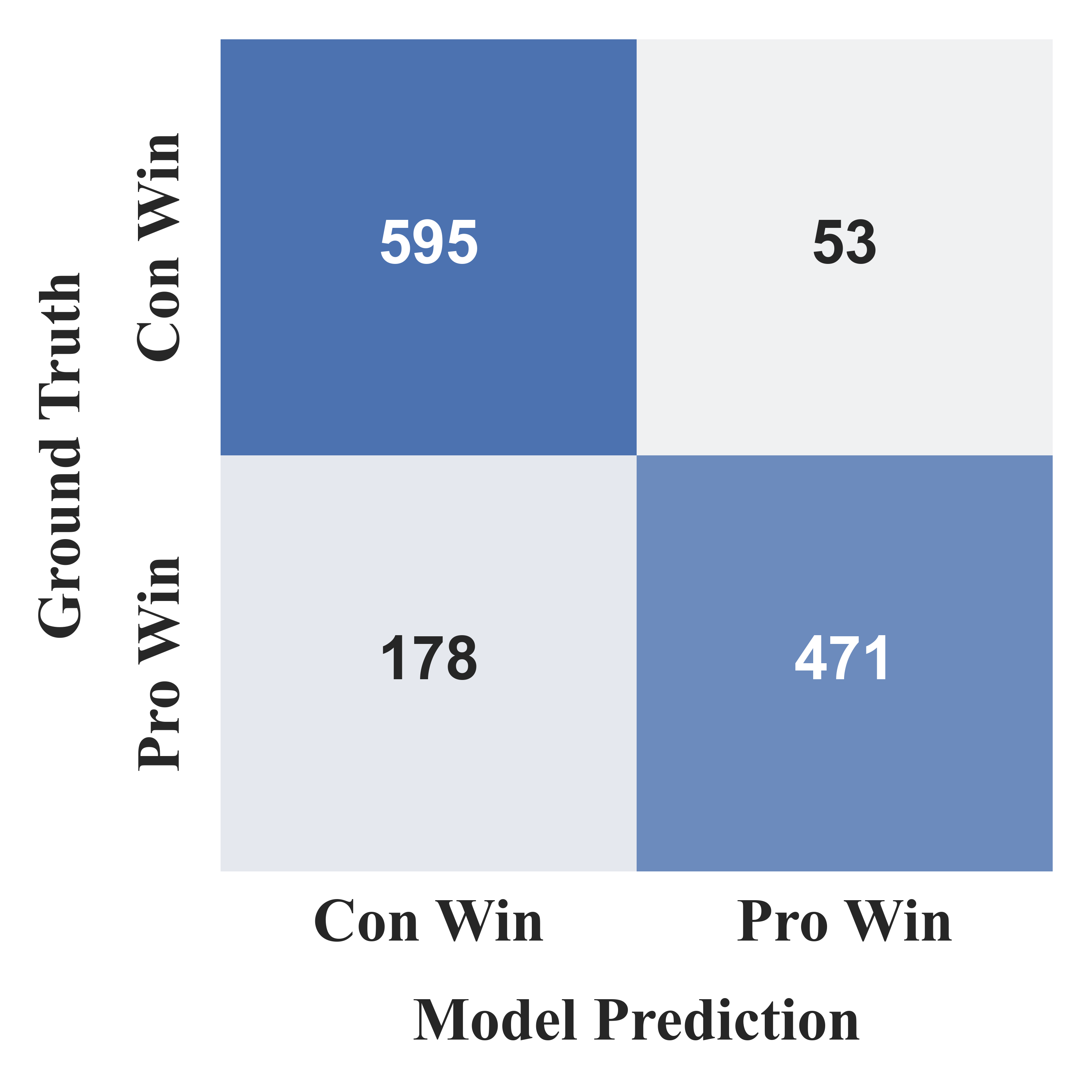}
        \caption{1/-1 label set}
        \label{fig:GPT4 sub3}
    \end{subfigure}
    \hfill
    \begin{subfigure}[b]{0.245\linewidth}
        \includegraphics[width=\linewidth]{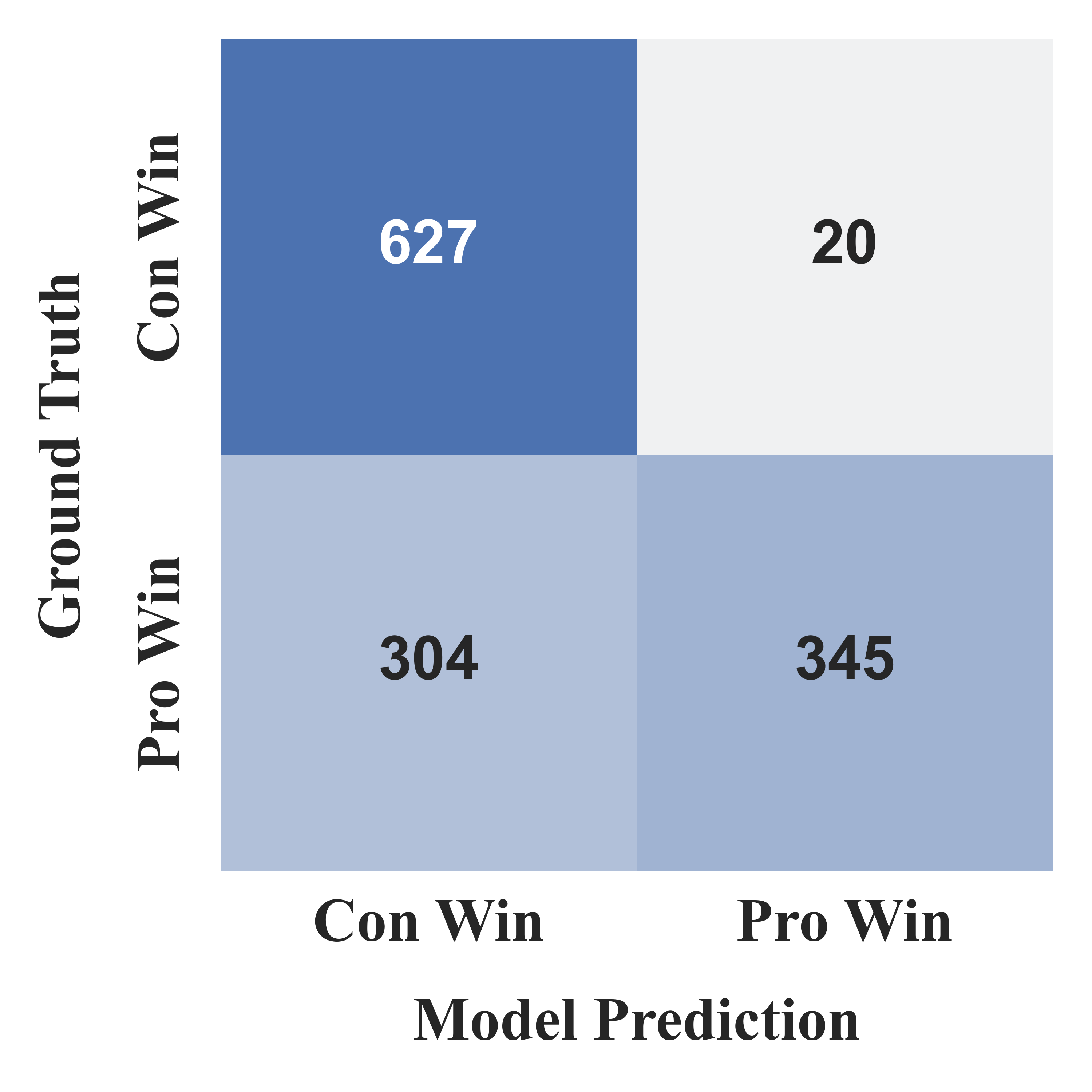}
        \caption{Pro/Con label set}
        \label{fig:GPT4 sub4}
    \end{subfigure}

    \vspace{1mm} 

    \begin{subfigure}[b]{0.245\linewidth}
        \includegraphics[width=\linewidth]{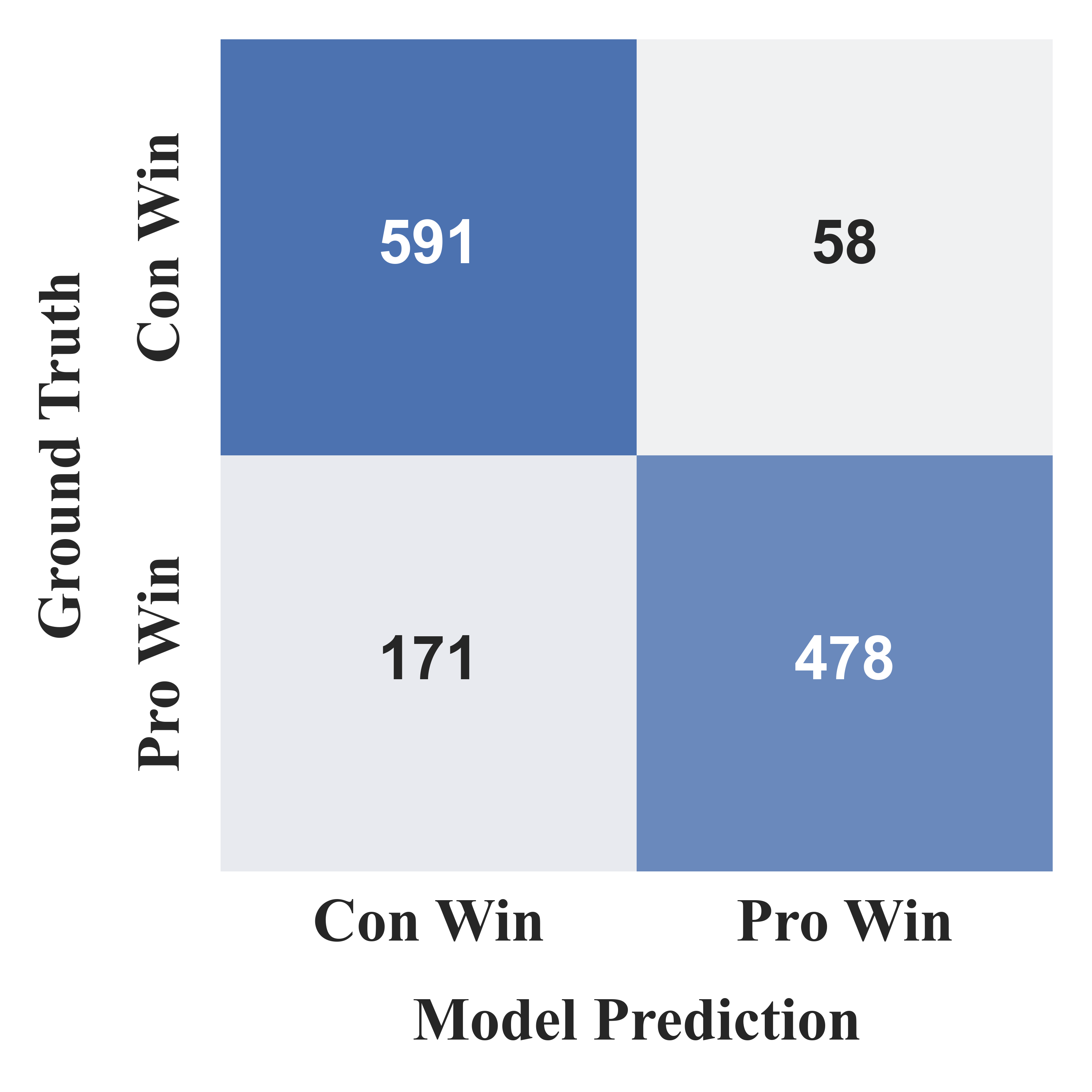}
        \caption{Shuffled A/B label set}
        \label{fig:GPT4 sub5}
    \end{subfigure}
    \hfill
    \begin{subfigure}[b]{0.245\linewidth}
        \includegraphics[width=\linewidth]{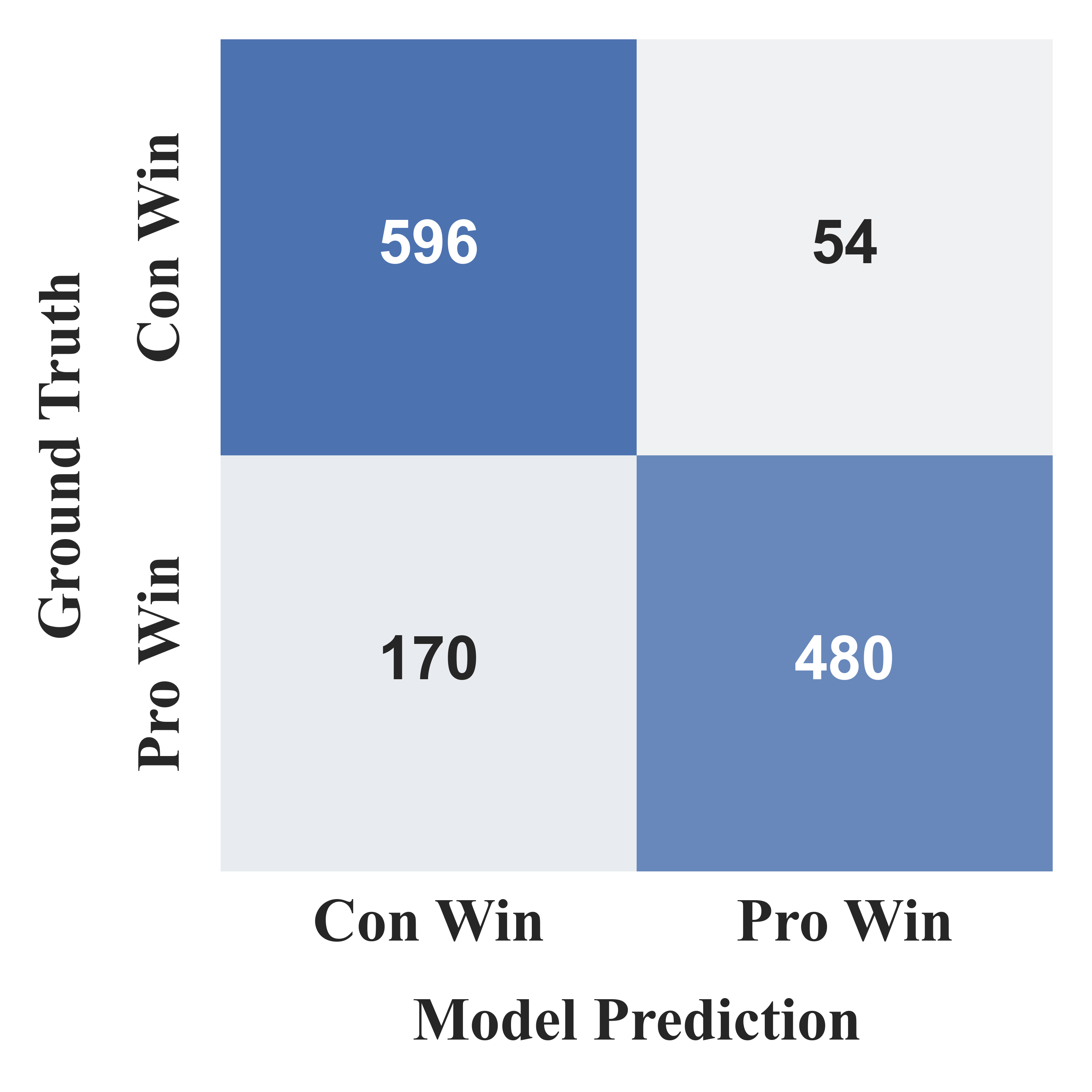}
        \caption{Shuffled P/C label set}
        \label{fig:GPT4 sub6}
    \end{subfigure}
    \hfill
    \begin{subfigure}[b]{0.245\linewidth}
        \includegraphics[width=\linewidth]{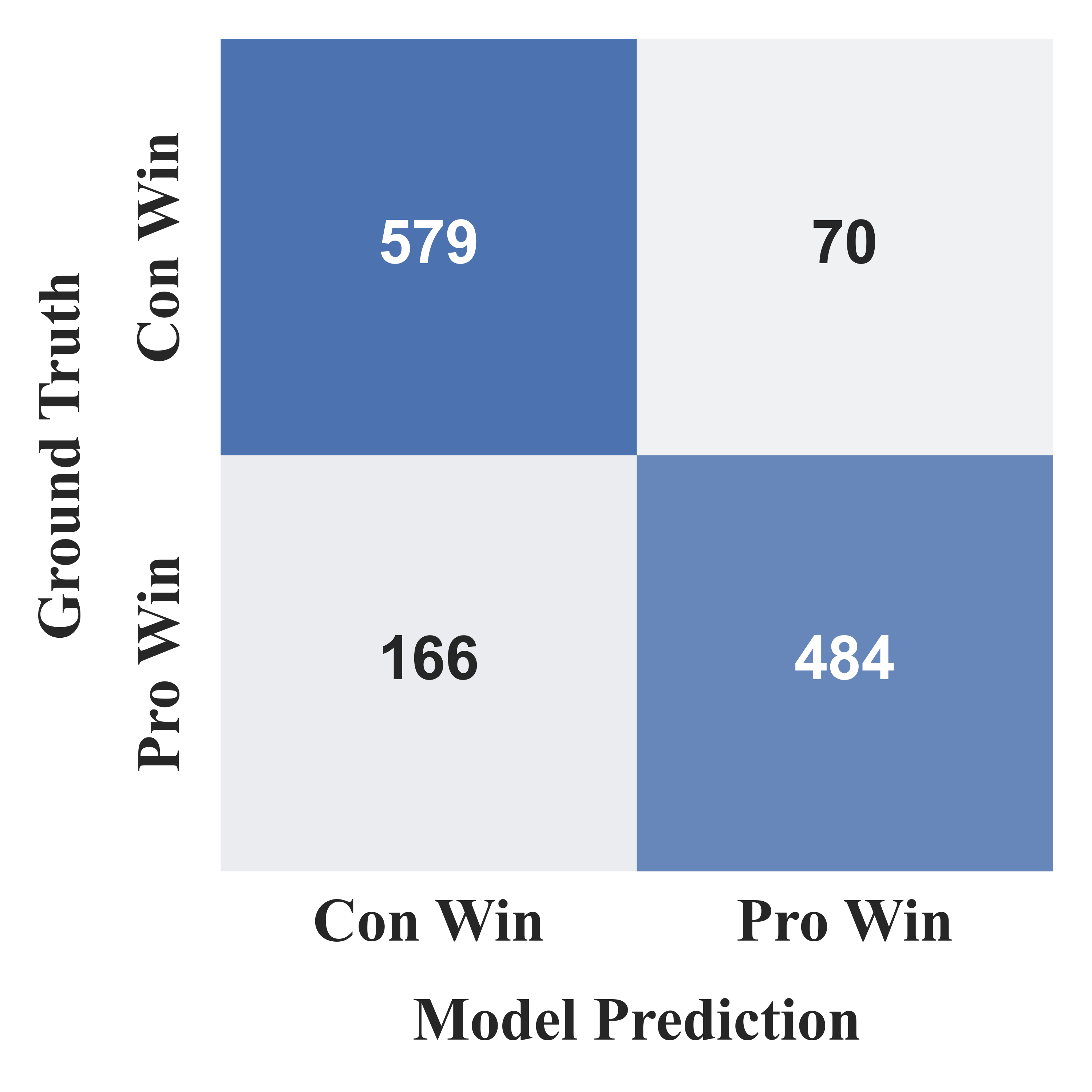}
        \caption{Shuffled 1/-1 label set}
        \label{fig:GPT4 sub7}
    \end{subfigure}
    \hfill
    \begin{subfigure}[b]{0.245\linewidth}
        \includegraphics[width=\linewidth]{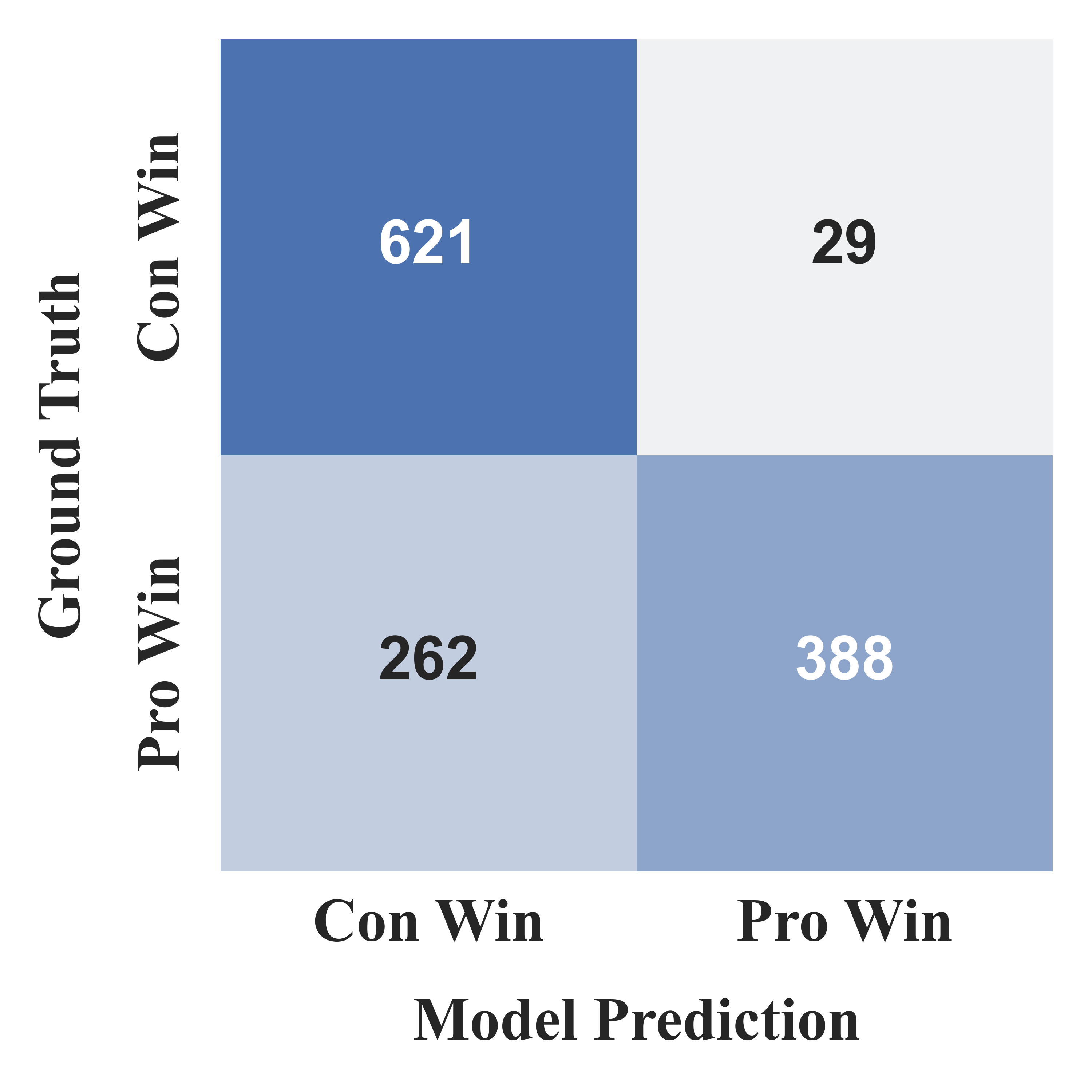}
        \caption{Shuffled Pro/Con label set}
        \label{fig:GPT4 sub8}
    \end{subfigure}

    \vspace{1mm} 

    \begin{subfigure}[b]{0.245\linewidth}
        \includegraphics[width=\linewidth]{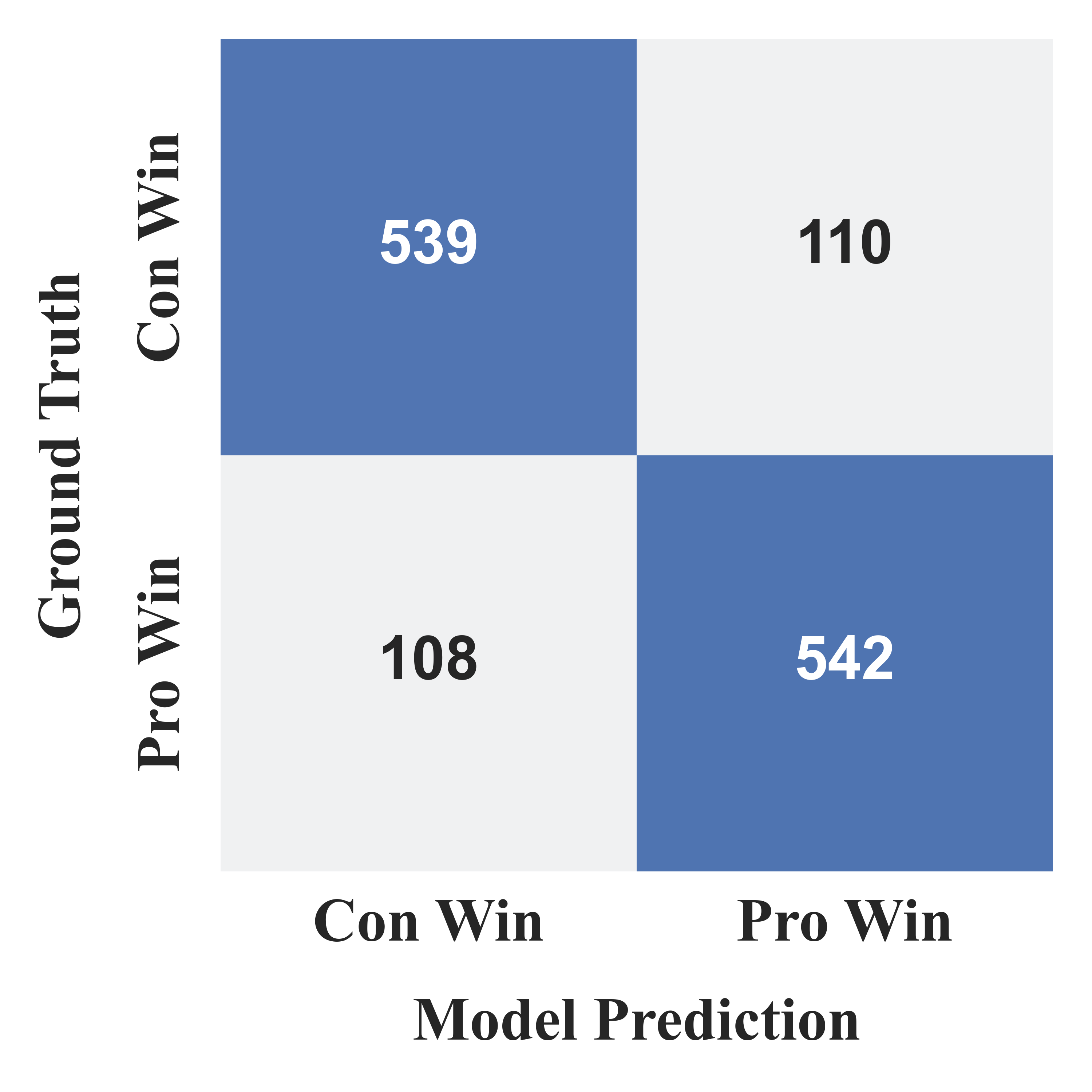}
        \caption{Shuffled B/A label set}
        \label{fig:GPT4 sub9}
    \end{subfigure}
    \hfill
    \begin{subfigure}[b]{0.245\linewidth}
        \includegraphics[width=\linewidth]{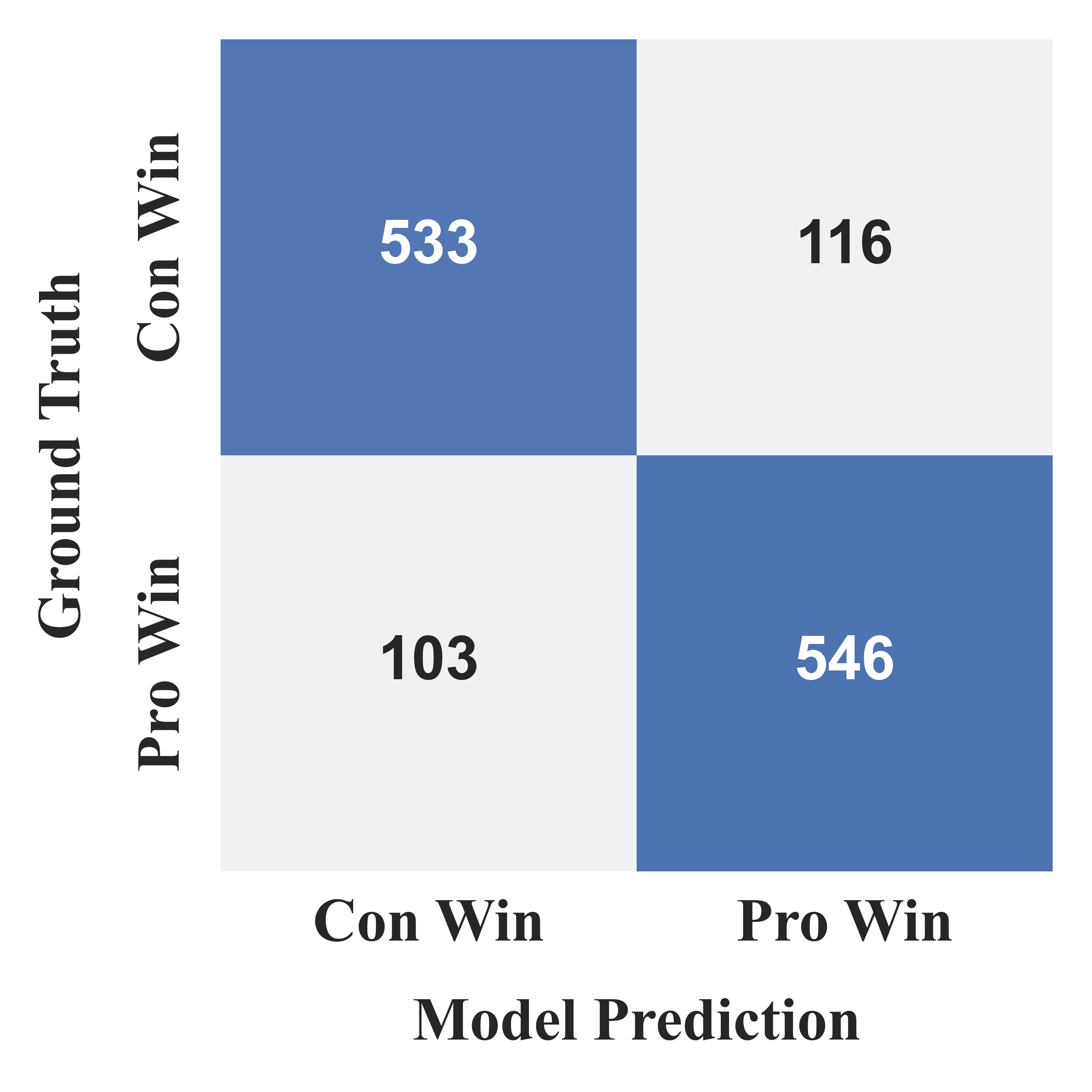}
        \caption{Shuffled C/P label set}
        \label{fig:GPT4 sub10}
    \end{subfigure}
    \hfill
    \begin{subfigure}[b]{0.245\linewidth}
        \includegraphics[width=\linewidth]{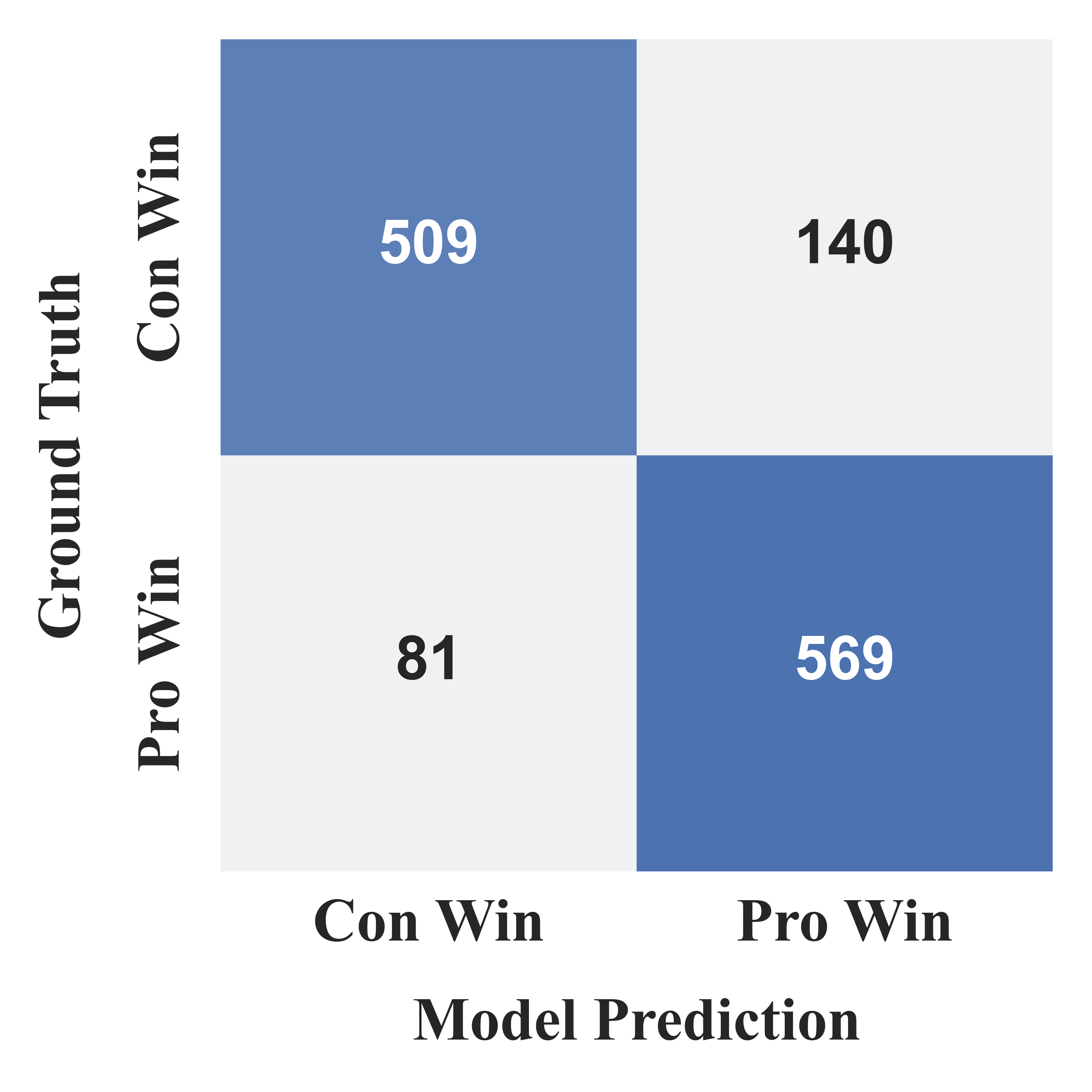}
        \caption{Shuffled -1/1 label set}
        \label{fig:GPT4 sub11}
    \end{subfigure}
    \hfill
    \begin{subfigure}[b]{0.245\linewidth}
        \includegraphics[width=\linewidth]{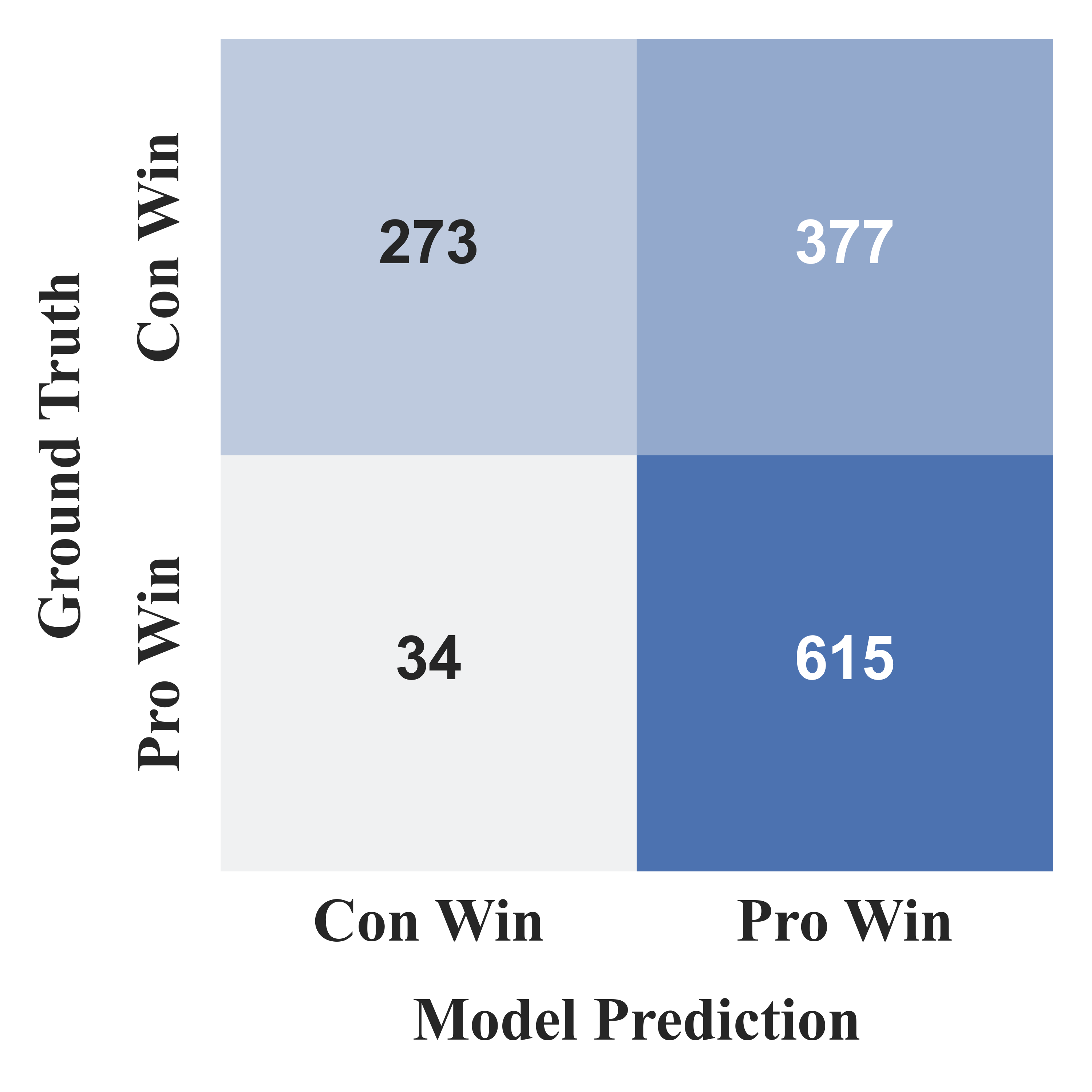}
        \caption{Shuffled Con/Pro label set}
        \label{fig:GPT4 sub12}
    \end{subfigure}
    
    \caption{This figure illustrates the impact of lexical bias on GPT-4 through confusion matrices for various Pro\_label/Con\_label sets. The matrices in the first row correspond to scenarios where the Pro\_label consistently occupies the leading position in the instruction prompt, potentially introducing a positional bias. In contrast, the second and third rows present matrices from experiments where the positions of Pro\_label and Con\_label are shuffled, aiming to mitigate this bias.}
    \label{fig:confusion matrcies GPT-4}
    \vspace{-6mm}
\end{figure*}

The detailed confusion matrices with various settings we experiment on balanced datasets can be found in Figure \ref{fig:confusion matrices GPT-3.5} for GPT-3.5 and in Figure \ref{fig:confusion matrcies GPT-4} for GPT-4.

\paragraph{Performance.} We also test GPT-3.5 and GPT-4 on the same subset of human-annotated data. The accuracies achieved by GPT-3.5 and GPT-4 are 79.73\% and 84.00\% respectively.

\paragraph{Positional Bias.}

For GPT-3.5, McNemar's tests \cite{mcnemar1947note} are also conducted for the settings with shuffled candidate response positions and fixed positions based on Table \ref{tab: mcnemar test for positional bias}, and the results are all significant.
\begin{table}[ht]
\centering
\begin{tabular}{ccccc}
\toprule
\textbf{Verbalizers} & $f_{\text{fixed\_shuffled}}$ & $f_{\text{shuffled\_fixed}}$ & $\chi^2$ & \textbf{P-Value} \\
\midrule
A/B & 25 & 86 & 33.52 & < 0.001 \\
P/C & 16 & 205 & 161.63 & < 0.001 \\
1/-1 & 38 & 124 & 45.65 & < 0.001 \\
Pro/Con & 34 & 79 & 17.92 & < 0.001 \\
\bottomrule
\end{tabular}
\caption{McNemar's test demonstrates that all positional biases are significant within GPT-3.5. $f_{\text{fixed\_shuffled}}$ indicates the number of debates predicted as Pro winning by the first verbalizer set but Con winning by the second verbalizer set. $f_{\text{shuffled\_fixed}}$ indicates the number of debates predicted as Pro winning with shuffled positions but Con winning by GPT-3.5 with fixing Pro as the first candidate response.}
\label{tab: mcnemar test for positional bias}
\end{table}

\begin{figure*}[h]
    \centering
    \captionsetup[subfigure]{font=scriptsize}

    \begin{subfigure}[b]{0.24\linewidth}
        \includegraphics[width=\linewidth]{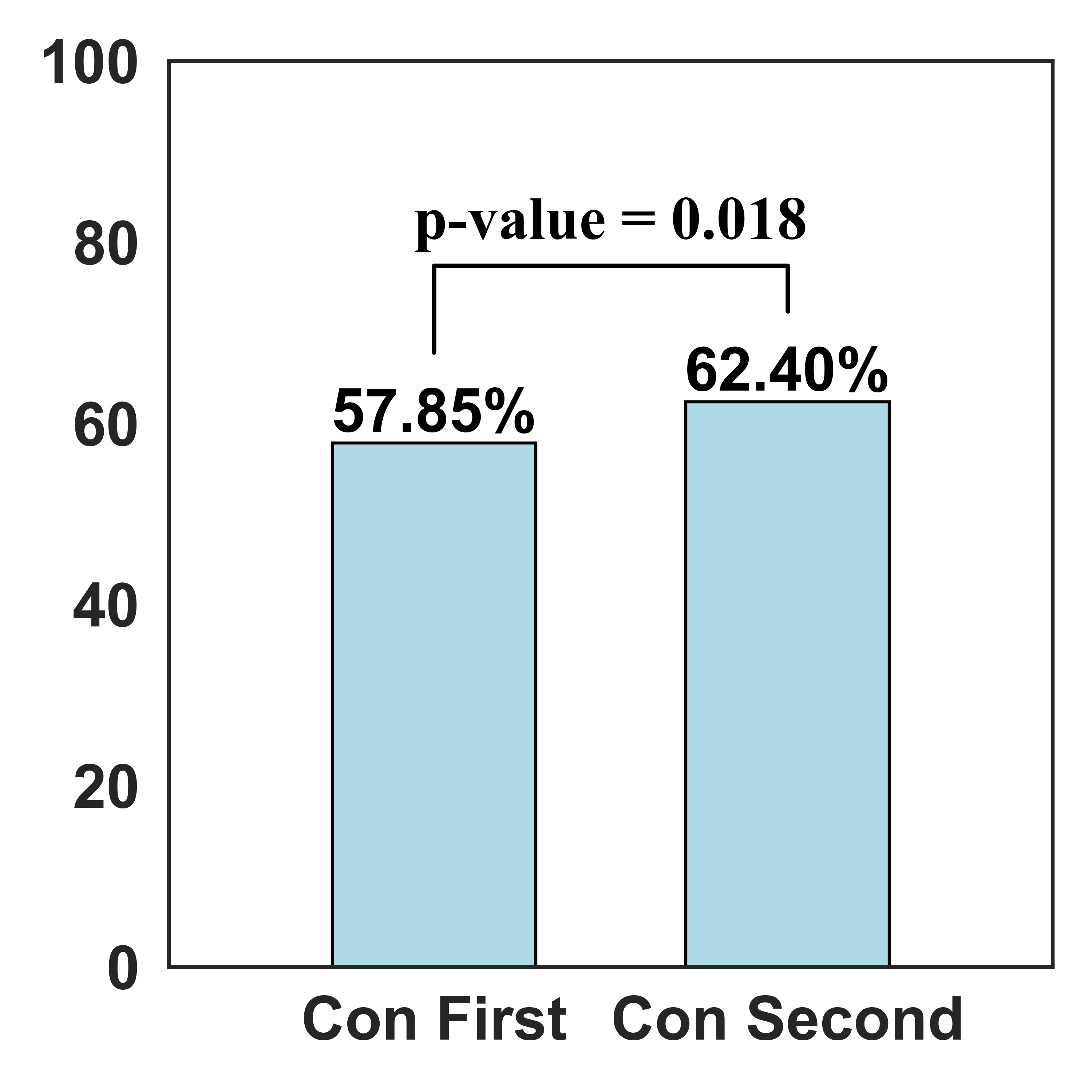}
        \caption{A/B label set}
        \label{fig:positional bias GPT4 sub1}
    \end{subfigure}
    \hfill 
    \begin{subfigure}[b]{0.24\linewidth}
        \includegraphics[width=\linewidth]{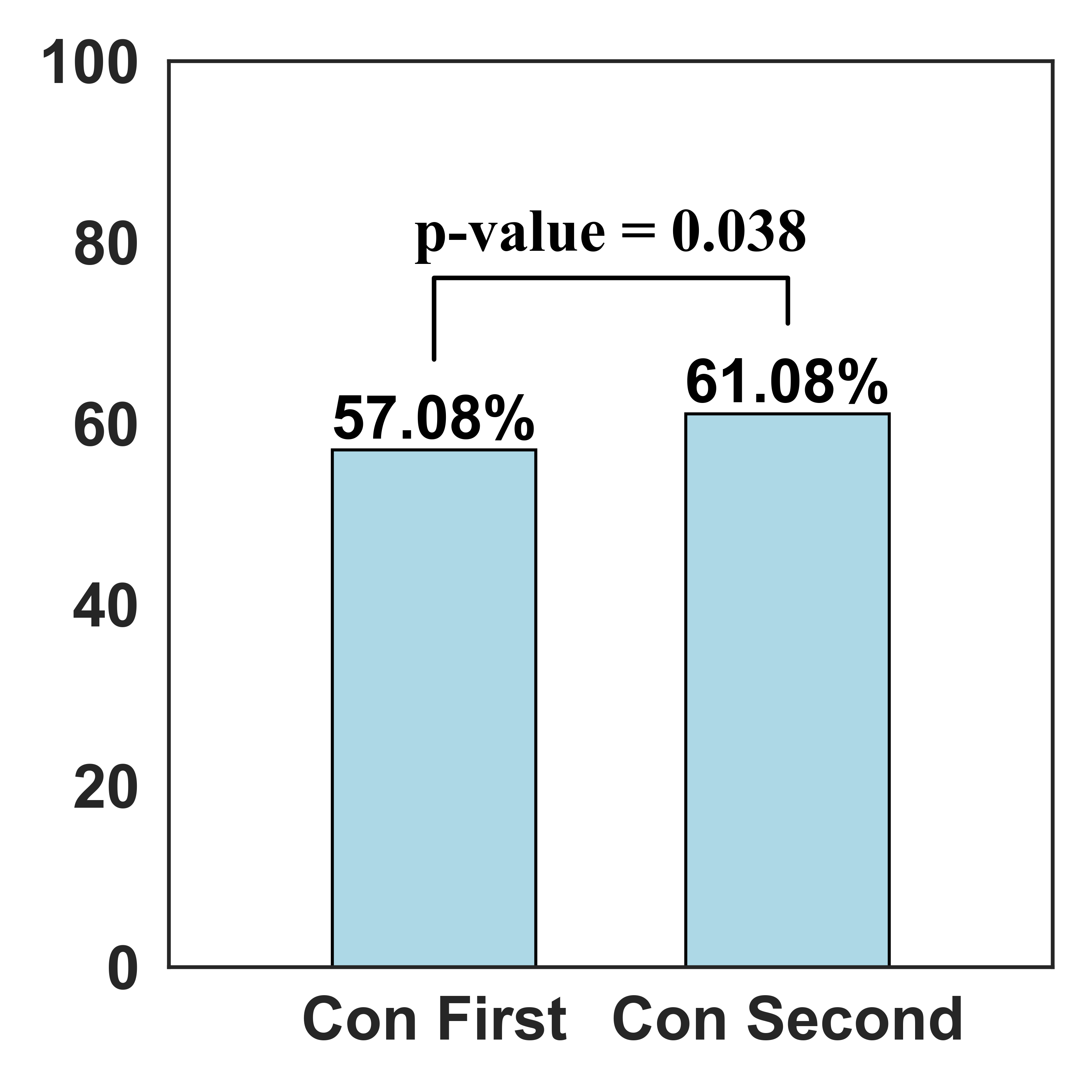}
        \caption{P/C label set}
        \label{fig:positional bias GPT4 sub2}
    \end{subfigure}
    \hfill 
    \begin{subfigure}[b]{0.24\linewidth}
        \includegraphics[width=\linewidth]{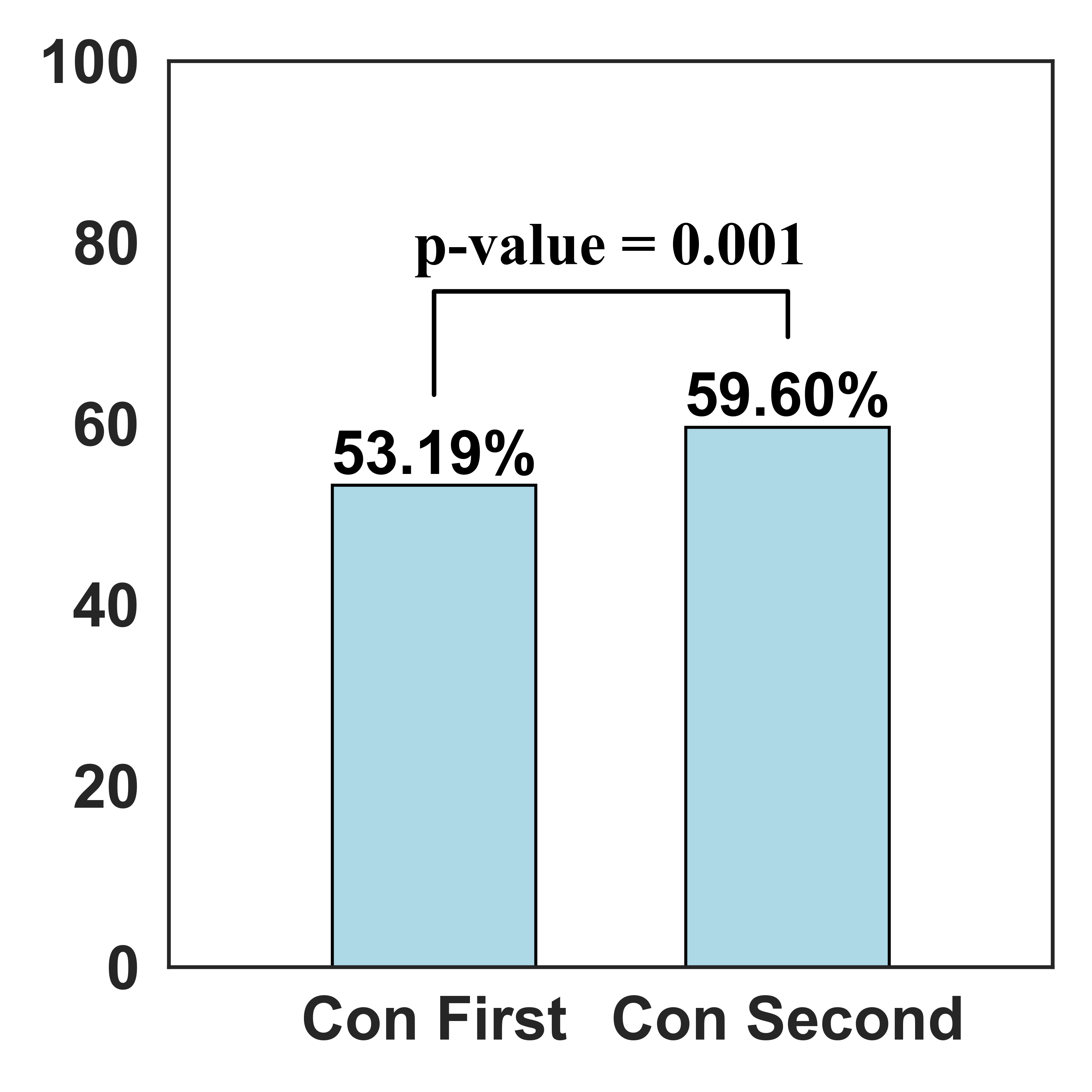}
        \caption{1/-1 label set}
        \label{fig:positional bias GPT4 sub3}
    \end{subfigure}
    \hfill 
    \begin{subfigure}[b]{0.24\linewidth}
        \includegraphics[width=\linewidth]{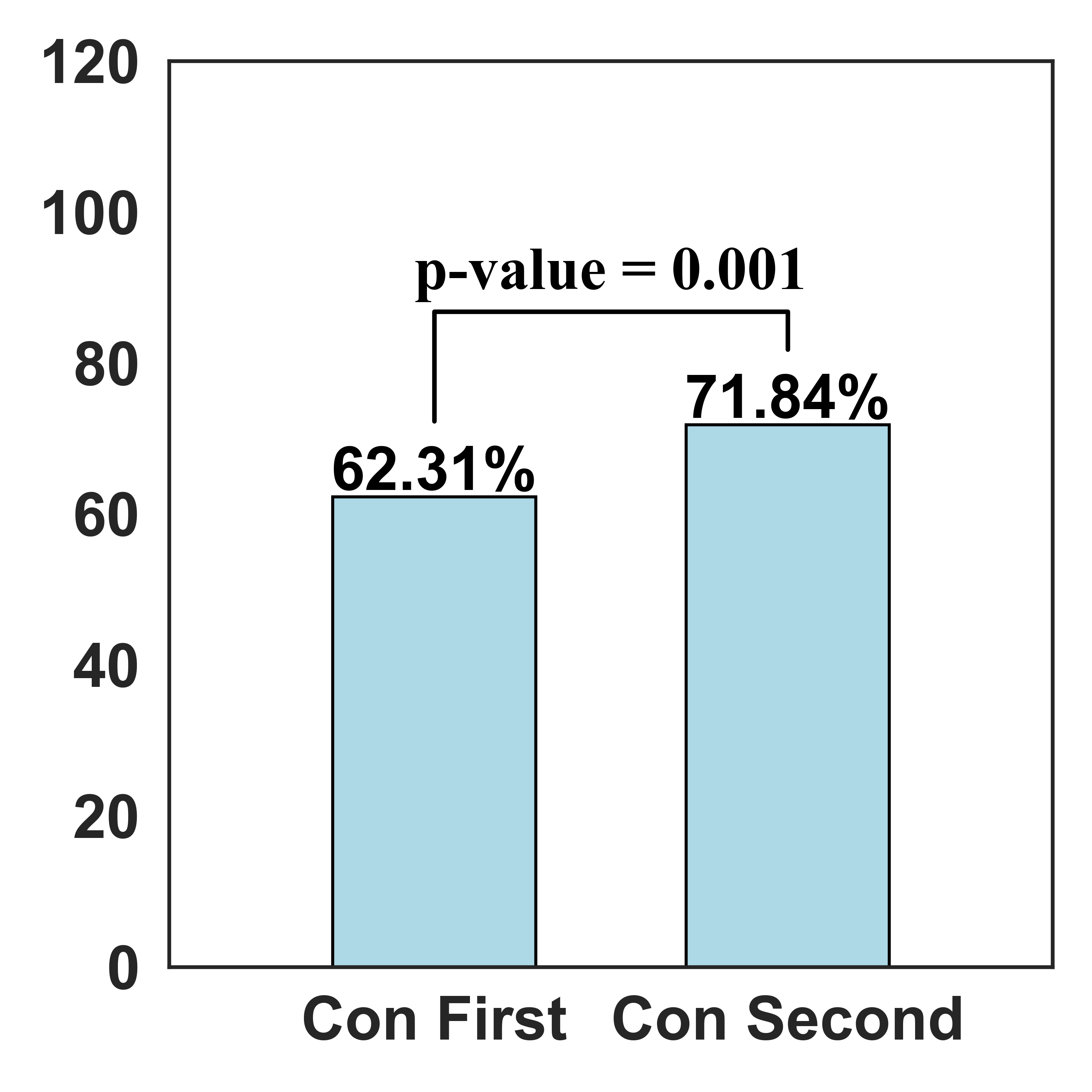}
        \caption{Pro/Con label set}
        \label{fig:positional bias GPT4 sub4}
    \end{subfigure}

    \caption{This figure illustrates the impact of positional bias on GPT-4 through the changes in the proportion of Predicted Con, shifting from when Con is fixed as the first candidate response to when it is positioned as the second. GPT-4 exhibits a positional bias towards the second candidate presented across all label set configurations.}
    \label{fig: positional bias in GPT-4}
\end{figure*}

The direction of the positional bias presented by GPT-4 is also shown towards the second position, contradicting the finding of the first position favorite illustrated by \citet{wang2023large}. The two-sided p-value from the two-portion z-tests demonstrates that the positional bias in GPT-4 as shown in Figure \ref{fig: positional bias in GPT-4} is also statistically significant.

\paragraph{Lexical Bias.}

\begin{figure}[ht]
    \centering
    \begin{subfigure}[b]{0.32\linewidth}
        \includegraphics[width=\linewidth]{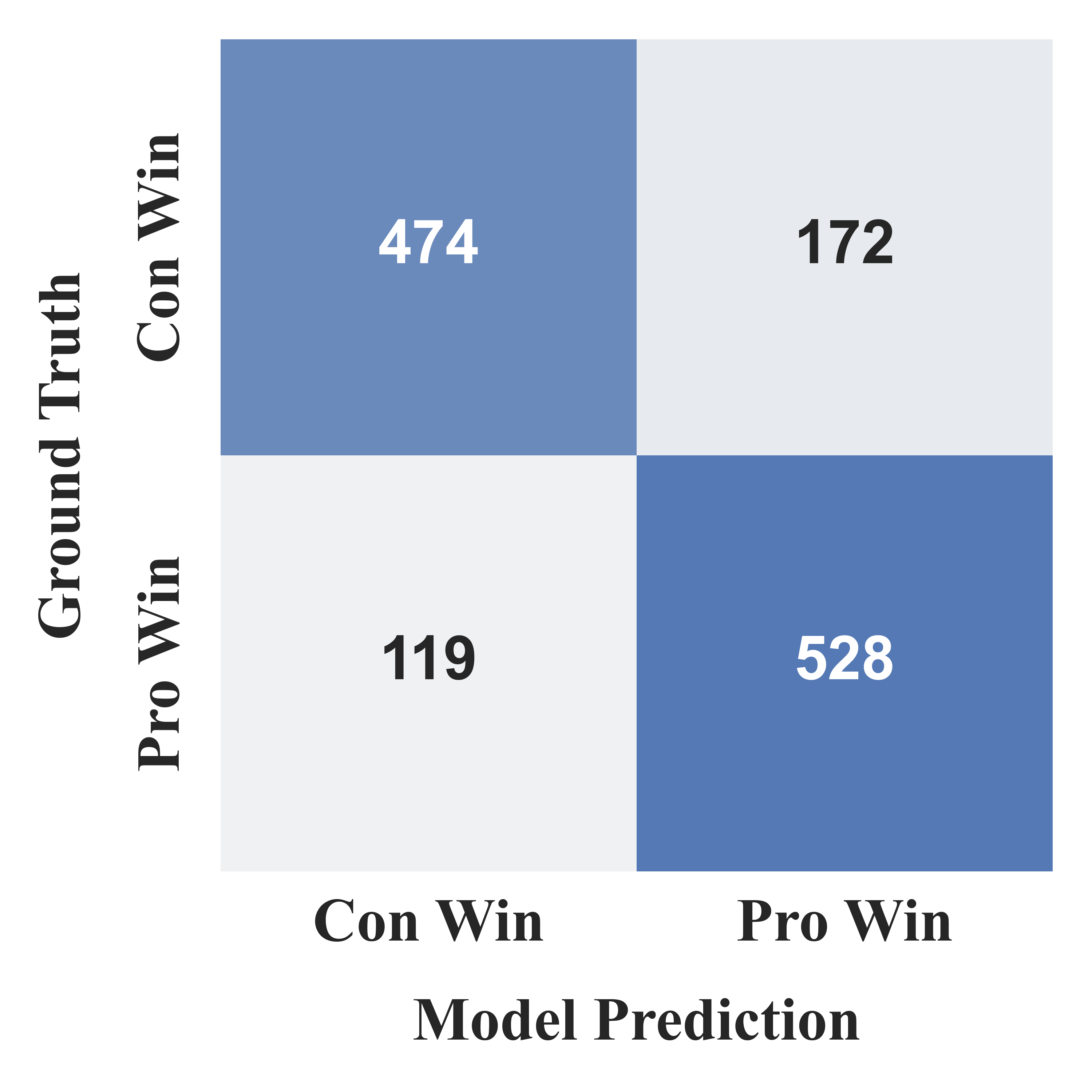}
        \caption{M/N label set}
        \label{fig:MN bias sub1}
    \end{subfigure}
    \hfill
    \begin{subfigure}[b]{0.32\linewidth}
        \includegraphics[width=\linewidth]{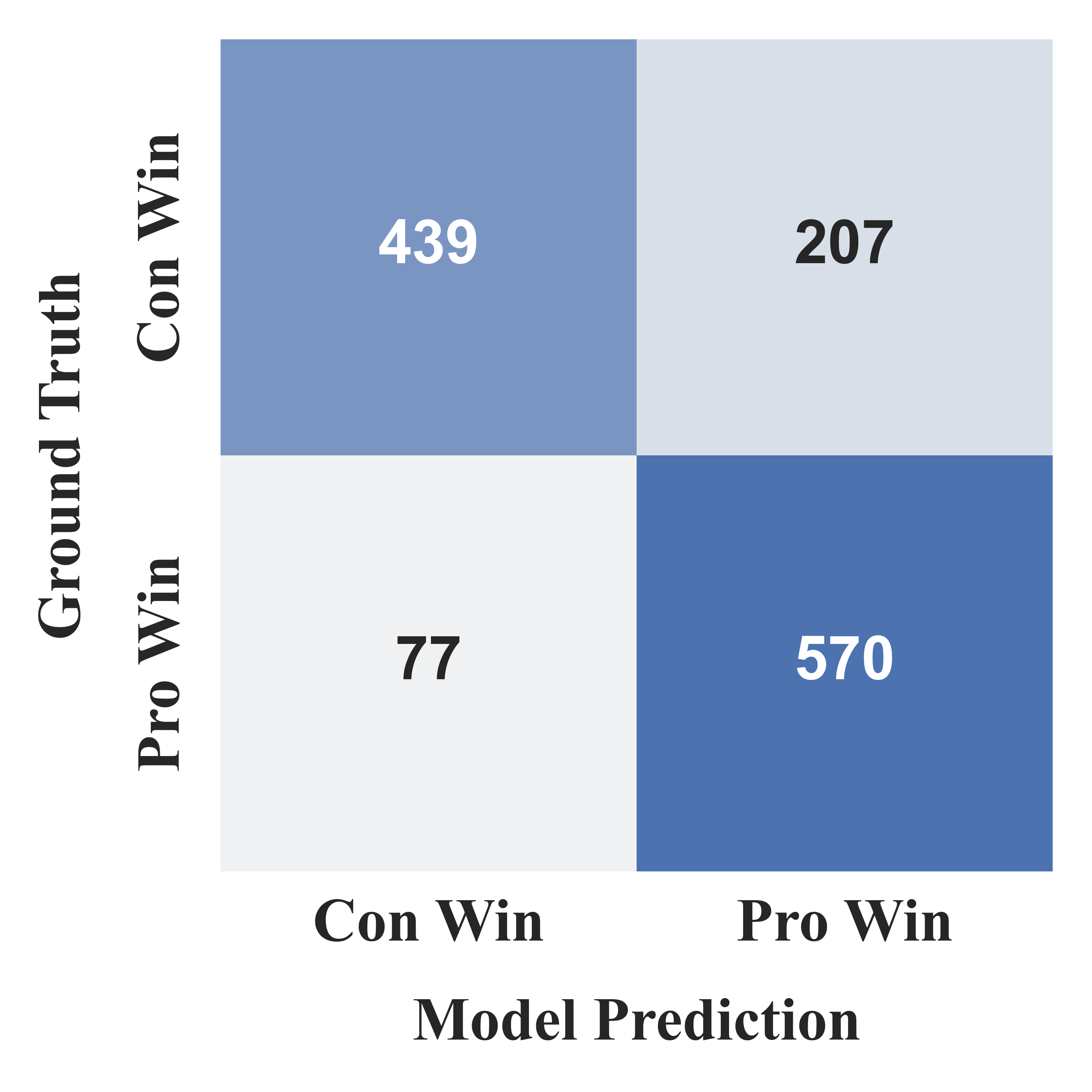}
        \caption{Shuffled M/N label set}
        \label{fig:MN bias sub2}
    \end{subfigure}
    \hfill
    \begin{subfigure}[b]{0.32\linewidth}
        \includegraphics[width=\linewidth]{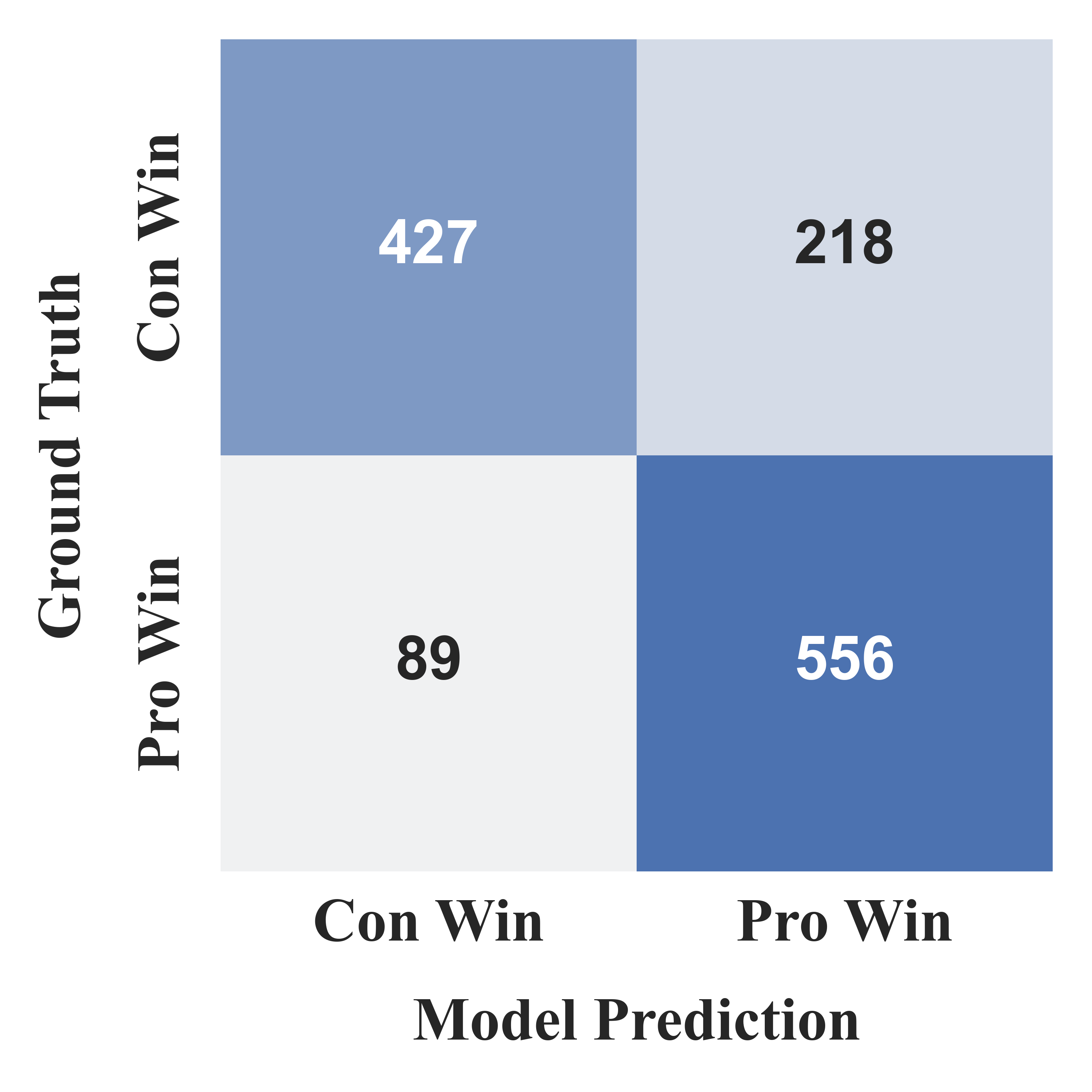}
        \caption{Shuffled N/M label set}
        \label{fig:MN bias sub3}
    \end{subfigure}
    \caption{Lexical bias of M/N label set in GPT-3.5}
    \label{fig:MN bias}
\end{figure}

The difference in the significance of lexical bias within the A/B label set and P/C label set could be due to the alphabetical distance they have or due to their common usage. To discern the underlying cause, we further experiment with the M/N label set for they are alphabetically adjacent but not typically associated with sequential interpretation. The results, detailed in Figure \ref{fig:MN bias}, reveal minimal lexical bias within the M/N group, suggesting that the bias originates from conventional usage rather than alphabetic proximity.

To further quantitatively assess the lexical bias in GPT-3.5, we employ McNemar's test to analyze instances of concordances (both predict Pro or Con), instances of discordances (one predicts Pro and the other predict Con) of each flipping group as shown in Table \ref{tab: mcnemar test for lexical bias}. All results are statistically significant.

\begin{table}[ht]
\centering
\small
\begin{tabular}{ccccc}
\toprule
\textbf{Verbalizers} & \textbf{$f_{12}$} & \textbf{$f_{21}$} & \textbf{$\chi^2$} & \textbf{P-Value} \\
\midrule
A/B vs B/A & 59 & 178 & 59.751 & < 0.001 \\
P/C vs C/P & 166 & 99 & 16.94 & < 0.001 \\
1/-1 vs -1/1 & 33 & 556 & 464.40 & < 0.001 \\
Pro/Con vs Con/Pro & 147 & 298 & 51.24 & < 0.001 \\
\bottomrule
\end{tabular}
\caption{McNemar's test demonstrates that all lexical biases are significant within GPT-3.5. $f_{12}$ indicates the number of debates predicted as Pro winning by the first verbalizer set but Con winning by the second verbalizer set. $f_{21}$ indicates the number of debates predicted as Pro winning by the second verbalizer set but Con winning by the first verbalizer set. The positions of verbalizers in the prompt are shuffled.}
\label{tab: mcnemar test for lexical bias}
\end{table}

Similar to GPT-3.5, GPT-4 also exhibits lexical bias towards 'B', '-1' and potentially 'Con' within the A/B, 1/-1, and Pro/Con label set. However, GPT-4 favors 'C' over 'P' significantly. McNemar's tests of lexical bias for GPT-4 are shown in Table \ref{tab: mcnemar test for lexical bias in GPT-4} for GPT-4.

\begin{figure*}[t]
    \centering
    \captionsetup[subfigure]{font=scriptsize}

    \begin{subfigure}[b]{0.24\linewidth}
        \includegraphics[width=\linewidth]{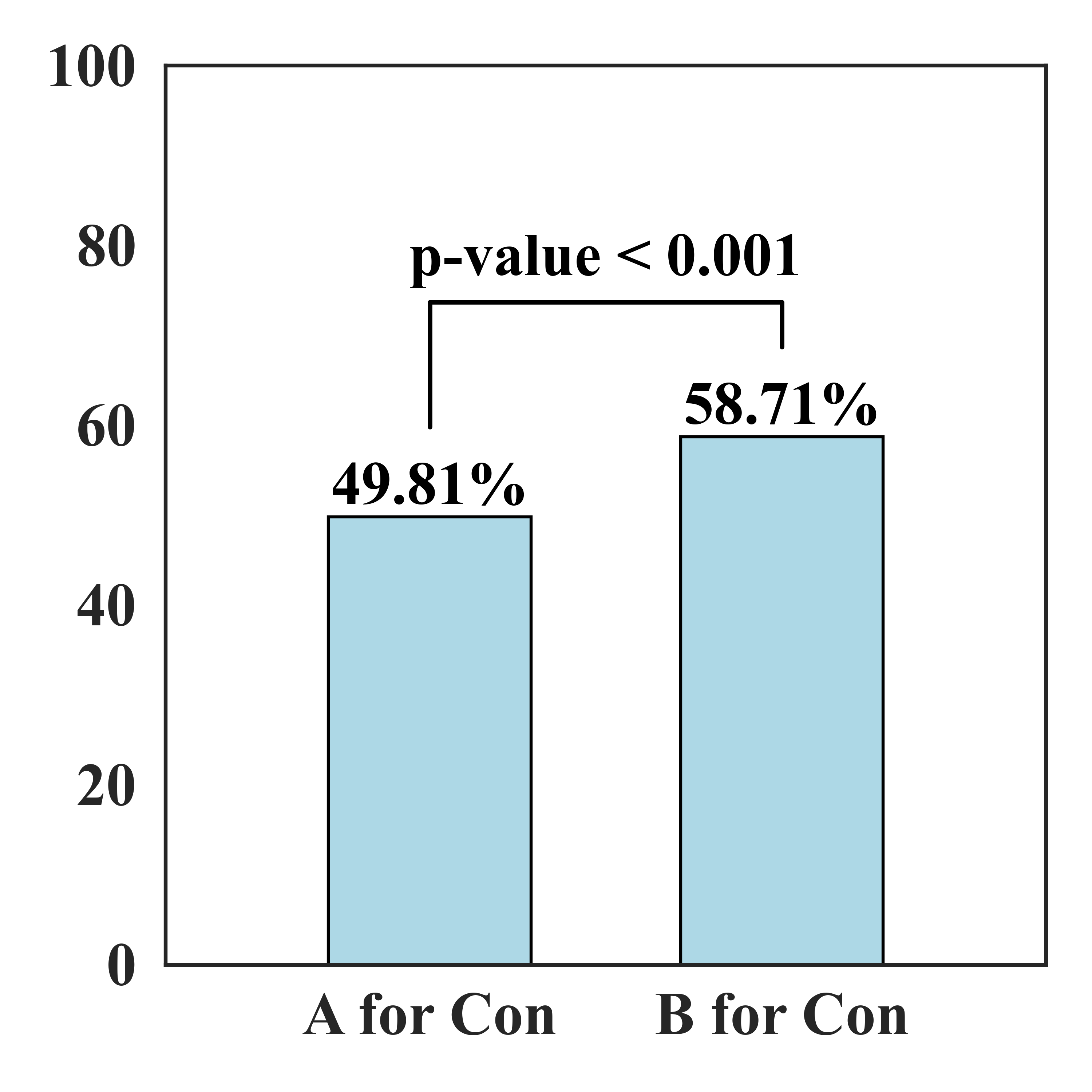}
        \caption{Shuffled A/B vs. Shuffled B/A}
        \label{fig:lexical bias GPT4 sub1}
    \end{subfigure}
    \hfill 
    \begin{subfigure}[b]{0.24\linewidth}
        \includegraphics[width=\linewidth]{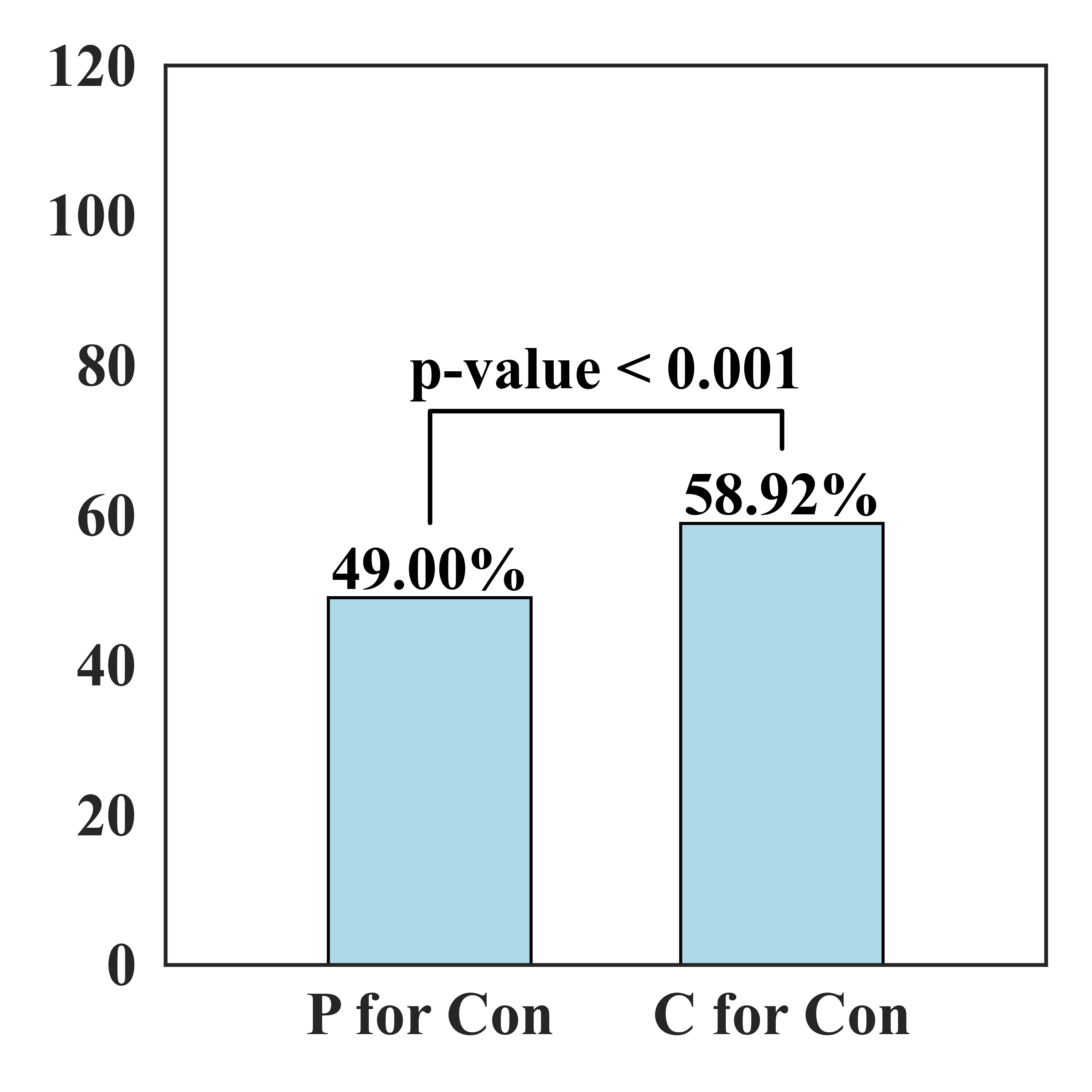}
        \caption{Shuffled P/C vs. Shuffled C/P}
        \label{fig:lexical bias GPT4 sub2}
    \end{subfigure}
    \hfill 
    \begin{subfigure}[b]{0.24\linewidth}
        \includegraphics[width=\linewidth]{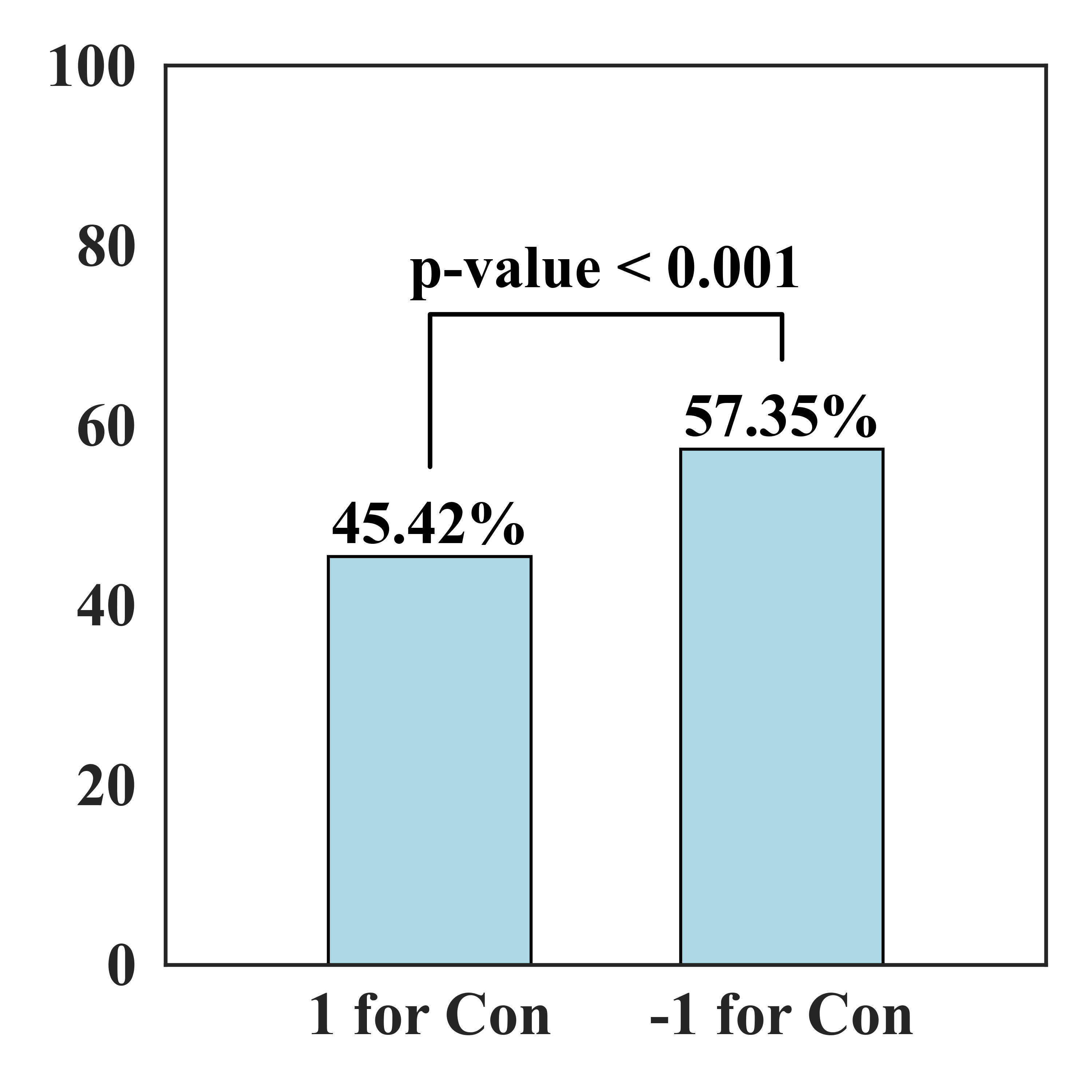}
        \caption{Shuffled 1/-1 vs. Shuffled -1/1}
        \label{fig:lexical bias GPT4 sub3}
    \end{subfigure}
    \hfill 
    \begin{subfigure}[b]{0.24\linewidth}
        \includegraphics[width=\linewidth]{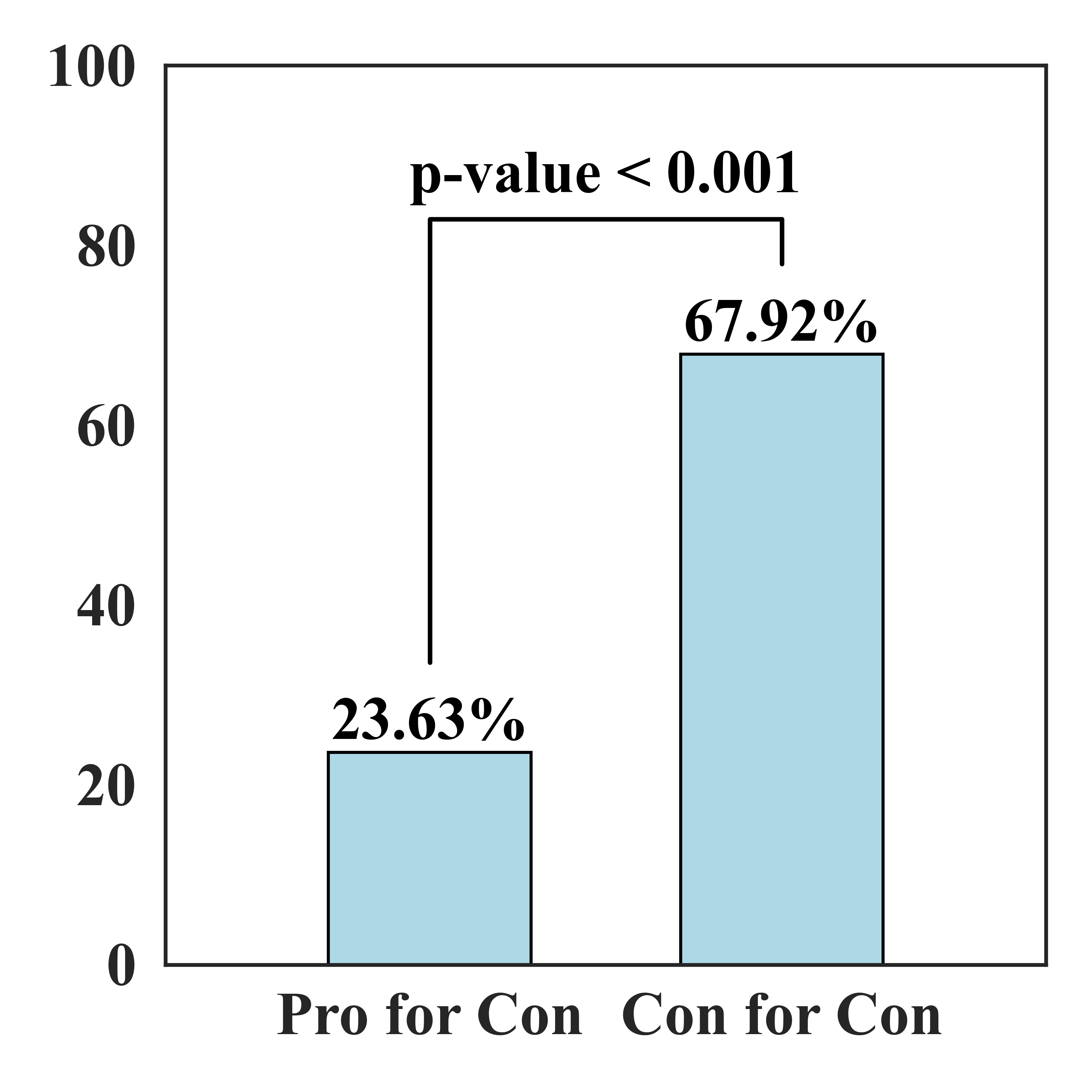}
        \caption{\tiny Shuffled Pro/Con vs. Shuffled Con/Pro}
        \label{fig:lexical bias GPT4 sub4}
    \end{subfigure}

    \caption{This figure illustrates the impact of lexical bias on GPT-4 through the changes in the portion of Predicted Con from switching the verbalizers for Pro and Con.}
    \label{fig: lexical bias in GPT-4}
\end{figure*}

\begin{table}[ht]
\centering
\small
\begin{tabular}{ccccc}
\toprule
\textbf{Verbalizers} & \textbf{$f_{\text{fixed\_shuffled}}$} & \textbf{$f_{\text{shuffled\_fixed}}$} & \textbf{$\chi^2$} & \textbf{P-Value} \\
\midrule
A/B & 6 & 54 & 36.82 & < 0.001 \\
P/C & 9 & 39 & 17.52 & < 0.001 \\
1/-1 & 15 & 45 & 14.02 & 0.002 \\
Pro/Con & 11 & 63 & 35.15 & < 0.001 \\
\bottomrule
\end{tabular}
\caption{McNemar's test demonstrates that all positional biases are significant within GPT-4. $f_{\text{fixed\_shuffled}}$ indicates the number of debates predicted as Pro winning by the first verbalizer set but Con winning by the second verbalizer set.  $f_{\text{shuffled\_fixed}}$ indicates the number of debates predicted as Pro winning with shuffled positions but Con winning by GPT-3.5 with fixing Pro as the first candidate response.}
\label{tab: mcnemar test for positional bias in GPT-4}
\end{table}


\begin{table}[ht]
\centering
\begin{tabular}{ccccc}
\toprule
\textbf{Verbalizers} & $f_{12}$ & $f_{21}$ & $\chi^2$ & \textbf{P-Value} \\
\midrule
A/B vs B/A & 4 & 36 & 24.03 & < 0.001 \\
P/C vs C/P & 6 & 134 & 115.21 & < 0.001 \\
1/-1 vs -1/1 & 11 & 166 & 133.99 & < 0.001 \\
Pro/Con vs Con/Pro & 6 & 581 & 561.29 & < 0.001 \\
\bottomrule
\end{tabular}
\caption{McNemar's test demonstrates that all lexical biases are significant within GPT-4. $f_{12}$ indicates the number of debates predicted as Pro winning by the first verbalizer set but Con winning by the second verbalizer set. $f_{21}$ indicates the number of debates predicted as Pro winning by the second verbalizer set but Con winning by the first verbalizer set. The positions of verbalizers in the prompt are shuffled.}
\label{tab: mcnemar test for lexical bias in GPT-4}
\end{table}


\paragraph{Order Bias.}
The Chi-squared test to show the association between the GPT-4's predictions and the sides that conclude the debates are shown in Table \ref{tab: chi-squared test for order bias in GPT-4}. Same as GPT-3.5, across all verbalizer choices, the order biases presented by GPT-4 are also statistically significant. In addition, the magnitude of the order bias within GPT-3.5 is much stronger than GPT-4, as measured by the Phi Coefficient. 

\begin{table}[htbp]
\centering
\begin{tabular}{crcccr}
\toprule
\textbf{Verbalizer} & \textbf{End-Side} & \textbf{\# P-Pro} & \textbf{\# P-Con} & \textbf{Phi Coeff.} & \textbf{P-Value} \\
\midrule
\multirow{2}{*}{A/B} & Pro & 359 & 291 & 0.099 & < 0.001   \\
                    & Con & 293 & 356 &   &       \\
\midrule
\multirow{2}{*}{P/C} & Pro & 277 & 372 & 0.076 & 0.006  \\
                     & Con & 227 & 419 &  &      \\
\midrule
\multirow{2}{*}{1/-1} & Pro & 286 & 362 & 0.074 & 0.007  \\
                     & Con & 238 & 411 &    &    \\
\midrule
\multirow{2}{*}{1/-1} & Pro & 203 & 445 & 0.069 & 0.001  \\
                     & Con & 162 & 486 &    &    \\

\bottomrule
\end{tabular}
\caption{GPT-4 predictions and debate conclusions association analysis with significance determined by Chi-square tests. \# P-Pro and \# P-Con denote the number of predicted Pro sides and Con sides as the winner by the model, respectively.}
\label{tab: chi-squared test for order bias in GPT-4}
\end{table}

\begin{figure}[ht]
    \centering
     \captionsetup[subfigure]{font=scriptsize}

    \begin{subfigure}[b]{0.35\linewidth}
        \includegraphics[width=\linewidth]{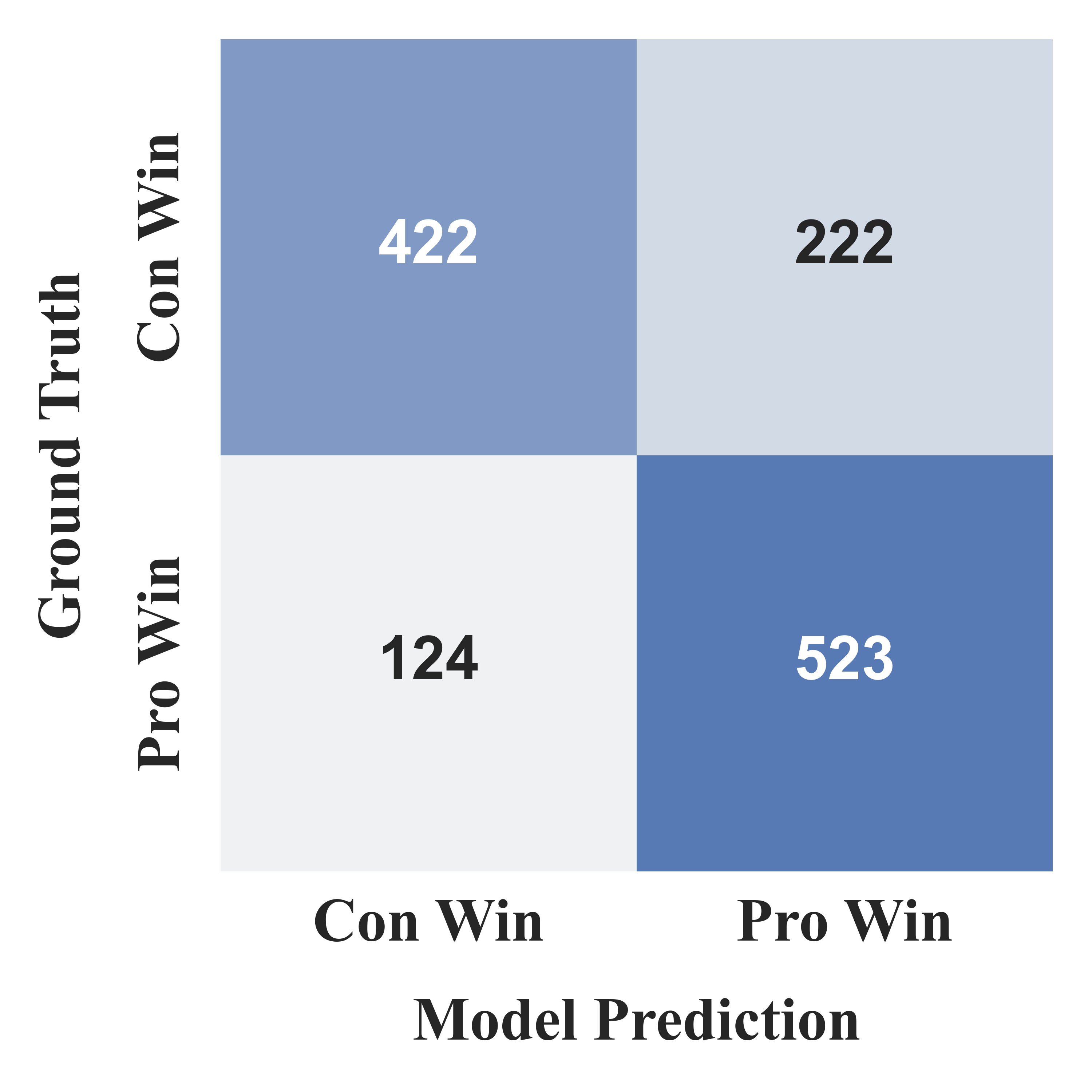}
        \caption{Double shuffled A/B in GPT-3.5}
        \label{fig:stance bias sub1}
    \end{subfigure}
    \hspace{1pt}
    \begin{subfigure}[b]{0.35\linewidth}
        \includegraphics[width=\linewidth]{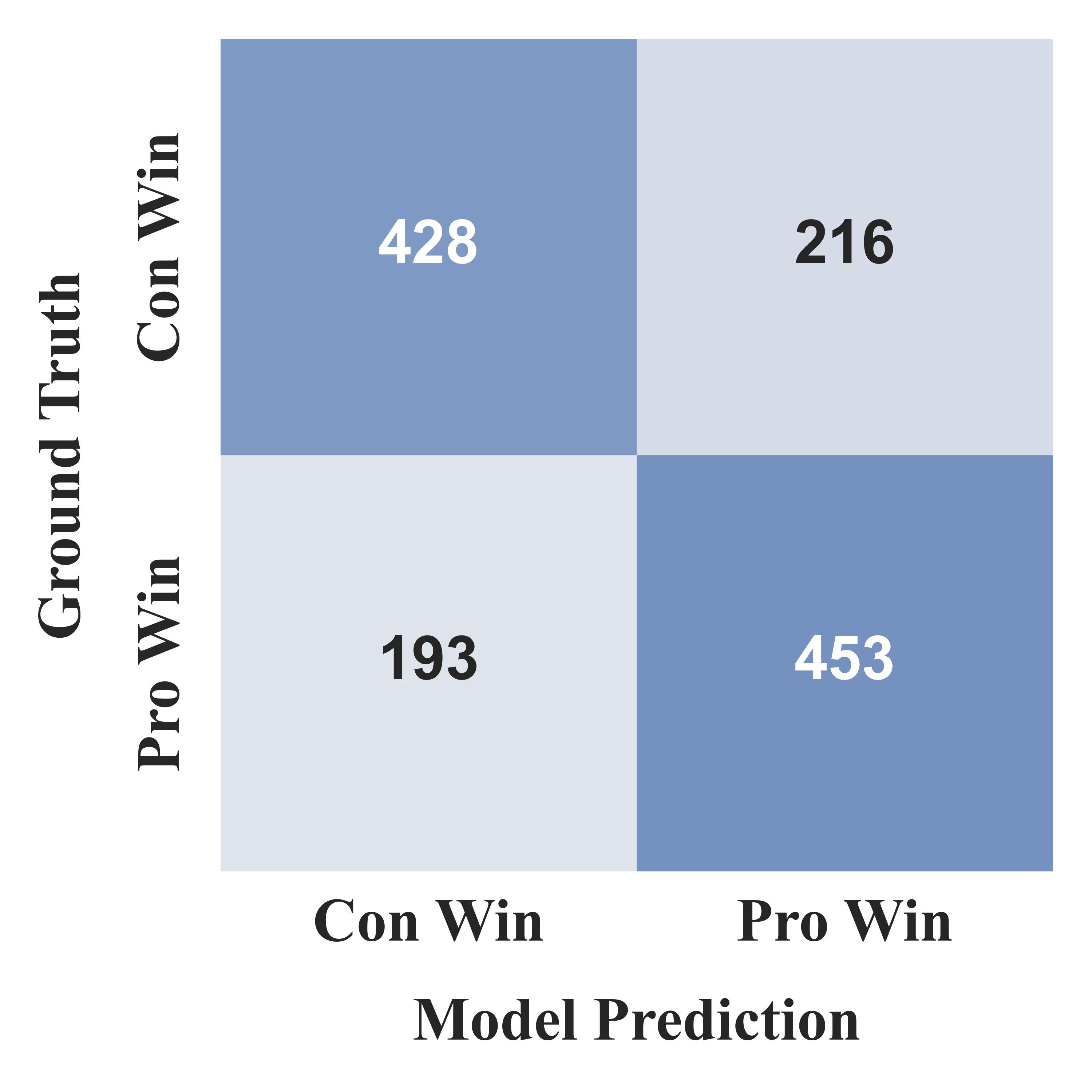}
        \caption{Double shuffled 1/-1 in GPT-3.5}
        \label{fig:stance bias sub2}
    \end{subfigure}
    \vspace{1mm}
    \begin{subfigure}[b]{0.35\linewidth}
        \includegraphics[width=\linewidth]{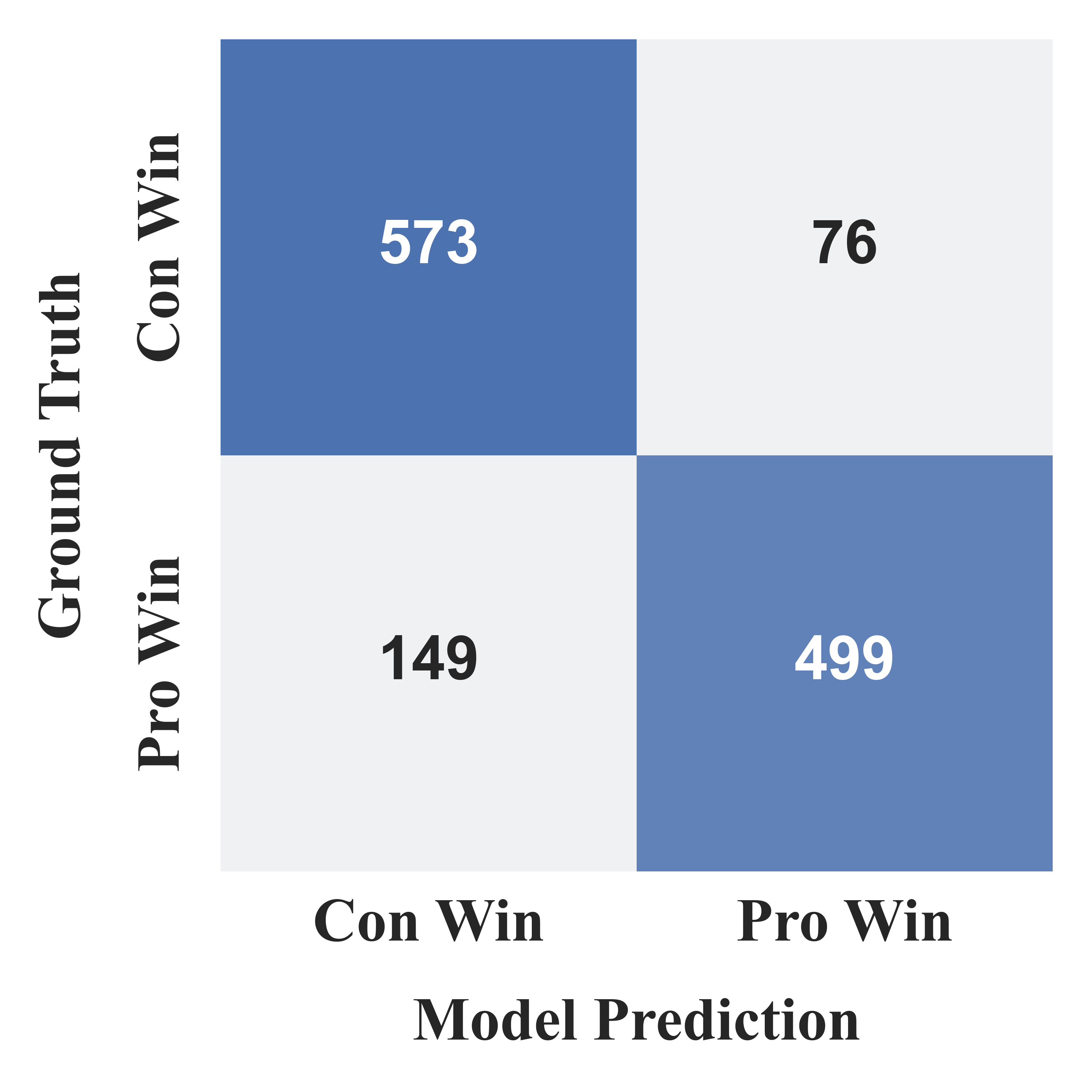}
        \caption{Double shuffled A/B in GPT-4}
        \label{fig:stance bias sub3}
    \end{subfigure}
    \hspace{1pt}
    \begin{subfigure}[b]{0.35\linewidth}
        \includegraphics[width=\linewidth]{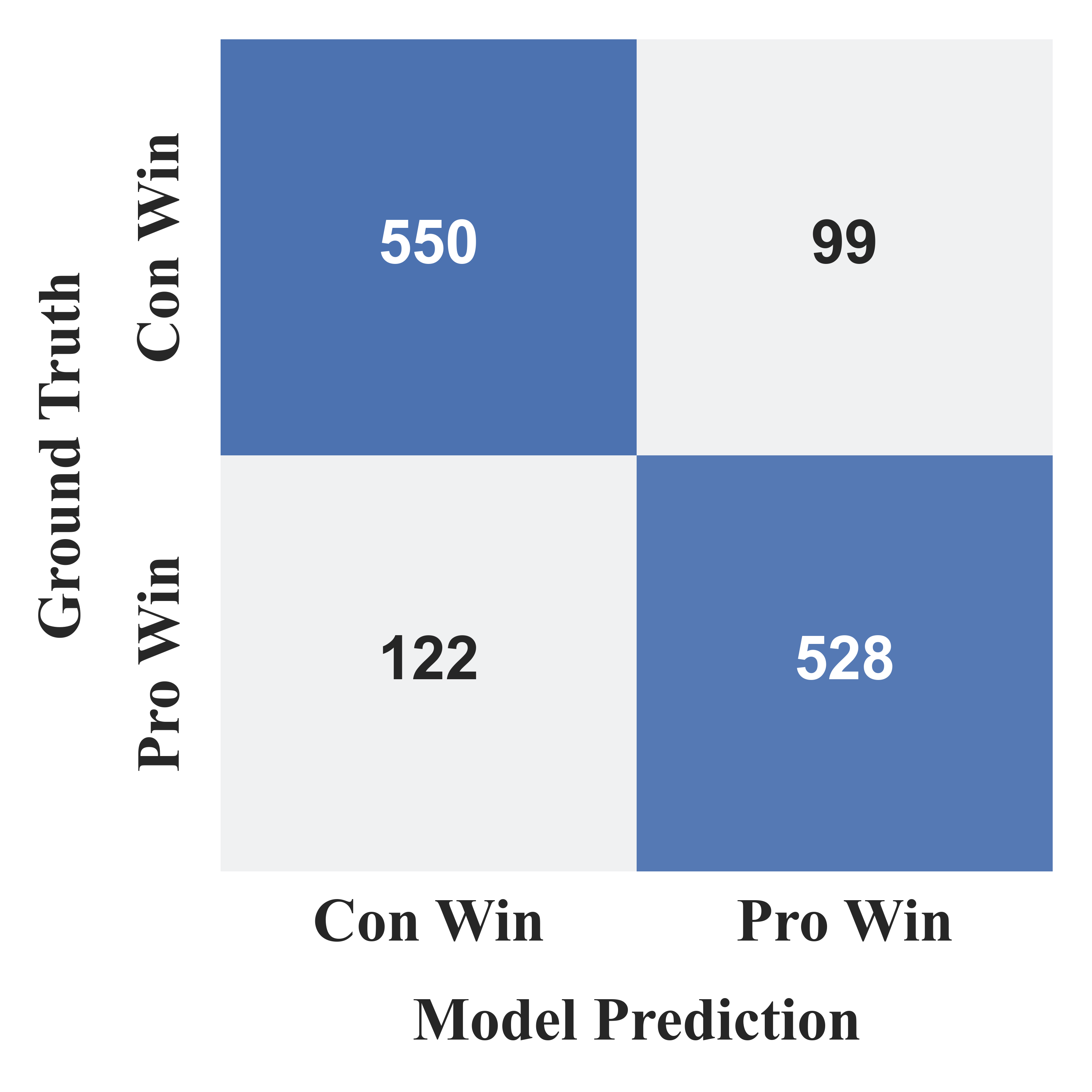}
        \caption{Double shuffled 1/-1 in GPT-4}
        \label{fig:stance bias sub4}
    \end{subfigure}
    \caption{Both the assignment of labels within each label set and the positions of labels are shuffled. These matrcies demonstrate that after eliminating the influence of the order bias, positional bias, and lexical bias, GPT-3.5 shows a stance bias towards the Pro stance, while GPT-4 shows a stance bias towards the Con stance.}
    \label{fig:stance bias}
\end{figure}

\paragraph{Stance Bias.}
Our hypothesis regarding stance bias is less evident in GPT-4, as it becomes overshadowed by lexical bias after positional bias is mitigated through shuffled positions. We conduct two experiments, employing shuffled label sets and positions under the A/B and 1/-1 configurations, as depicted in Figure \ref{fig:stance bias}. The findings reveal a contrasting residual bias in GPT-4 compared to GPT-3.5, after addressing positional, lexical, and order biases.

\subsection{Extension to Unbalanced Setting of DDO Dataset}
\label{sec: appendix 2}

\paragraph{Positional Bias.}
We additionally explore variations in "Pro" and "Con" predictions when alternating between shuffled and fixed candidate response placements in unbalanced data that reflects the original distribution. These observations, detailed in Table \ref{tab:positional bias on the data set with original distribution}, highlight a consistent pattern.

\begin{table}[h]
\centering
\small
\begin{tabular}{cccc}
\toprule
\textbf{Verbalizer} & \textbf{Position} & \textbf{\# Pred Pro} & \textbf{\# Pred Con} \\
\midrule

\multirow{2}{*}{A/B} & Fixed & 533 & 954 \\
                    & Shuffled & 610 & 881 \\
\midrule
\multirow{2}{*}{P/C} & Fixed & 494 & 1000 \\
                      & Shuffled & 727 & 769 \\
\midrule
\multirow{2}{*}{1/-1} & Fixed & 230 & 1267 \\
                       & Shuffled & 340  & 1155 \\
\midrule
\multirow{2}{*}{Pro/Con} & Fixed & 517 & 977 \\
                         & Shuffled & 590 & 904 \\
\bottomrule
\end{tabular}
\caption{Upon fixing and shuffling the positions of labels set as candidate responses in an unbalanced dataset that replicates the original data distribution, the analysis systematically reveals a positional bias towards the second position in GPT-3.5.}
\label{tab:positional bias on the data set with original distribution}
\vspace{-4mm}
\end{table}

\paragraph{Lexical Bias.}
The same experiments applied to the unbalanced dataset with the original distribution yield consistent results for the direction of lexical bias in GPT-3.5 (see Table \ref{tab:lexical bias on the data set with original distribution }), except for the non-significance P/C set.

\begin{table}[th]
\centering
\small
\begin{tabular}{ccc}
\toprule
\textbf{Verbalizer} & \textbf{\# P-Pro} & \textbf{\# P-Con} \\
\midrule

A/B & 610 & 882 \\
B/A & 755 & 734 \\
\midrule
P/C & 727 & 769 \\
C/P & 717 & 835 \\
\midrule
1/-1 & 340 & 1155 \\
-1/1 & 943  & 553 \\
\midrule
Pro/Con & 590 & 904 \\
Con/Pro & 877 & 619 \\
\bottomrule
\end{tabular}
\caption{Upon flipping label sets and shuffling their positions in an unbalanced dataset which replicates the original data distribution, the analysis systematically reveals lexical biases in GPT-3.5 that align directionally with those identified in a balanced dataset. \# P-Pro and \# P-Con denote the number of predicted Pro sides and Con sides as the winner by the model, respectively.}
\label{tab:lexical bias on the data set with original distribution }
\vspace{-4mm}
\end{table}

\subsection{Enhancing Bias Reduction through Prompt Engineering}
\label{sec: appendix 3}


\paragraph{Winning Definition} We find no significant difference in the models’ performance between giving a definition and not giving a definition in the prompt in our preliminary experiments. Therefore, we stick with the more concise version that we illustrate in the main body of the paper. We spectacle it is because our definition of ‘winning’ is consistent with the common understanding of the term. 

\paragraph{LLM-Eval}
In a further step, we direct the LLMs to provide reasons for their judgments before they generate the outcomes using the prompt template shown in Table \ref{tab:extra_template}. Such a method is reported by ~\citet{wang2023large} to be able to reduce the positional bias. We do a pilot experiment using GPT-3.5 with a single A/B label set to see if the effect comes from 'reducing' the bias or from providing a bias in the opposite direction and thus counteract it.

\begin{figure}[ht]
    \centering
    \begin{subfigure}[b]{0.32\linewidth}
        \includegraphics[width=\linewidth]{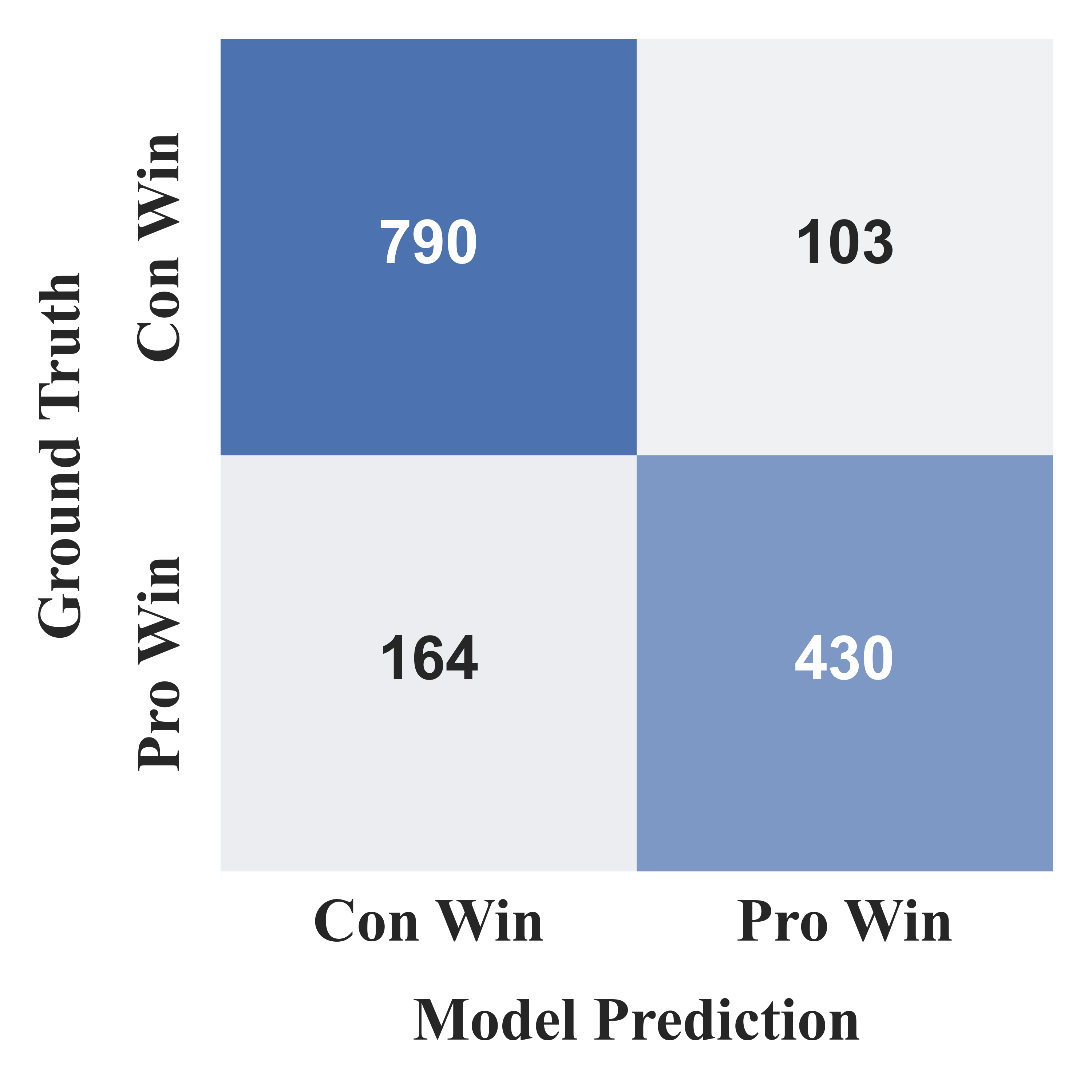}
        \caption{A/B label set}
        \label{fig:eval analysis sub1}
    \end{subfigure}
    \hfill
    \begin{subfigure}[b]{0.32\linewidth}
        \includegraphics[width=\linewidth]{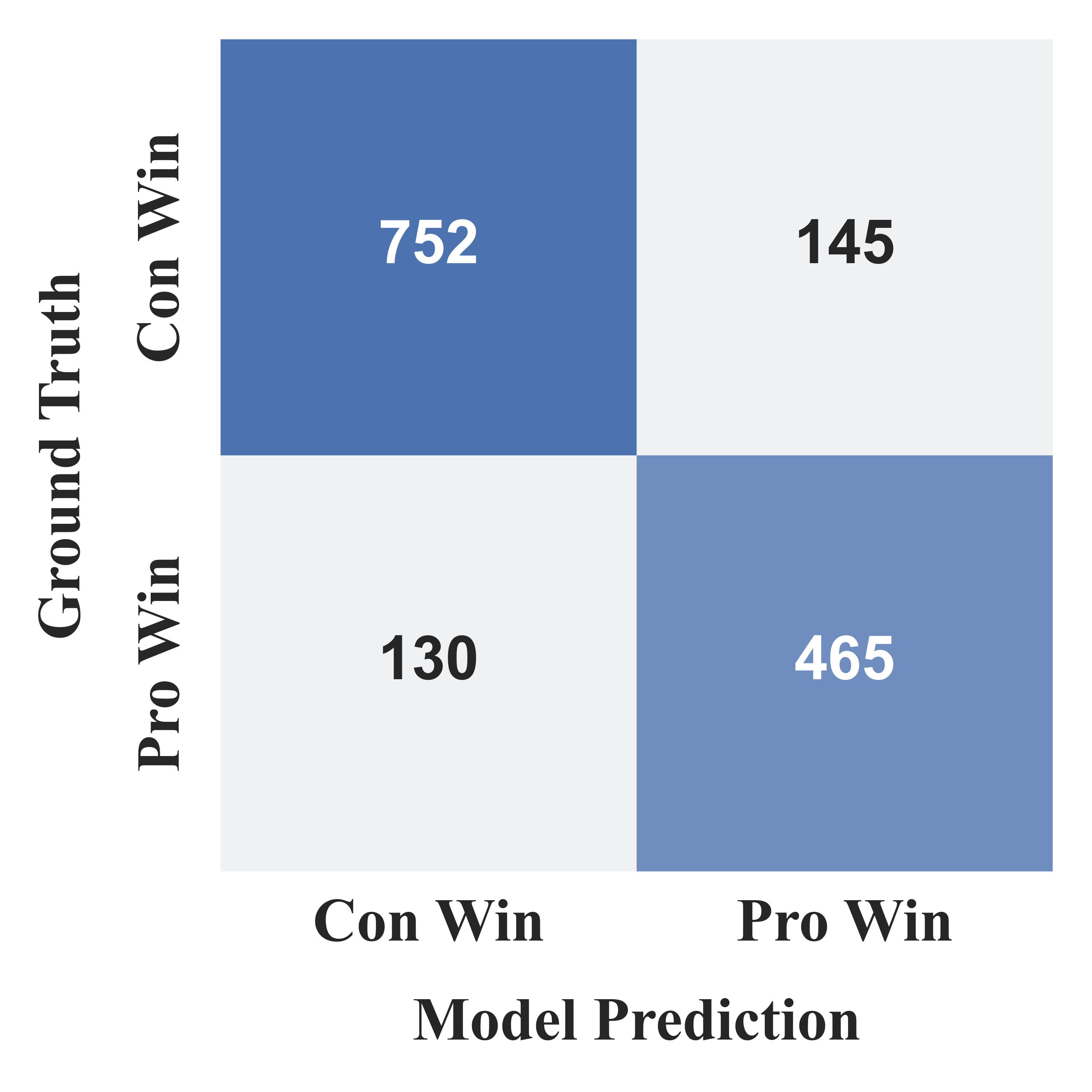}
        \caption{Shuffled A/B label set}
        \label{fig:eval analysis sub2}
    \end{subfigure}
    \hfill
    \begin{subfigure}[b]{0.32\linewidth}
        \includegraphics[width=\linewidth]{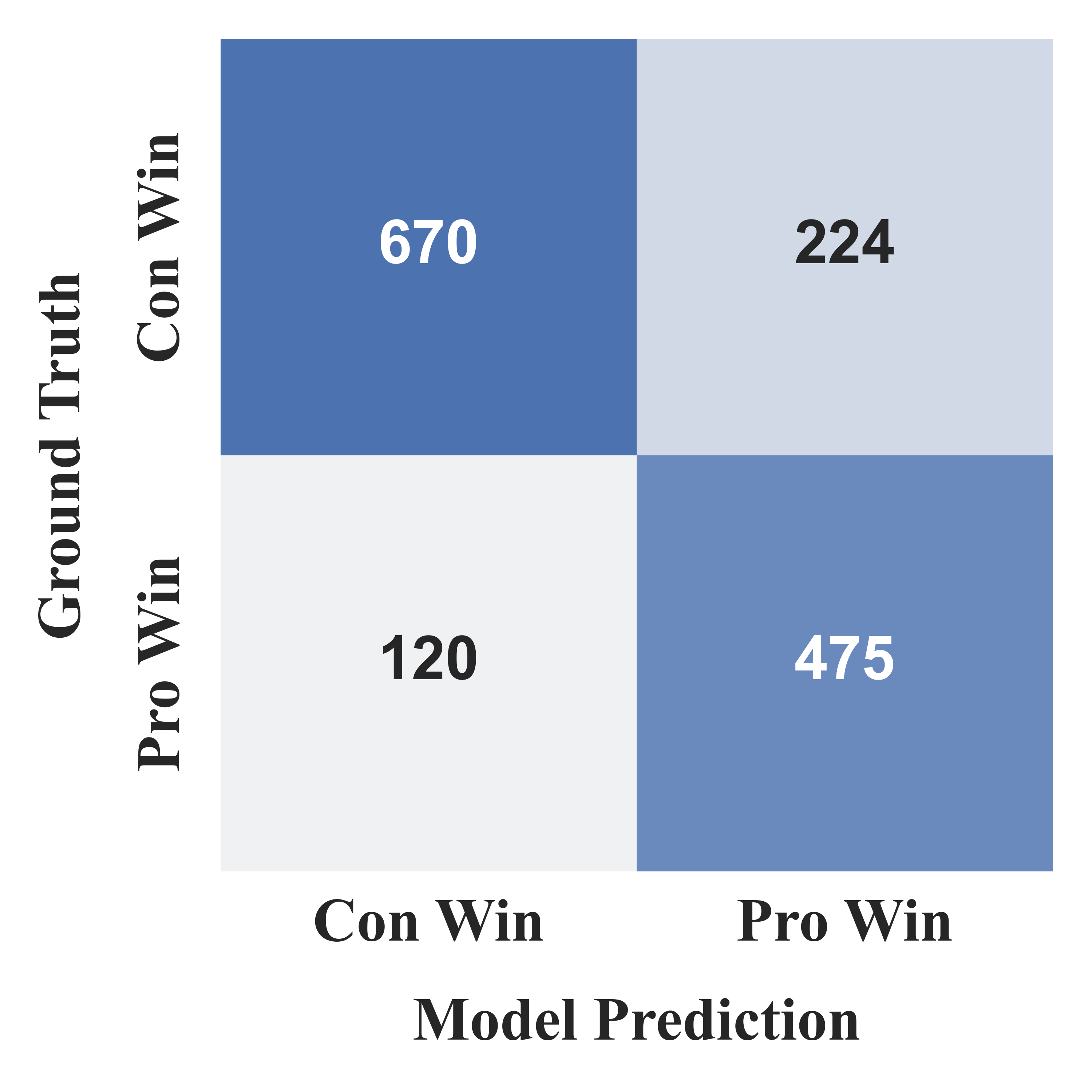}
        \caption{A/B label set with Evaluation}
        \label{fig:eval analysis sub3}
    \end{subfigure}
    \caption{Analysis of the effect of generating analysis on reducing positional bias.}
    \label{fig: eval analysis}
\end{figure}

As the results shown in Figure \ref{fig: eval analysis}, GPT-3.5 exhibits a greater bias towards Pro when generating analysis compared to when positions are shuffled to eliminate the positional bias. Therefore, it is more likely that prompting GPT-3.5 to generate the analysis first introduces a new bias towards Pro, which is in the opposite direction of the positional bias, since Con is consistently positioned as the second candidate response. However, arriving at a definitive answer necessitates further experimentation, which we defer to future research.

\begin{table*}[t]
\centering 
\small
\begin{tcolorbox}
\textbf{Content Prompt}

The content of the whole debate:

The current speech in the debate is from the user \textcolor[rgb]{0,0,0.9}{\{Side1\_label\}}:

[The content of the side1]

The current speech in the debate is from the user \textcolor[rgb]{0,0,0.9}{\{Side2\_label\}}:

[The content of the side2]

The current speech in the debate is from the user \textcolor[rgb]{0,0,0.9}{\{Side1\_label\}}:

[The content of the side1]

\ldots

\textbf{Vanilla Prompt}

Assume you are a debate evaluator, there are two participants in this debate. Given the above context of the whole debate. Please give the decision on which participant is the winner, you only need to give the character(number) of either \textcolor[rgb]{0,0,0.9}{\{Side1\_label\}}, or \textcolor[rgb]{0,0,0.9}{\{Side2\_label\}}. \textcolor[rgb]{0,0,0.9}{\{Side1\_label\}} means user \textcolor[rgb]{0,0,0.9}{\{Side1\_label\}} wins. \textcolor[rgb]{0,0,0.9}{\{Side2\_label\}} means user \textcolor[rgb]{0,0,0.9}{\{Side2\_label\}} wins. Please only give the result without any other words.

\textbf{Eval Prompt}

Assume you are a debate evaluator, there are two participants in this debate. Given the above context of the whole debate, please provide a comprehensive explanation of your evaluation, avoiding any potential bias and ensuring that the order in which the responses were presented does not affect your judgment. Finally, decide who wins the debate. Output with the following format:\\
Evaluation: \\
<your comprehensive evaluation explanation here>\\
<winner \textcolor[rgb]{0,0,0.9}{(\{Side1\_label\}} or \textcolor[rgb]{0,0,0.9}{\{Side2\_label\}})>\\
The final line of your output should contain only one word: \textcolor[rgb]{0,0,0.9}{\{Side1\_label\}} if you conclude that user \textcolor[rgb]{0,0,0.9}{\{Side1\_label\}} wins, or \textcolor[rgb]{0,0,0.9}{\{Side2\_label\}} if you conclude that user \textcolor[rgb]{0,0,0.9}{\{Side2\_label\}} wins. No tie or inconclusive results are allowed.

\end{tcolorbox}
\vspace{-0.3cm}
\caption{The "Vanilla Evaluation" prompts the model to predict results directly based on the content prompt. 
The "Eval Prompt" mandates the model to evaluate arguments for both sides and provide a holistic assessment based on the “Content prompt”.}
\label{tab:extra_template}
\end{table*}

\subsection{Debate Example}
\label{sec: appendix 4}

The debate example can be found in Table \ref{tab:debate_example 1}, \ref{tab:debate_example 2} and \ref{tab:debate_example 3}.

\begin{table*}[th]
\centering 
\small
\begin{tcolorbox}
\textbf{The current speech in the debate is from the user \textcolor[rgb]{0,0,0.9}{\{Side1\_label\}}}:

Thank you, to whoever accepts this challenge, I look forward to this debate.  
  
Now, to start off, I will go over some definitions.  
  
Morality: [conformity to the rules of right conduct]  
Evil: [morally wrong or bad; immoral; wicked]  
Right: [in accordance with what is good, proper, or just]  
Atheist: [a person who denies or disbelieves the existence of a supreme being or beings.]  
Theist: [the belief in one God (in this debate I am referring to the Christian God) as the creator and ruler of the universe, without rejection of revelation] 
  
  http://dictionary.reference.com...  
  
To begin with, what makes something wrong or right? The law of a specific nation? Yourself? This question was simple when we were kids, for instance if John hit Sue then John was wrong and then gets in trouble. But as we get older this topic becomes more complicated. For instance, who said it was wrong for John to hit Sue? Who said it was wrong for someone to steal, cheat, lie, murder, torture, rape? The point is, in some cultures its acceptable and even encouraged to do these things. Just look at Hitler, Stalin or any other evil dictator/government. Anyone can read about the trattorias acts that have been accurately recorded through out history. But here's the thing, all these men committed terrible acts without believing that they themselves were 'wrong.' For example, Hitler murdered 10 million people for ethnic cleansing reasons, and through out his entire life as ruler over Germany, never once thought he was doing an immoral act. In fact, he believed he was doing just the opposite, Hitler thought, that through killing 10 million people he was "glorifying the Father Land" and doing the world a huge favor. Plus, Hitler not only was evil himself, but he had a whole nation behind him. Millions swore true allegiance to him, and his ideas.  
  
Now, given the above paragraph, it is impossible to say that Hitler's actions were immoral under an Atheistic world view. Why? Because in an Atheistic world view there is no God to judge such acts. The only thing that can judge Hitler in an Atheistic world are other people, but what if every single person on the planet became a Nazi. So there must be an ultimate judge, or over seer, in order for Hitler's actions to be held accountable.  
  
So, if one wants to debate that morality is defined by the law of a specific nation, or ones ability to justify there own actions, then the voters and my oppenent should be able to see clearly that an Atheistic world view can not account for morality.  
  
Please answer the following questions in your next argument.  
  
How can Atheism account for morality? And what will you base you morality off of, if not God?

\textbf{The current speech in the debate is from the user \textcolor[rgb]{0,0,0.9}{\{Side2\_label\}}}:
  
This should be an interesting debate.... I love this sort of debate, ie. what are morals and why do we have them sort of thing  
  
"Because in an Atheistic world view there is no God to judge such acts. The only thing that can judge Hitler in an Atheistic world are other people, but what if every single person on the planet became a Nazi."  
  
First If everyone was a Nazi there would be no problem with Nazism because they wouldn't have a WW2 Repeat due to the fact that everyone would agree....  
  
Second the people/self being the judge is what I am arguing. The Ultimate judge is humanity. The concept of the Other best applies here. When we look at another acting, we judge them. When we look at ourselves acting the same way, we remember that judgment. We don't have to actually see someone else, but imagine that there is that Other judging us. Also if we look to the roots of morality we don't find God, but humanity. Why is it immoral to kill? Because if it was allowed then people would freely kill us. If we look at what we would think had we seen the event happen, or been the recipient, we will agree that the event is bad. From All this we can take morality to really be a golden rule of sorts. Judge ourselves as we would judge others. Do to others as we would have done to ourselves. Neither of these concepts require God, in fact they function just as well with a God as without.  
  
"How can Atheism account for morality? And what will you base you morality off of, if not God?"  
I have already sort of answered this but I will do it again for sake of order and clarity. Atheism accounts for morality via Humanity. The roots of our morals exist in an atheist society, they were created not by God but by human conscience and need for order and safety. I don't want to retype the explanation of the Other(which was admittedly pretty bad) but that is a general concept of how atheism can account for and provide a base for morality. The golden rule is another base for morality. Morals Exist for human safety primarily. Why is it immoral to kill? because we don't want to be killed.  
  
God is not the source of Morals, and therefore an atheistic world view can account for morals just as well as a theistic world view can.

\end{tcolorbox}
\vspace{-0.3cm}
\caption{The first round of a debate example.}
\label{tab:debate_example 1}
\end{table*}

\begin{table*}[th]
\centering 
\scriptsize
\begin{tcolorbox}
\textbf{The current speech in the debate is from the user \textcolor[rgb]{0,0,0.9}{\{Side1\_label\}}}:

Thanks for your response.  
  
Metz said, "Second the people/self being the judge is what I am arguing. The Ultimate judge is humanity."  
  
To say that humanity is the ultimate judge is not saying anything. For instance, in one part of the world it may be morally acceptable to murder your wife if she disobeys her husband. In another part of the world that particular act may be unacceptable. But, which view of the issue is right? Who decides it? The point is, that to base what is considered right or wrong off humanity is ridiculous, since humanity can not agree on an absolute, universal view of what is considered moral or immoral. Since this is true anything could be acceptable, such as murder, rape, lying cheating, abusing, drugs ect... Why? Again, because morality is totally arbitrary under the jurisdiction of humanity, since all humans have different standards of morals. And since all humans have different standards on morals, then this just illustrates my point, there must be a God to judge people's actions. In an Atheistic world there are no absolutes for morals.  
  
Also, if there are seven hundred billion people on the planet and half say gay marriage is right but the other half say gay marriage is wrong, then who decides? What makes one view right and the other wrong? This question can not be answered in an Atheistic universe, since all the opinions given by the people are different. So, humanity, can not, on it's own make a rational decision, dealing with morality. This is why there must be an objective standard for people to base their judgement off of. Again, under an Atheistic world view morals can not be accounted for.  
  
"Why is it immoral to kill? Because if it was allowed then people would freely kill us."  
  
What about the people who could care less about whether or not death is a reaction of killing another person. For instance, a man could be very enraged at a particular moment, so, what if he decides to kill everyone in the town regardless of wether he dies that day or lives, in the process of committing all the murders he can. Not only that flaw, but there are people who murder people all the time without getting caught, or getting killed back in the process. So for these murderers there is no incentive what so ever for them to not go out and murder another human being.  
  
Plus, saying that its immoral to murder because you will get murdered back is not even answering the question of why it is immoral to murder another human being. You need to tell me why murder is wrong in the first place.  
  
Metz said "Do to others as we would have done to ourselves."  
  
Its amazing how Atheists think, they will always claim there world has morals, and do things such as feed the poor and help many in need ect... These are all good things, its just the principles in which these acts are found, are in the Bible. You see, Atheists take morals from the Christian world view but do not acknowledge the basis of which those morals came from, which is ultimately God. Now I'm not saying that all of the morals in an Atheistic world view are taken from Christianity, but a lot of them are, Along with many other religions that acknowledge the presence of a god.  
  
Metz said, "Also if we look to the roots of morality we don't find God, but humanity."  
  
Prove to me that we find humanity, don't just say it, prove it or at least tell expand on that reasoning. I do not agree with that statement at all and until you try to prove it it is just your word against mine. Which is exactly what an atheistic world view consists of, one man's word against another, which is no absolutes or universal ideas  
  
I also encourage the voters to check out this link, it will help illustrate my point.  
  
Thank you  
charles 15  
  
Good Luck

\textbf{The current speech in the debate is from the user \textcolor[rgb]{0,0,0.9}{\{Side2\_label\}}}:
"To say that humanity is the ultimate judge is not saying anything. For instance, in one part of the world it may be morally acceptable to murder your wife if she disobeys her husband. In another part of the world that particular act may be unacceptable. But, which view of the issue is right? Who decides it? The point is, that to base what is considered right or wrong off humanity is ridiculous, since humanity can not agree on an absolute, universal view of what is considered moral or immoral."  
  
But this accounts for morality... it just doesn't account for my opponents version of morality. Also this really doesn't say why Theism can actually account for universal morals. People disagree on religion. If Morals were universal then the scenario my opponent laid out wouldn't exist. But yet he claimed it does.... So what my opponent is essentially arguing is that Morality doesn't work.  
  
"In an Atheistic world there are no absolutes for morals."  
  
Ok... Same thing in a Theist world. But lets look at the topic for a moment shall we? It never says Atheism needs to account for universal morals, just morals. This really doesn't attack my case at all. The Definition of Morality my opponent gives is "conformity to the rules of right conduct" But it never says these rules must be universal. If we have laws they do not hold everyone accountable worldwide, likewise morality doesn't have to be universal.  
  
"Again, because morality is totally arbitrary under the jurisdiction of humanity, since all humans have different standards of morals."  
  
That is how I argue we can account for morality. If we want to find acceptable morality we need people to disagree, this is how democracy works and how morality would inevitable work. And yet again, Theism is different how?  
  
"Also, if there are seven hundred billion people on the planet and half say gay marriage is right but the other half say gay marriage is wrong, then who decides? What makes one view right and the other wrong? This question can not be answered in an Atheistic universe, since all the opinions given by the people are different. So, humanity, can not, on it's own make a rational decision, dealing with morality."  
  
Oh yeah... and God is doing so much better? The reason so many people disagree is primarily religion...granted there are other factors but religion and tradition are massive players.  
  
"So, humanity, can not, on it's own make a rational decision, dealing with morality"  
  
Well actually we live in a largely theist world... so what you meant to say was " So, God and religion cannot make a rational decision dealing with morality"  
  
"This is why there must be an objective standard for people to base their judgement off of"  
  
yeah, its called survival mate.... people see other and judge themselves... People tell others that a certain action is wrong because they don't want what they see done to other done to themselves...  
  
"You need to tell me why murder is wrong in the first place."  
  
Its wrong because people say its wrong... you essentially made my argument for me there; "its immoral to murder because you will get murdered back" it isn't moral to Murder because you are ending that persons existence. I don't want to end my existence so I tell people that it is wrong to kill. If I wanted to be killed would I say it is wrong to kill?  
  
"You see, Atheists take morals from the Christian world view but do not acknowledge the basis of which those morals came from, which is ultimately God"  
  
Um... Alright... The First appearance of the golden rule was I believe in the Analects of Confucious... Not the bible. Also it really doesn't matter where the Morals came from as long as an Atheist world can account for them... I personally have a justification for all my moral opinions that has nothing to do with god but with how I perceive humans.  
  
"Metz said, "Also if we look to the roots of morality we don't find God, but humanity."  
Prove to me that we find humanity, don't just say it, prove it or at least tell expand on that reasoning."  
  
That I will be glad to do.... Name any generally accepted moral principle and I will show how it can be traced back to humanity. Also my opponent again makes the mistake of saying Atheism cannot account for UNIVERSAL MORALS, but sadly neither can Thiesm as we have seen and that is not the subject of this debate.  
  
Lets do an example of morality being human using the Moral principle that killing is wrong.  
  
1. Humans don't want to be killed  
2. People, as a general rule, want to do what they feel is right.  
3. Therefore people(in general), because they don't want to be killed, have said that killing is wrong  
4. Therefore it is generally accepted among people killing is wrong  
5. Hence killing is considered an Immoral Act.  
  
Justification behind 1-5.  
  
1. The Urge for Survival in all things is primary, it has been seen through the existence of life  
2. The concept of the conscious tells us that we want to do the right thing. So people are deterred by the idea that what they may be doing is wrong.  
3. Combination of 1\&2 plus the fact that people made this decree to create the deterrence I mentioned in 2  
4. A summary of 4 as a general rule  
5. Putting the concept of right/wrong into Morals  
  
Thank you,  
Matt  
  
Good luck to my opponent, and I urge everyone to look critically at all arguments

\end{tcolorbox}
\vspace{-0.3cm}
\caption{The second round of a debate example.}
\label{tab:debate_example 2}
\end{table*}

\begin{table*}[th]
\centering 
\scriptsize
\scalebox{0.83}{
\begin{tcolorbox}
\textbf{The current speech in the debate is from the user \textcolor[rgb]{0,0,0.9}{\{Side1\_label\}}}:

Metz said, "Also this really doesn't say why Theism can actually account for universal morals. People disagree on religion. If Morals were universal then the scenario my opponent laid out wouldn't exist."  
  
I thought I made this clear in my opening statement: the God I am referring to is the Christian God. So, when you say that a theistic world view can not account for morality, because of all the different religions, then yes I would agree with you, because if there are many different religions that judge humans, then there would not be one standard to which morality is based upon. So, when I mention God I am only referring to the Christian God. Now that my view on the issue has been restated, any argument used by Metz (con), about why a theistic universe can not account for morality either; because of all the different God's derived from different religions, will be irrelevant. Since I am only referring to One religion, which is Christianity. And since there is just one God then there is only one moral standard, thus God can account for what is wrong or right.  
  
"It never says Atheism needs to account for universal morals, just morals."  
  
Okay, lets have it Metz's way, Atheism does not need to account for universal laws, just morals in general, very well. If there are no universal morals that prohibit certain acts of crime such as rape, murder, polygamy, theft, ect... then why am I obligated to obey those morals? Why can't I just abide by my own moral standards, since there are know Universal ones? For instance, I could think that its just fine to murder, rape, steal ect... because that's what I believe is right. So if there are just MORALS, to be defined by anybody, and no UNIVERSAL MORALS then who is to say that my morals are wrong? Whose to say anything is wrong for that matter? Once again the argument for an Atheistic world view on morals collapses on itself because it can not account for what is truly right or wrong.  
  
Metz said, "Its wrong because people say its wrong..." this quote is in response to me asking why murder is wrong.  
  
So are you saying that if the majority of the human population say that gays should not be aloud to marry, then that is automatically the moral standard? This is exactly my point, if a moral act is defined by what people say is moral then anything from the act of murder to a little white lie must be accepted by humanity. For instance, I could say murder is right because I said so. Also, a real life example is, 'people' started 'saying' that Jews should be considered sub human and thrown into concentration camps, but did this make it right? No, of course not. You see, I can say that Hitler was wrong because God commands it in the Bible, "thall shall not murder," its the 6th commandment. But, the best that my opponent can say, is, "Hitler's acts of genocide were immoral because the Jews were being murdered against there own will." Well, my opponent's statement just begs the question, So? Who says the Jews have a right to live in the first place? After all, millions were saying that Jews did not have the right to live. So which side is right, and why? For, to simply say that Hitler was wrong because he murdered Jews against there will is NOT answering the WHY? It only states a mere fact.  
  
Another example, John Locke, a well known philosopher who came up with idea of the Social Contract, this contract was to ensure that every human being was born with the right to live. Now, the question I have for Locke's thinking, along with anyone else who agrees with him, is this, WHY? Why are humans born with the right to live? I do not see a logical answer without God in the equation.  
  
1. Humans don't want to be killed.  
2. People, as a general rule, want to do what they feel is right.  
3. Therefore people(in general), because they don't want to be killed, have said that killing is wrong  
4. Therefore it is generally accepted among people killing is wrong  
5. Hence killing is considered an Immoral Act.  
  
Again, not only does this example have nothing to do with WHY murder is wrong. But, what you have described here is Western Civilization for the past 200 years or so, ONLY. This certainly is not the case in the Philippines, the Middle East, or any other extremely violent area in the world. This totally disproves your point above. And not only that, but what about in past history, such as the Dark Ages where many people considered murder to be a normal act, in order to get food or money so they could fill there bellies. So, when murder became an act that was generally accepted among the people, such as in the Philippines, the Middle East, and any other extremely violent areas in the world or from times in the past, such as the Dark Ages, is it then morally acceptable to murder? I see no reason why not, under an Atheistic universe.  
  
In conclusion, I still believe that my opponent has failed to answer the why for his reasoning? For instance, everyone knows that people don't want to be murdered. But the question I am asking is, why is it wrong for people to be murdered? To say because people don't want to be murdered is not answering the question. Because why should a murderer care about what his or her victim wants if its just a question of morals and not universal morals? Also my opponent argue that a theistic world view can not answer for this question either. Well, that isn't answering the question, that's just pointing fingers.  
  
As I have said, I am a Christian and will be basing my arguments off a single religion and a single God. Now, my opponent may take this as a opportunity to criticize my religion like he did in his last argument, some what. If my opponent starts to argue that Christianity is not perfect and why should God be the ultimate judge this is still not answering the question of why anything is right or wrong to do anything. Again its just pointing fingers.  
  
Now this is something I have only touched on a little, I can say something is wrong or right because I believe there is an ultimate judge, God. This means there are universal laws of morality, that are absolute, and everyone must abide by them. In an Atheistic universe the only thing that can judge morality is humanity which I have proved is inconsistent and ultimately can not account for morality at all.  
  
Again I encourage the voters to listen to the video above it really illustrates my point.  
  
My dad also had a personal relationship with Dr. Bahnsen (the man debating in the video). My dad told me that after the debate between Bahnsen (Christian) and Stein (Atheist) they continued debating each other through emails and letters, after a couple weeks of going back and forth with their arguments Stein eventually wrote "I don't really have any answers for you, but I'm just not ever going to agree with you."  
  
Please answer the fallowing questions...  
  
1)Why should a murderer care about what his or her victim wants if its just a question of morals?  
2)Why are humans born with the right to live? I do not see a logical answer without God in the equation.  
3)Are you saying that if the majority of the human population say that gays should not be aloud to marry, then that is automatically the moral standard?  
  
Thank you,  
charles15

The current speech in the debate is from the user \textcolor[rgb]{0,0,0.9}{\{Side2\_label\}}:

I will start with the three questions my opponent proposed to me at the end of his last argument.  
  
1)Why should a murderer care about what his or her victim wants if its just a question of morals?  
There are, obviously exceptions to my rule of moral deterrence. But remember my proof established it as a general rule. This has nothing to do with Atheism at all, when someone murders someone they are not in a state of mind that would disregard any moral background whatsoever. Even if we assume a theist stance, these people have committed a sin, so therefore God as much fails to uphold morals as would Atheism. Also the Psychological consequences would be felt later as philosopher and psychologist Fyodor Dostoevsky laid out in his book Crime and Punishment.  
  
2)Why are humans born with the right to live? I do not see a logical answer without God in the equation.  
  
I hate to say this but its the shocking truth... We are born with the right to live because we have a will to live. If nobody wanted to vote would it be considered a right? This will to live is also not traceable to god, but to the fact that humans are just animals with the ability to reason. Unless my opponent wishes also to deny evolution and biological fact then this has to be accepted. The most primal instinct of live is to preserve itself. This is where morals come from as I have repetedly argued. Humans judging others and therefore judging themselves.  
  
3)Are you saying that if the majority of the human population say that gays should not be aloud to marry, then that is automatically the moral standard?  
  
This is Mob rule, not necessarily morality. But not to criticize to much but I have that the same would be said of God. If the Bible says it then its wrong, which seems to be a common belief about gay marriage. As I said The base of Morality is humans, Gay marriage does not threaten anybody, so it is therefore it is not sought to prevent like killing would be. People Judge others in Gay Marriage but it does not affect them so the link between natural morals is flawed. A society may come to the belief that gay marriage is immoral, but it is not intrinsically immoral, and this seems to be what is happening in the world today.  
  
Now on to the remaining arguments:  
  
"Since I am only referring to One religion, which is Christianity. And since there is just one God then there is only one moral standard, thus God can account for what is wrong or right."  
  
This really doesn't mean that everyone would follow this God, so are these people immoral? People believe do different extents, and so therefore have different morals even assuming the same God and religious texts and Church structure. In order for God to be as great a source for morals as my opponent claims we would need to abandon any remaining Autonomy and become almost robotic in our beliefs, an act which is, ironically, immoral in either world.  
  
"So if there are just MORALS, to be defined by anybody, and no UNIVERSAL MORALS then who is to say that my morals are wrong? "  
  
Not defined by anybody, defined by humanity. Humans Judge, you are judged by your fellows, you judge others and so judge yourself. Every step of the way there are checks.  
  
" If there are no universal morals that prohibit certain acts of crime such as rape, murder, polygamy, theft, ect... then why am I obligated to obey those morals? Why can't I just abide by my own moral standards, since there are know Universal ones? For instance, I could think that its just fine to murder, rape, steal ect... because that's what I believe is right."  
  
First, I Never said Atheism CAN'T account for universal morals merely that it was not my burden to prove that it did. Also, you can have your own moral standards, I know many people that have there own and are not killers, for example I think we have a moral obligation to fairness and to help people, I have friends that have a more sink or swim attitude. These morals can be relative, this is part of what shapes humanity, to accept that all morals are dictated to us really destroys that humans element. However when we get into killing, people judge more carefully, people are afraid. For the sake of protection and for moral order HUMANS establish moral rules, such as that against killing. Atheism can account for Morality because it was humans all along that accounted for morality.  
  
"Who says the Jews have a right to live in the first place? "  
They do... They have a will to live that is as strong as that of any other. This turns Life into a right intrinsic of humanity. Thus when the Jews were killed Hitler was taking an intrinsic right and the act was thus, immoral. I already addressed the other problem at the beginning.  
  
"everyone knows that people don't want to be murdered. But the question I am asking is, why is it wrong for people to be murdered?"  
  
You gave me the answer right there. This bring me back to the same Will to Live argument. It is wrong because people have a will to live. Because they have this will it becomes a recognized right to live. Thus when someone violates this right the act is immoral in most circumstances(there are exceptions to every moral idea).  
  
Voters,  
When you are reading this debate you need to think about whether or not you would logically do some of the things my opponent has said in his examples, and whether you would want them done to yourself. You also must recognize that Murder's generally have an altered or disturbed state of mind that could be influenced by such things as Alcohol that means in Either world these people don't respect morals.  
  
The key question here is: Did my opponent prove that without God morals COULD NOT exist? Or did I prove that morals COULD exist in such a world. Remember the resolution asks could, which means "is it possible"  
  
Thanks,  
Metz

\end{tcolorbox}}
\vspace{-0.3cm}
\caption{The third round of a debate example.}
\label{tab:debate_example 3}
\end{table*}

\subsection{Extension to IQ2 Dataset in the Balanced Setting with GPT-4}
\label{sec: appendix 5}
There are 108 debates in the IQ2 dataset. The average number of words contained in each debate, including all contexts, is 17579, exceeding the current maximum length constraint of GPT-3.5 (16k tokens). Only 24 debates in IQ2 have a word count below this limit, which would result in a sample size too small to derive meaningful results. While excluding the context from the host or audience involved could reduce the average length of each debate to 12801 words, it could also lead to a lack of context in some parts of the debaters’ conversation. Therefore, we only analyze IQ2 dataset on GPT-4 with a 32k token limit. We again use the balanced setting as we explained in the Methodology section. Based on the smallest category (Con end with Con win) among the four conditions, we sampled IQ2 to be 13 for pro/con side end with pro/con win, a total of 52 samples. 

\paragraph{Positional Bias.} GPT-4 exhibits consistent positional bias on the IQ2 dataset, as shown in Table \ref{tab: positional bias on IQ2}. The second position is preferred over the first position, proved by the higher proportion of Predicted Con when Con is positioned as the second candidate response.

\begin{table}[h]
    \centering
    \small
    \begin{tabular}{c|c|c}
         \hline
          Verbalizer (Pro/Con) & Positions & Predicted Con Proportion \\
         \hline
          A/B & Con Second & 94.23\%  \\
          A/B & Pro Second & 34.62\% \\
          \hline
          B/A & Con Second & 34.62\%  \\
          B/A & Pro Second & 3.85\% \\
         \hline
    \end{tabular}
    \caption{GPT-4 shows positional bias on IQ2 with 52 balanced samples.}
    \label{tab: positional bias on IQ2}
\end{table}

\paragraph{Lexical Bias.} We find consistent lexical biases in the IQ2 dataset with GPT-4, as shown in Table \ref{tab:lexical bias on IQ2}. `B’(`-1’) is preferred over `A’(`1’), indicated by the higher proportion of Predicted Con when `B’(`-1’) represents Con compared to when `A’(`1’) represents Con. 

\begin{table}[h]
    \centering
    \small
    \begin{tabular}{c|c|c}
         \hline
          Verbalizer (Pro/Con) & Positions & Predicted Con Proportion \\
         \hline
          A/B & Con Second & 94.23\%  \\
          B/A & Con Second & 34.62\%  \\
          \hline
          A/B & Pro Second & 34.62\% \\
          B/A & Pro Second & 3.85\% \\
          \hline       
          1/-1 & Con Second & 65.38\% \\
          -1/1 & Con Second & 42.31\% \\
         \hline
    \end{tabular}
    \caption{GPT-4 shows lexical bias on IQ2 with 52 balanced samples.}
    \label{tab:lexical bias on IQ2}
\end{table}

\paragraph{Order Bias.} GPT-4 exhibits order bias on the IQ2 dataset, which is also consistent with our finding on DDO dataset, as demonstrated by Table \ref{tab:chi_test_for_order_bias_in_GPT4}. The ending side of a debate is more likely to be predicted as the winner.
\begin{table}[h]
\centering
\footnotesize 
\scalebox{0.95}{
\begin{tabular}{cccc}
\toprule
\textbf{Verbalizer} & \textbf{End-Side} & \textbf{\# P-Pro} & \textbf{\# P-Con} \\
\midrule
\multirow{2}{*}{1/-1} & Pro & {\bf 11} & 15 \\\
             & Con &  7 & {\bf 19}  \\
\bottomrule
\end{tabular}}
\caption{Order bias shown by GPT-4 on IQ2 with 52 balanced samples.}
\vspace{-5mm}
\label{tab:chi_test_for_order_bias_in_GPT4}
\end{table}

